\documentclass[pmlr]{jmlr}% new name PMLR (Proceedings of Machine Learning)
\RequirePackage{graphicx}
 \usepackage[section]{algorithm}
\usepackage{booktabs}
\usepackage{fancybox} % For plotting tiktz boxes.
\usepackage{tikz}
\usepackage{bm}
\usetikzlibrary{positioning}
\usepackage{longtable}% for long tables
\makeatletter
\def\set@curr@file#1{\def\@curr@file{#1}} %temp workaround for 2019 latex release
\makeatother
\usepackage[load-configurations=version-1]{siunitx} % newer version

\theorembodyfont{\upshape}
\theoremheaderfont{\scshape}
\theorempostheader{:}
\theoremsep{\newline}

\def\mR{\mathbb{R}}

\def\d{\mathrm{d}}

\def\refeq#1{(\ref{#1})}

\def\marg#1{{\underline{#1}}}
\DeclareMathOperator{\Beta}{Beta}
\DeclareMathOperator{\Dir}{Dir}
\DeclareMathOperator{\Mult}{Mult}

\DeclareMathOperator{\Gam}{Gam}
\DeclareMathOperator{\Pois}{Pois}
\DeclareMathOperator{\DirMult}{DirMult}

\DeclareMathOperator{\CRT}{CRT}
\DeclareMathOperator{\NB}{NB}

\DeclareMathOperator{\Log}{Log}
\DeclareMathOperator{\SumLog}{SumLog}

\DeclareMathOperator{\U}{U}

\DeclareMathOperator{\GCRTP}{CRTCP}
\DeclareMathOperator{\Polya}{Polya}

\newcommand{\ignore}[1]{}
\newcommand*\mathinhead[2]{\texorpdfstring{$\boldsymbol{#1}$}{#2}}
\newcommand{\subf}[2]{%
    {\small\begin{tabular}[t]{@{}c@{}}
    #1\\#2
    \end{tabular}}%
  }
\jmlrvolume{[VOLUME \# TBD]}
\jmlryear{2024}
\jmlrworkshop{Machine Learning for Healthcare}

\title[Multinomial belief networks]{Multinomial belief networks for healthcare data}

\author{\Name{H. C. Donker}
       \Email{h.c.donker@umcg.nl}\\
       \addr Department of Epidemiology\\
       University Medical Center Groningen\\
       Groningen, Groningen, the Netherlands.
       \AND
       \Name{D. Neijzen}
       \Email{d.neijzen@umcg.nl}\\
       \addr Department of Epidemiology\\
       University Medical Center Groningen\\
       Groningen, Groningen, the Netherlands.
       \AND
       \Name{J. de Jong}
       \Email{johann.de\_jong@boehringer-ingelheim.com}\\
       \addr Statistical Modeling \\
       Global Computational Biology and Data Sciences\\
       Boehringer Ingelheim Pharma GmbH \& Co. KG\\
       Biberach an der Ri{\ss}, Germany.
       \AND
       \Name{G. A. Lunter}
       \Email{g.a.lunter@umcg.nl}\\
       \addr Department of Epidemiology\\
       University Medical Center Groningen\\
       Groningen, Groningen, the Netherlands.
       }

\begin{document}

\maketitle

\begin{abstract}
  Healthcare data from patient or population cohorts are often characterized by sparsity, high missingness and relatively small sample sizes.  In addition, being able to quantify uncertainty is often important in a medical context.
  To address these analytical requirements we propose a deep generative Bayesian model
  for multinomial count data. We develop a collapsed Gibbs sampling procedure that takes advantage
  of a series of augmentation relations, inspired by the Zhou--Cong--Chen model. We visualise the model's ability to identify coherent substructures in the data using a dataset of handwritten digits.  We then apply it to a large experimental dataset of DNA
  mutations in cancer and show that we can identify biologically meaningful clusters of mutational signatures in a fully data-driven way.
\end{abstract}

\section{Introduction}
\begin{figure}[!t]
  \begin{tabular}{cc}
  \subf{\includegraphics[width=0.5\textwidth]{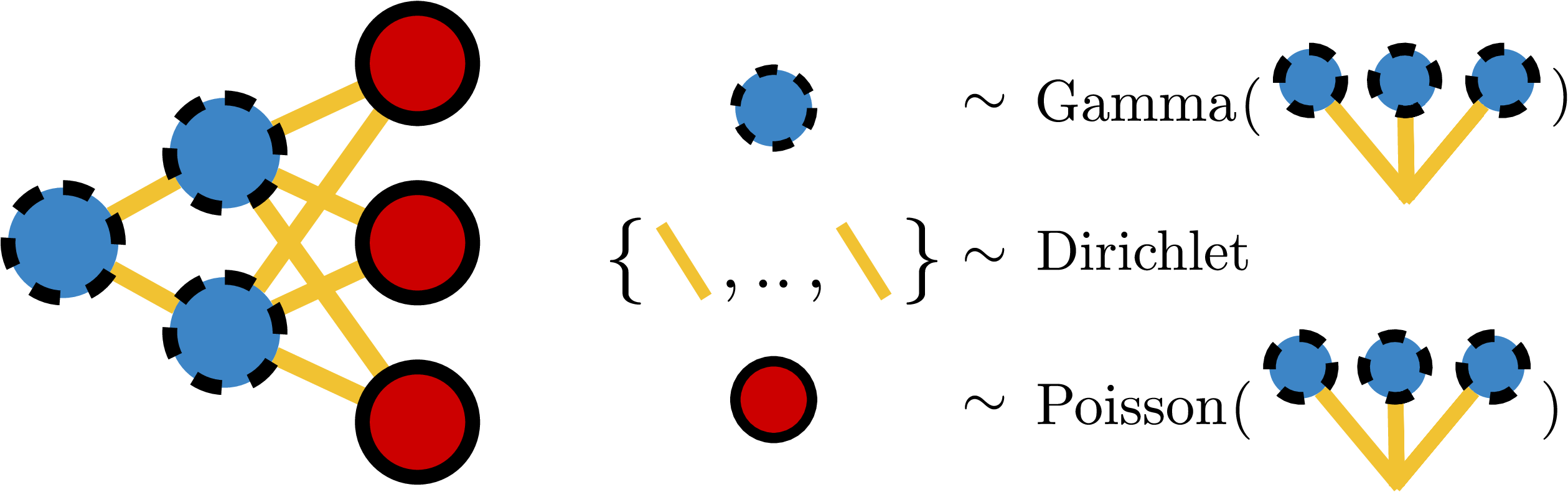}}{Poisson gamma belief network.}
  &
  \subf{\includegraphics[width=0.5\textwidth]{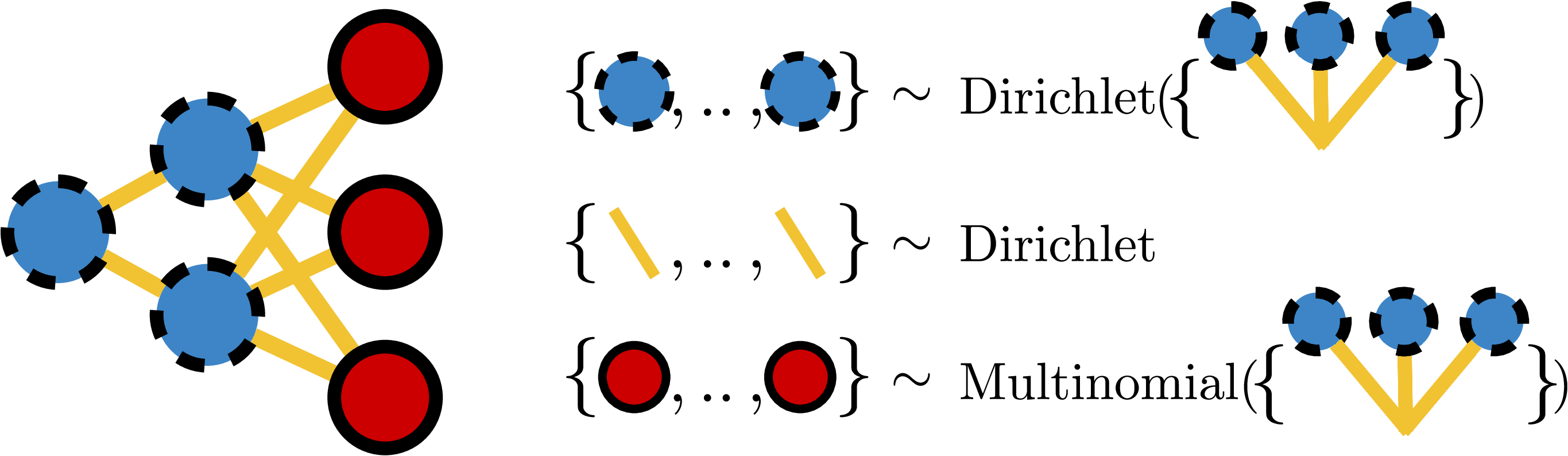}}{Multinomial belief network.}
    %\subf{0.45\textwidth}{\includegraphics[width=\textwidth]{poisson_gamma_belief_net}}
  \end{tabular}
  \caption{Schematic representation of the two belief networks. Red nodes are observations,  blue dashed circles are latent hidden units, and edges are latent weights. }
  \label{fig:architecture}
\end{figure}

Healthcare data is expensive and limited.  In addition, events of interest may be infrequent, and high missingness is a feature of most practical data sets. These features, along with the need to handle
uncertainty, present challenges to traditional maximum
likelihood-based machine learning methodologies, which often give rise to biased results, as highlighted by several
studies~\citep{smith87,beerli06,alzubaidi}. In addition, the predictions from these methodologies are often overconfident on
out-of-distribution data and fail to adequately address situations where
data incompleteness or uncertainty plays an important role~\citep{acharya2015gamma,emmanuel21,MURP23}.

As was argued persuasively in \cite{papamarkou24}, an approach that promises to overcome these limitations is to use generative and fully Bayesian methods. Examples include latent Dirichlet (LDA) allocation~\citep{blei2003latent},
deep belief nets \citep{HintonETAL:06,journals/jmlr/ZhouCC16}, and variational
autoencoders \citep{kingma2013autoencoding}. In principle, Bayesian methods are
data efficient, guard against overtraining,  account for
uncertainty, and deal with missing data in a principled way. However, current implementations
fall short of this ideal. One class of
methods uses variational approaches
to enable learning, which involves approximations such as fixing the posterior form
beforehand or by making a mean field assumption~\citep{RANG15,RANG16,FERR22,soleimani2017scalable,MURP23}. Other methods use exact
sampling but are
limited in their representational power by using binary variables, a shallow architecture, or use Poisson-distributed variables as (intermediate) output
\citep{WELL04,HintonETAL:06,journals/jmlr/ZhouCC16,zhao2018dirichlet,PANDA19,DONK21}.

In this paper, we tailor Bayesian belief networks to healthcare data by choosing a multinomial distribution for the output variable (Fig.~\ref{fig:architecture}). The versatility
of the multinomial distribution makes it well-suited for modelling data
types commonly found in healthcare data, from categorical variables found in patient questionnaires to text
documents~\citep{GRIF04} and DNA mutations in cancer. Using augmentation techniques real-valued, ordinal, and survival (i.e., censored) data can be modelled as well.
Importantly, it is straightforward to model several multinomials simultaneously (each
with their own dimension and observation count) which naturally enables modelling of
heterogeneous data, while missingness by setting relevant observation counts to 0. We will develop those extensions elsewhere; here we focus on
modelling a single multinomial-distributed observation.

A prominent example of a model with multinomial output variables is
LDA~\citep{blei2003latent}. In one representation of the model, the latent parameter
matrix of the multinomial distributions across samples is factorized as a low-rank
product of (sample-topic and topic-feature) matrices whose rows are drawn from Dirichlet
distributions. The low-rank structure, the sparsity induced by the Dirichlet priors, and
the existence of effective inference algorithms have resulted in numerous applications
and extensions of LDA. Despite this success, LDA has some limitations. One is that
inference of the Dirichlet hyperparameters is often ignored or implemented using
relatively slow maximum likelihood methods \citep{minka2003,george17}. This was
elegantly addressed by \citet{hdp06} by endowing the Dirichlet distribution with another
Dirichlet prior in a hierarchical structure, allowing information to be borrowed across
samples. Another limitation is that LDA ignores any correlation structure among the
topic weights across samples, which for higher latent (topic) dimensions becomes
increasingly informative.  An effective approach that addresses this issue was developed
by \citet{journals/jmlr/ZhouCC16}, who developed a multi-layer fully-connected Bayesian
network using gamma variates and Poisson, rather than multinomial, observables. Here we
combine and extend these two approaches in the context of multinomial
observations, resulting in a model whose structure resembles a fully connected
multi-layer neural network but retains the efficient inference properties of LDA.
This paper is structured as follows. In section \ref{sec:pgbn} we review
Zhou--Cong--Chen's Poisson-gamma belief network (PGBN) and subsequently introduce our
model in section \ref{sec:multinomial}. We apply the model to handwritten digits and mutations
in cancer in Sec.~\ref{sec:experiments}. We end with conclusions and a discussion
(Sec.~\ref{sec:dicussion_conclusion}).

\subsection*{Generalizable Insights about Machine Learning in the Context of Healthcare}
We introduce a Bayesian machine learning model, the multinomial belief network (MBN), to analyze healthcare data by decomposition. This unsupervised approach is well suited for analysing 'omics data such as gene expression and mutation profiles or to model the heterogeneity in clinical presentation, risk factors, and the underlying disease mechanisms of patient populations. Compared to e.g. non-negative matrix factorisation, this method is less prone to overtraining, and inferences come with uncertainty estimates. Unlike traditional topic modelling methods, MBNs can capture topic interactions across multiple layers. Initiatives like the 100k genomes project, which aim to integrate signature analysis into standard patient care~\citep{EVER23}, highlight the need for reliable deconvolution and uncertainty quantification to support treatment decisions.

\section{Poisson gamma belief network}\label{sec:pgbn}
  \subsection{Generative model}
  We first review the Gamma belief network of \citet{journals/jmlr/ZhouCC16}. The backbone of the model is a stack of Gamma-distributed hidden units
  $\theta^{(t)}_{vj}$ ($K_t$ per sample $j$), where the last unit parameterizes a
  Poisson distribution generating observed counts $x_{vj}$, one for each sample $j$ and
  feature $v$. The generative model is
  \begin{align}
      a_{vj}^{(T+1)} &= r_v \nonumber %\label{eq:thetaprior1},
       \\
      \theta^{(t)}_{vj} &\sim \Gam(a_{vj}^{(t+1)}, {c_j^{(t+1)}}), && t=T,\ldots,1
      \nonumber %\label{eq:thetaprior}
      \\
      a_{vj}^{(t)} &= \sum_{k=1}^{K_{t}} \phi^{(t)}_{vk}\theta^{(t)}_{kj}, && t=T,\ldots,1 \nonumber % %\label{eq:updatea}
      \\
      x_{vj} &\sim \Pois( a_{vj}^{(1)} ).
      \nonumber %\label{eq:poislikel}
  \end{align}
  For $T=1$ we only have one layer, and the model reduces to Poisson Factor Analysis,
  $x_{vj} = \Pois([\pmb{\phi\theta}]_{vj})$ \citep{zhou2012beta}. For multiple layers,
  the features $\pmb{\theta}^{(t+1)}$ on layer $t+1$ determine the shape parameters of
  the gamma distributions on layer $t$ through a non-negative connection weight matrix
  $\pmb{\phi}^{(t+1)}\in\mR_+^{K_{t}\times K_{t+1}}$, so that $\pmb{\phi}^{(t+1)}$
  induces correlations between features on level $t$. The rate parameter for $\pmb{\theta}^{(t)}$
  is $c_j^{(t)}\sim \Gam(e_0,f_0)$, one for each sample $j$ and layer $t$;  for  $t>1$ the $c^{(t)}_j$ act as concentration parameters for the activations below.
  %with hyperparameters $e_0$ and $f_0$.
  The weights $\pmb{\phi}^{(t)}$ that connect latent states between layers are normalised as $\sum_v \phi^{(t)}_{vk}=1$ owing to their Dirichlet priors $\phi_{vk}^{(t)} \sim \Dir(\{ \eta_v^{(t)}\}_v)$%with hyperparameters $\{ \eta_v^{(t)}\}_v$
  ; here we use curly braces to denote vectors, with the subscript indicating the index
  variable; we drop the subscript if the index variable is unambiguous. The top-level
  activation is controlled by $r_v\sim\Gam(\gamma_0/K_T,c_0)$ where hyperparameters
  $\gamma_0$ and $c_0$  determine the typical number and scale, respectively, of active
  top-level hidden units. The lowest-level activations $\pmb{a}^{(1)}$ parameterize
  Poisson distributions that generate the observed count variables $x_{vj}$ for sample
  (individual, observation) $j$. This completes the specification of the generative
  model.

  This model architecture is similar to a $T$-layer neural network, with activations $\pmb{a}^{(t)}$
   representing the activity of features (topics,
  factors) of increasing complexity as $t$ increases.

  \def\nk{\kern-2pt}

  \subsection{Deep Poisson representation}
\begin{figure*}
      \centering
      %\figuresize{2.0}
      \begin{tikzpicture}[node distance=1.5cm,>=latex]
          % Nodes for a^3, theta^2, a^2, theta^1, a^1, and phi^1
          \node (a3) {$\pmb{a}^{(3)}$};
          \node (theta2) [below of=a3] {$\pmb{\theta}^{(2)}$};
          \draw [->] (a3) -- (theta2);
          \node (a2) [below of=theta2] {$\pmb{a}^{(2)}$};
          \draw [-|] (theta2) -- (a2);
          \node (theta1) [below of=a2] {$\pmb{\theta}^{(1)}$};
          \draw [->] (a2) -- (theta1);
          \node (a1) [below of=theta1] {$\pmb{a}^{(1)}$};
          \draw [-|] (theta1) -- (a1);
          \node (phi1) [left of=a1] {$\pmb{\phi}^{(1)}$};
          \draw [-|] (phi1) -- (a1);

          % Node for c^3 and arrow to theta^2
          \node (c3) [left of=theta2] {$\pmb{c}^{(3)}$};
          \draw [->] (c3) -- (theta2);

          % Node for c^2 and arrow to theta^1
          \node (c2) [left of=theta1] {$\pmb{c}^{(2)}$};
          \draw [->] (c2) -- (theta1);

          % Nodes for x^i, q^i, and arrows to them
          \node (xi3) [text=gray,right of=a3] {$\pmb{x}^{(3)}$};
          \node (q3) [text=gray,right of=xi3] {$\pmb{q}^{(3)}$};
          \node (xi2) [text=gray,right of=a2] {$\pmb{x}^{(2)}$};
          \node (q2) [text=gray,right of=xi2] {$\pmb{q}^{(2)}$};
          \node (xi1) [right of=a1] {$\pmb{x}^{(1)}$};
          \node (q1) [right of=xi1] {$\pmb{q}^{(1)}$};
          \draw [->,gray] (a3) -- (xi3);
          \draw [->,gray] (a2) -- (xi2);
          \draw [->] (a1) -- (xi1);
          \draw [-|,gray] (q1) -- (q2);
          \draw [-|,gray] (c2) -- (q2);
          \draw [-|,gray] (q2) -- (q3);
          \draw [-|,gray] (c3) -- (q3);

          \draw [->,gray] (q3) -- (xi3);
          \draw [->,gray] (q2) -- (xi2);
          \draw [->] (q1) -- (xi1);

          % Node for phi^2 and arrow to a^2
          \node (phi2) [left of=a2] {$\pmb{\phi}^{(2)}$};
          \draw [-|] (phi2) -- (a2);

          % Node for phi^3 and arrow to a^3
          \node (phi3) [left of=a3] {};

          \node (labelA) [left of=phi3] {\bf a.};
          \node (labelB) [right of=q3] {\bf b.};
          \node (cempty) [right of=labelB] {};
          \node (ca3) [right=0.25 cm of cempty] {$\pmb{a}^{(3)}$};
          \node (cx3) [right of=ca3] {$\pmb{x}^{(3)}$};
          \node (cq3) [right of=cx3] {$\pmb{q}^{(3)}$};
          \node (cm2) [below=0.42 cm of cx3] {$\pmb{m}^{(2)}$};
          \node (cy2) [below=0.42 cm of cm2] {$\pmb{y}^{(2)}$};
          \node (cx2) [below=0.42 cm of cy2] {$\pmb{x}^{(2)}$};
          \node (cm1) [below=0.42 cm of cx2] {$\pmb{m}^{(1)}$};
          \node (cy1) [below=0.42 cm of cm1] {$\pmb{y}^{(1)}$};
          \node (cx1) [below=0.42 cm of cy1] {$\pmb{x}^{(1)}$};
          \node (cc3) at  (cm2 -| cempty) {$\pmb{c}^{(3)}$};
          \node (cc2) at  (cm1 -| cempty) {$\pmb{c}^{(2)}$};
          \node (cphi2) at  (cy2 -| cempty) {$\pmb{\phi}^{(2)}$};
          \node (cphi1) at  (cy1 -| cempty) {$\pmb{\phi}^{(1)}$};
          \node (cq2) at  (q2 -| cq3) {$\pmb{q}^{(2)}$};
          \node (cq1) at  (q1 -| cq3) {$\pmb{q}^{(1)}$};
          \draw [->] (ca3) -- (cx3);
          \draw [->] (cq3) -- (cx3);
          \draw [->] (cx3) -- (cm2);
          \draw [->] (cm2) -- (cy2);
          \draw [-|] (cy2) -- (cx2);
          \draw [->] (cx2) -- (cm1);
          \draw [->] (cm1) -- (cy1);
          \draw [-|] (cy1) -- (cx1);
          \draw [-|] (cc3) -- (cq3);
          \draw [-|] (cc2) -- (cq2);
          \draw [->] (cphi2) -- (cy2);
          \draw [->] (cphi1) -- (cy1);
          \draw [-|] (cq1) -- (cq2);
          \draw [-|] (cq2) -- (cq3);
          \draw [->] (cq3) -- (cm2);
          \draw [->] (cq2) -- (cm1);
      \end{tikzpicture}
      \caption{Two equivalent generative models for a count variable $\pmb{x}^{(1)}$ from the Poisson gamma belief network, using ({\bf a}) a tower of real-valued latent variables $\pmb{\theta}$, $\pmb{a}$,  or ({\bf b}) latent counts $\pmb{m}$, $\pmb{y}$, $\pmb{x}$.  Blunt arrows indicate deterministic relationships.  The variable $\pmb{q}^{(1)}$ is a dummy and has a fixed value $1$.  In representation ({\bf a}) the grayed-out counts $\pmb{x}^{(t)}$ and variables $\pmb{q}^{(t)}$, $t>1$,  are included for clarity (and have the same distribution as the variables in the right model) but are not used to generate the outcome $\pmb{x}^{(1)}$, and so can be marginalized out.}
      \label{fig:equiv}
  \end{figure*}
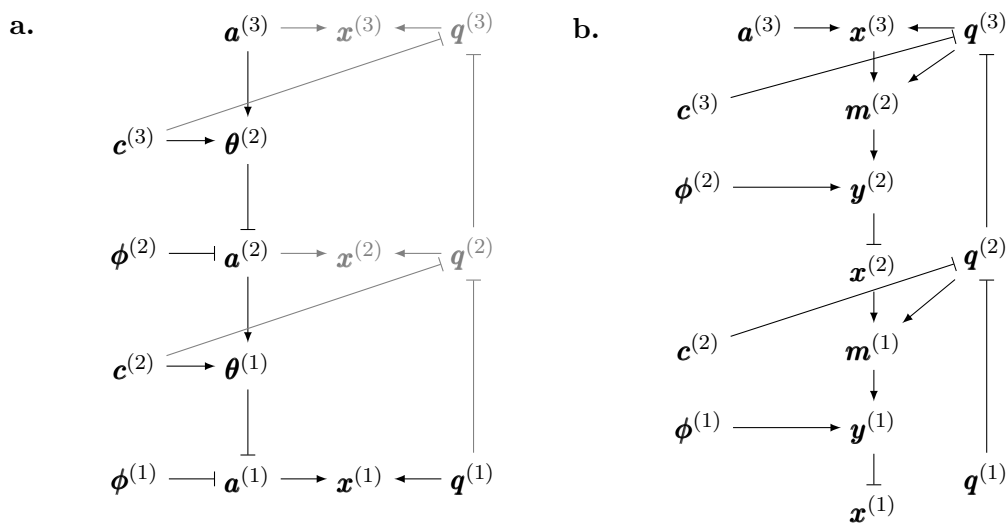

  An alternative and equivalent representation is obtained by integrating out the hidden
  units $\pmb{\theta}$ and augmenting with a sequence of latent counts $\pmb{x}^{(t+1)}
  \nk\rightarrow\nk \pmb{m}^{(t)} \nk\rightarrow \nk\pmb{y}^{(t)}\nk \rightarrow\nk
  \pmb{x}^{(t)}$. Specifically, let $\Log(p)$ be the logarithmic distribution, with
  probability mass function $\Log(k;p) \propto {p^k/ k}$ where $0\nk<\nk p\nk<\nk 1$,
  and define $n\sim\SumLog(l,p)$ by $n = \sum_{i=1}^l u_i$ where each $u_i \sim
  \Log(p)$. Henceforth, underlined indices denote summation, so that $x_\marg{j} := \sum_j x_j$.
  Augmenting each layer with counts
  %Starting from the top-most $a^{(T+1)}_{kj}$, augment
  $x_{kj}^{(t)}\sim\Pois(q_j^{(t)}a_{kj}^{(t)})$, it turns out these can be generated as
  \begin{align*}
  &m_{jk}^{(t)}\sim\SumLog(x_{kj}^{(t+1)},1-e^{-q_j^{(t+1)}});\qquad\\
  &\{y_{vjk}^{(t)}\}_v\sim\Mult(m_{jk}^{(t)},\{\phi_{vk}^{(t)}\}_v);\qquad\\
  &x_{vj}^{(t)}:=y_{vj\marg{k}}^{(t)},
  \end{align*}
  where     $q_j^{(t+1)} = \ln
      [1+{q_j^{(t)}}/ c_j^{(t+1)}]$; see \citet{journals/jmlr/ZhouCC16} and Supplementary Material sections~\ref{sec:dpfa}, \ref{sec:altdpfa}.
  Starting with $t=T+1$ this shows how to eventually generate the observed counts $x^{(1)}_{vj}$ using only count variables, with all $\pmb{\theta}^{(t)}$ integrated out.
  The two alternative schemes are shown graphically in Fig.~\ref{fig:equiv}.

  \subsection{Inference}
  At a high level, inference consists of repeatedly moving from the first representation
  to the second, and back.  This is achieved by swapping, layer-by-layer, the direction
  of the arrows to sample upward $\pmb{x}^{(t)} \rightarrow \pmb{y}^{(t)} \rightarrow
  \pmb{m}^{(t)} \rightarrow \pmb{x}^{(t+1)}$; these counts are then used to sample
  $\pmb{\phi}^{(t)}$ and $\pmb{\theta}^{(t)}$, after which the procedure starts again.
  To propagate latent counts upwards, we use the identity from Theorem 1 of~\citet{journals/pami/ZhouC15}:
  \begin{equation}
      \begin{aligned}
%         p(x, m|a, q) = & 
\Pois(x|qa) \SumLog(m|x,1-e^{-q}) %\\
         = %& 
         \NB(m|a,1-e^{-q}) \CRT(x|m,a),
      \end{aligned}
      \label{eq:factorpoisson}
  \end{equation}
  to turn $ \pmb{m} \rightarrow \pmb{x} $ into $\pmb{x} \rightarrow \pmb{m}$; here
  $\CRT(x|m,a)$ is the number of occupied tables in a Chinese restaurant table
  distribution over $m$ customers with concentration parameter $a$, and $\NB(k|r,p)$ is
  the negative-binomial distribution with $r$ successes of probability $p$. Finally, we
  use that independent Poisson variates conditioned on their sum are multinomially
  distributed with probabilities proportional to the individual Poisson rates, to
  convert $\pmb{y} \rightarrow \pmb{x}$ to $ \pmb{x} \rightarrow \pmb{y}$ as well as
  $\pmb{m} \rightarrow \pmb{y}$ to $ \pmb{y} \rightarrow \pmb{m}$. To Gibbs sample the
  variables, we first make an upward pass from $\pmb{x}^{(1)} \rightarrow \dots
  \rightarrow \pmb{x}^{(T+1)}$ followed by a downward pass where the
  multinomial-Dirichlet conjugacy is used to update $\pmb{\phi}^{(t)}$, the gamma-gamma
  rate conjugacy to update $\pmb{c}^{(t)}$ and the Poisson-gamma conjugacy to update
  $\pmb{\theta}^{(t)}$ and $\pmb{r}$. Details are provided in the supplement.

  \section{Multinomial belief network}
  \label{sec:multinomial}

  \subsection{Generative model}
  We now introduce the multinomial belief network (MBN). To model multinomial observations, we replace Poisson observables with multinomials,  and we replace the gamma-distributed hidden activations with Dirichlet-distributed weights $\{ \theta^{(t)}_{vj} \}_v$.
  The generative model is
  \begin{align}
      a_{vj}^{(T+1)} &= r_v, \\
      \{\theta^{(t)}_{vj}\}_v &\sim \Dir(\{c^{(t+1)} a_{vj}^{(t+1)}\}_v), && t=T,\ldots,1 \\ %\label{eq:dmfatheta} \\
  %    c^{(t)} & \sim  \Gam(e_0, f_0), && t=T,\ldots,2\\
      a_{vj}^{(t)} &= \sum_{k=1}^{K_{t}} \phi^{(t)}_{vk}\theta^{(t)}_{kj}, && t=T,\ldots,1 \label{eq:dmfaa} \\
      \{x_{vj}\}_v &\sim \Mult( n_j, \{a_{vj}^{(1)}\}_v ).
      \label{eq:dmfamult}
  \end{align}
  As before, the weights are Dirichlet distributed $\phi_{vk}^{(t)} \sim \Dir(\{ \eta_v^{(t)}\}_v)$ with hyperparameters $\{ \eta_v^{(t)}\}_v$.
  Different from the PGBN model we choose one $c^{(t)} \sim \Gam(e_0, f_0)$ per dataset (and per layer) instead of one per sample $j$, reducing the number of free parameters per sample, and allowing the variance across samples to inform the $c^{(t)}$.
  Finally we let the top-level activations $r_v$ be Dirichlet distributed, $\{r_v\}_v \sim \Dir(\{ \gamma_0 / K_T \}_v)$ with $\gamma_0$ a hyperparameter. This completes the definition of the generative model.

  \subsection{Deep multinomial representation}
  Integrating out $\pmb{\theta}^{(t)}$, the generative model can be alternatively represented as a deep multinomial factor model, as follows (Supplementary Material, Sec.~\ref{supplementary_sec:multinomial_factorisation},):
  \begin{equation}
      \begin{aligned}
          n_j^{(t+1)}&\sim\CRT(n_j^{(t)},c^{(t+1)});\\
          \{x_{kj}^{(T+1)}\}_k & \sim\Mult(n_j^{(T+1)}, %\{a^{(T+1)}_{kj}\}_k);\\
          \{r_k\});\\
          \{m^{(t)}_{kj}\}_k & \sim \Polya(n_j^{(t)}, \{x_{kj}^{(t+1)}\}_k);\\
          \{y^{(t)}_{vjk}\}_v & \sim\Mult(m_{jk}^{(t)},\{\phi^{(t)}_{vk}\}_v);\\
          x^{(t)}_{vj} & = y^{(t)}_{vj\marg{k}}.
      \end{aligned}
  \end{equation}
  Here underlined subscripts denote summation; and $\Polya(n, \{y_k\})$ is the distribution of the contents of an urn after running a Polya scheme: starting with $y_k$ balls of color $k$, repeatedly drawing a ball, returning the drawn ball and a new identically colored one each time, until the urn contains $n$ balls. The two representations of the model are structurally identical to the two representations of the PGBN shown in Fig.~\ref{fig:equiv}, except that the $q^{(t)}$ are replaced by $n^{(t)}$, and the relationship between successive $n^{(t)}$ is stochastic instead of deterministic.

  \subsection{Inference}
  Similar to the PGBN, we reverse the direction of $\pmb{x}^{(t+1)} \rightarrow \pmb{m}^{(t)} \rightarrow \pmb{y}^{(t)} \rightarrow \pmb{x}^{(t)}$ to propagate information upward. To reverse $\pmb{x}^{(t+1)} \rightarrow \pmb{m}^{(t)}$ into $\pmb{m}^{(t)} \rightarrow \pmb{x}^{(t+1)}$ we use the following:
  \begin{theorem}
      \label{theorem1}
      The joint distributions over $n$, $\{x_k\}$ and $\{m_k\}$ below are identical:
      \begin{align}
          %& p(n,\{x_k\},\{m_k\}|n_0,c,\{a_k\})= \\
           &\DirMult(\{m_k\}|n_{0}, \{ca_k\}) \left[\prod_{k} \CRT(x_k|m_k, ca_k) \right] \delta_{n, x_{\marg{k}}}= \nonumber \\
          & \CRT(n|n_{0}, c) \Mult(\{x_k\}|n,\{a_k\}) \Polya(\{m_k\}|n_{0}, \{x_k\}).
      \nonumber
      \end{align}
  \end{theorem}
  (For the proof see Sec.\ref{sec:proof}, Supplementary Material.)
  Here $\DirMult(\{x_k\}|n,\{c a_k\})$ is the Dirichlet-multinomial distribution of $n$ draws with concentration parameters $\{c a_k\}$, and $\delta_{i,j}$ denotes the Kronecker delta function that is $1$ when $i=j$ and zero otherwise; here it expresses that $n=x_{\underline{k}}$.  Note that $a_{\underline{k}}=1$, as it parameterizes a multinomial.

  In words, the theorem states that observing $\{m_k\}$ from a multinomial parameterized by probabilities from a Dirichlet distribution that itself has parameters $\{c \alpha_k\}$, provides information about the probabilities $\{a_k\}$ through an (augmented) multinomial-distributed observation $\{x_k\}$, and information about the concentration parameter $c$ through a CRT-distributed observation $n=x_{\underline{k}}$.
  Therefore, by augmenting  with $\{x_k\}$ and
  %choosing priors for $\{\alpha_k\}$ and $c$ that are conjugate to the multinomial and $\CRT$ distributions respectively, posteriors for $\{a_k\}$ and $c$ can be obtained.
  choosing conjugate priors to the multinomial and $\CRT$ distributions for $\{a_k\}$ and $c$ respectively, we can obtain posteriors for these parameters.

  The remaining arrows can be swapped by augmenting and marginalizing multinomial distributions.
  Taken together, to sample in the MBN we use augmentation and the Dirichlet-multinomial conjugacy to update $\pmb{\phi}^{(t)}$, $\pmb{\theta}^{(t)}$, and $\pmb{r}$, while to update $c^{(t)}$ we sample from the Chinese restaurant table conjugate prior $\GCRTP(\alpha | m, \{n_j\}_j, a, b) \propto \Gam(\alpha|a, b) \alpha^m \prod_j \frac{\Gamma(\alpha)}{\Gamma(\alpha + n_j)}$ using the method described by~\citet{hdp06}.  In more detail, sampling proceeds as follows.  Identifying $x_{vj}^{(1)} \equiv x_{vj}$,    $n_j^{(1)} \equiv x_{\marg{v}j}$  and $a^{(T+1)}_{kj} \equiv r_v$:
  \begin{algorithm}[H]
      \caption{Procedure to Gibbs sample weights $\pmb{\phi}^{(t)}$, hidden units $\pmb{\theta}^{(t)}$, concentration parameters $c^{(t)}$, and activations $\pmb{r}$ of an MBN given training data $x_{vj}^{(1)}$.}
      \begin{tabbing}
          \qquad \enspace For $t=1,\ldots,T$: \\
             \qquad \qquad Sample $\{y_{vjk}^{(t)}\}_{k} \sim
              \Mult(x^{(t)}_{vj}, \{\phi^{(t)}_{vk}\theta^{(t)}_{kj}\}_{k})$\\
              \qquad \qquad Compute $m_{jk}^{(t)} = y_{\marg{v}jk}^{(t)}$, \\
              \qquad \qquad Sample $x_{kj}^{(t+1)} \sim \CRT(m_{jk}^{(t)}, c^{(t+1)} a_{kj}^{(t+1)})$;\\
              \qquad \qquad Compute $n_j^{(t+1)} = x_{\marg{v}j}^{(t+1)}$,  \\
              \qquad \qquad Sample $\{\phi^{(t)}_{vk}\}_v \sim \Dir(\{\eta^{(t)}_v
          +y^{(t)}_{v\marg{j}k}\}_v)$, \\
          \qquad \enspace Sample $\{r_v\} \sim\Dir(\{\gamma_0/K_T+x_{v\marg{j}}^{(T+1)}\}_v)$ \\
          \qquad \enspace For $t=T,\ldots,1$: \\
              \qquad \qquad Sample $\{\theta_{kj}^{(t)}\}_k \sim \Dir(\{c^{(t+1)}a_{kj}^{(t+1)}+m_{jk}^{(t)}\}_k)$ \\
              \qquad \qquad Sample $c^{(t+1)} \kern-2pt \sim\kern-1pt  \GCRTP (n^{(t+1)}_{\marg{j}}\kern-2pt , \{n^{(t)}_{j}\}_j, \kern-1pt e_0, \kern-1pt f_0)$ 

      \end{tabbing}
  \end{algorithm}
  In practice, sampling might proceed per observation $j$, and resampling of global
  parameters $\pmb{\phi}^{(t)}$ and $c^{(t)}$, which involve summing over $j$, is done
  once all observations have been processed. For details see Supplement section
  \ref{supplementary_sec:multinomial}.

  \section{Experiments}\label{sec:experiments}
  \subsection{Performance evaluation}
  Having reviewed the PGBN and introduced our model, we now illustrate its application on small images of handwritten digits and on DNA point mutations in cancer. To evaluate performance, we hold out 50\% of the pixel quanta (resp.\ mutations)  from the images (resp.\ patients) to form a test set, $\mathbf{x}^{\mathrm{test}}$, and evaluate the held-out perplexity as:
  \begin{align}\label{eq:perplexity}
      \mathcal{L}(\mathbf{x}^{\mathrm{test}}) = \exp\left(-\frac{1}{J}\sum_{j=1}^J  \sum_{v=1}^V \frac{x_{vj}^{\mathrm{test}}
      \ln p_{vj}}{x_{\marg{v}j}^{\mathrm{test}}}\right),
  \end{align}
  where $p_{vj}$ is the probability of feature $v$ in example $j$. For non-negative
  matrix factorisation (NMF) trained on a Kullback-Leibler (KL) loss (which is
  equivalent to a Poisson likelihood,~\cite{LEE99}), the probability
  $p_{vj}=a_{vj}/a_{\marg{v}j}$ is the training set reconstruction $a_{vj} \equiv \sum_k
  \phi_{vk} \theta_{kj}$ (so that $a_{vj} \approx x_{vj}$) normalised across features
  $v$.

  Similarly, $p_{vj}=S^{-1} \sum_{\sigma=1}^{S}
  a^{(\sigma)}_{vj}/a^{(\sigma)}_{\marg{v}j}$ for the PGBN and MBN where $a_{vj} =
  \sum_{k=1}^{K_1}\phi^{(1)}_{vk} \theta^{(1)}_{kj}$ is the bottom layer activation
  normalised and averaged over $S$ posterior samples $\sigma=1,\dots,S$ (for the
  MBN $a_{vj}$ is normalised by construction) similar
  to~\citet{zhou2012beta,RANG15}.

  Unlike in~\citet{journals/jmlr/ZhouCC16}, for the PGBN we use a gamma distribution to
  model the scale $c_j \sim \Gam(e_0, f_0)$ for all layers, instead of a separate
  beta-distributed $p^{(2)}_j$ to set $c_j^{(2)}=p^{(2)}_j/(1-p^{(2)}_j)$ for the first
  layer only. In addition, we consider $\gamma_0$ a fixed hyperparameter for both belief
  networks. For each experiment, four Markov chains were initialised using the prior to
  ensure overdispersion relative to the posterior (as suggested in~\citet{GELM13}). Each
  chain was run in parallel on a separate nVidia A40 device.

  \subsection{UCI ML handwritten digits}
  \begin{figure}
      \centering
      \begin{tabular}{cc}
          \subf{\includegraphics[width=0.5\textwidth]{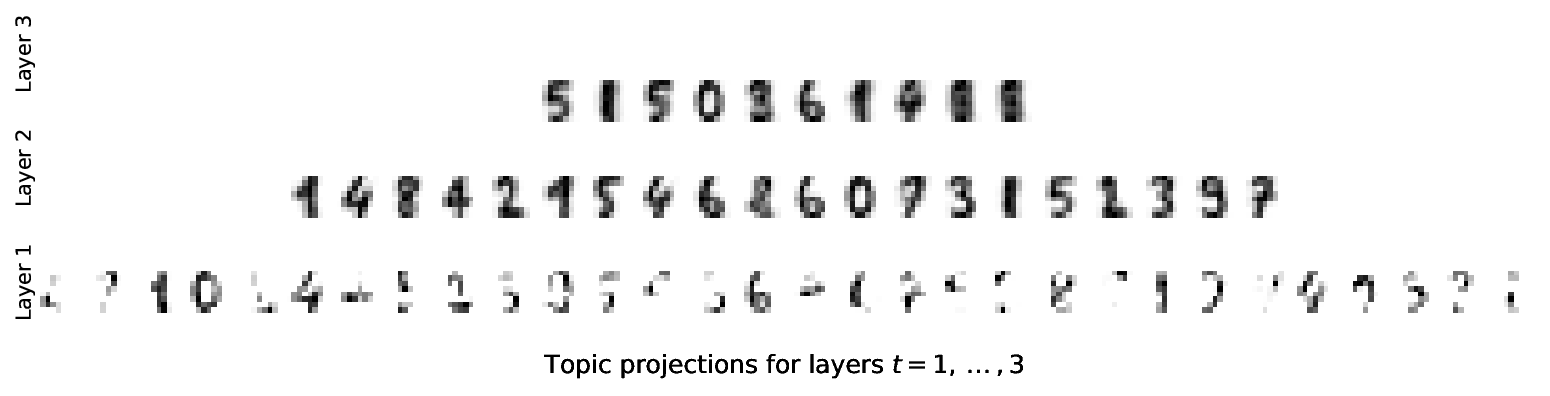}}{Chain 1}
          &
          \subf{\includegraphics[width=0.5\textwidth]{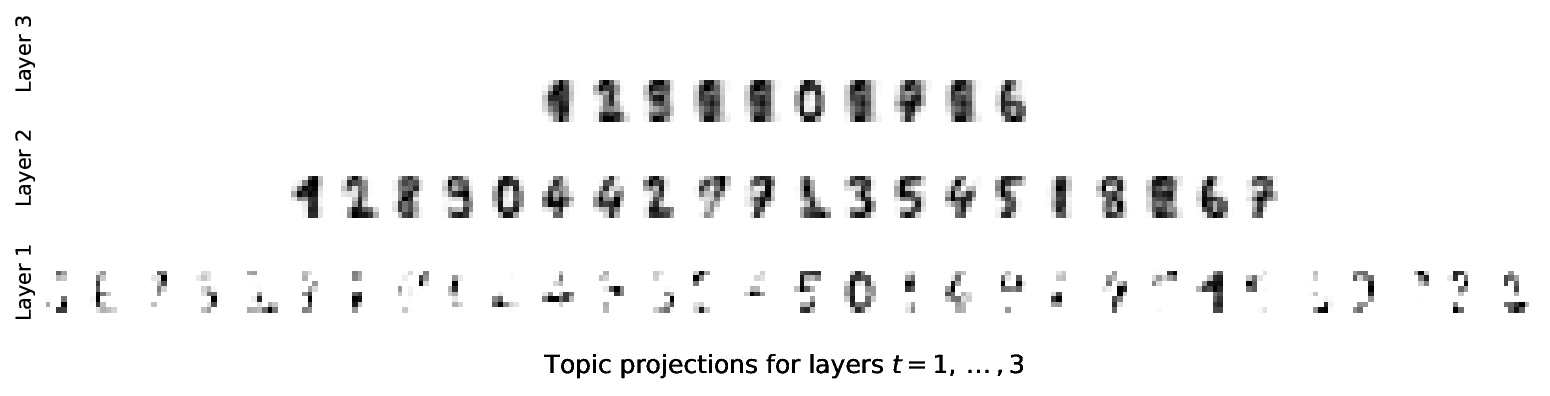}}{Chain 2}
          \\
          \subf{\includegraphics[width=0.5\textwidth]{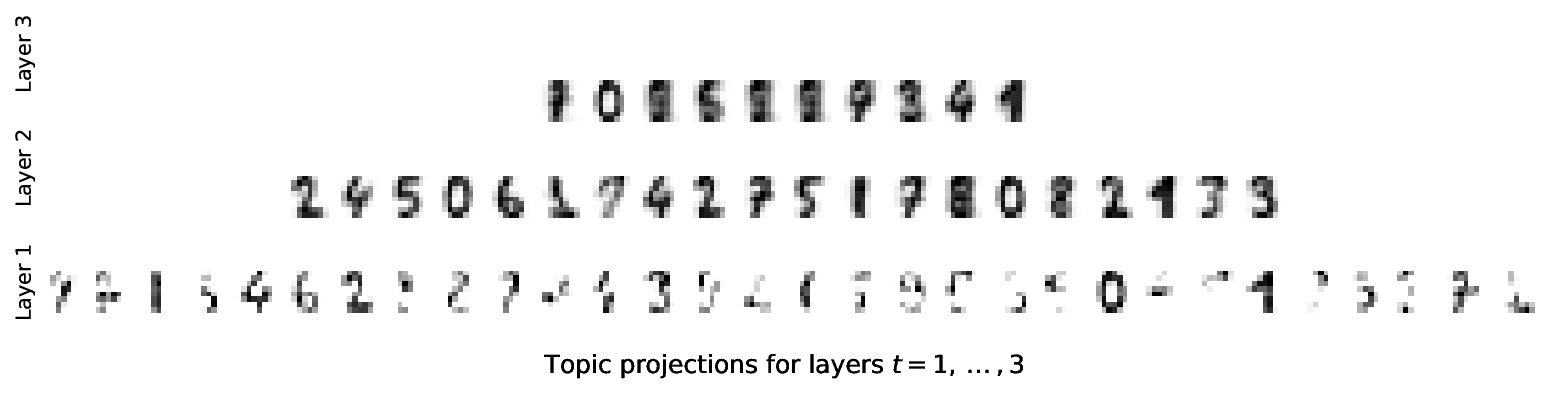}}{Chain 3}
          &
          \subf{\includegraphics[width=0.5\textwidth]{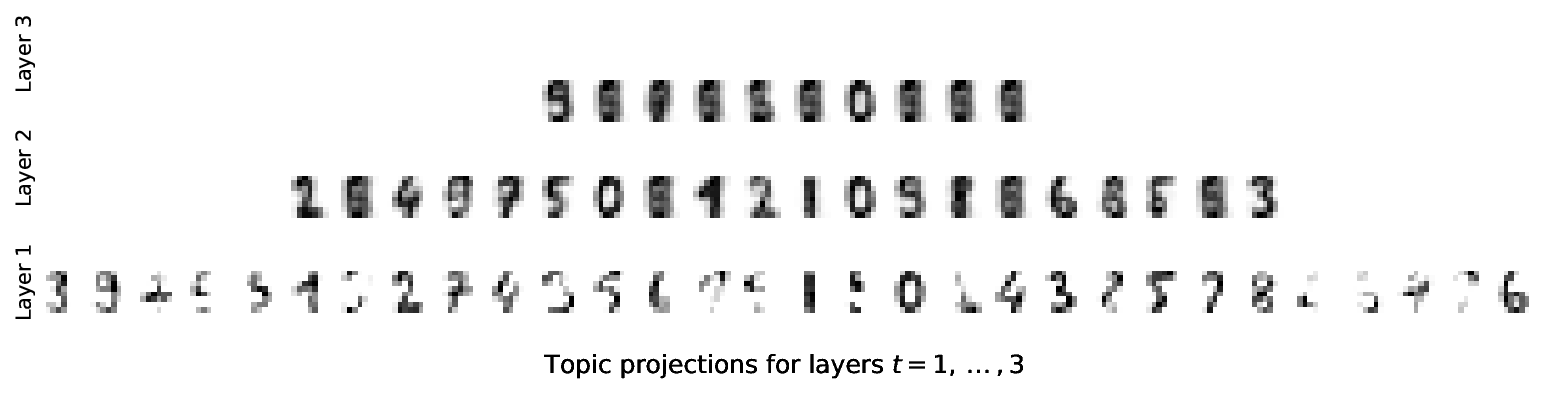}}{Chain 4}
      \end{tabular}
      \caption{Hierarchy of topics learned by a three-layer MBN (with $[K_1, K_2, K_3] = [30, 20, 10]$ latent components) after training on the Optical Recognition of Handwritten Digits dataset from the UC Irvine Machine Learning Repository~\citep{ALPA98}. Topics are represented by their projection $\prod_{l=1}^{t}\pmb{\phi}^{(l)}$ onto the pixels. Separate panels refer to individual Markov chains that were run in parallel.}
      \label{fig:digits}
  \end{figure}

  We considered the Optical Recognition of Handwritten Digits dataset from the UC Irvine Machine Learning Repository~\citep{ALPA98} 
  %(part of Sci-kit learn~\citep{SKL11}) 
  containing 1797 images of handwritten digits.
  %numbered zero to nine. 
  Each pixel in the $8\times 8$ images had a discrete intensity
  ranging from 0 to 15 %(i.e., four bits) 
  which we modelled as counts. Three separate
  models ($[K_1, K_2, K_3] = [30, 20, 10]$
  latent components; $\gamma_0 =e_0 = f_0 = 1$, $\eta=0.05$) with one through three layers were run
  for $10^5$ Gibbs burn-in steps and 1280 samples were collected from each chain. 
  In terms of likelihood, all three MBN models converged within 5000 iterations of
  burn-in. 
  %Nevertheless, convergence metrics such as $\hat{R}$~\citep{GELM13}, comparing
  %between- and within-chain estimates, indicated that the sampler did not fully explore
  %the posterior (not counting the model's intrinsic symmetries), even after $10^5$
  %burn-in steps. We attributed this to the large number of examples in the dataset.
  %Indeed, running the sampler on a much smaller dataset with only 10 examples leads to
  %satisfactory convergence ($\hat{R} \leq 1.01$ for all latent states on all network
  %configurations after being made permutation symmetry invariant). Furthermore,
  Extensive simulation-based calibration tests~\citep{TALT18} for various network
  configurations on small datasets (we tested up to $J=19$ examples) indicate that our
  sampler was correctly implemented. 
  %The poor mixing is presumably an intrinsic property
  %of our Gibbs sampling approach.

    The perplexity on holdout pixel intensity quanta were
  31.0$^{+0.1}_{-0.1}$, 30.7$^{+0.1}_{-0.1}$, and 30.7$^{+0.1}_{-0.1}$ for one to three
  layers (lower is better, bootstrapped 95\% confidence intervals), respectively. The
  relatively modest improvement with depth is typical for these models and also
  observed e.g.\ in the PGBN on 20 newsgroup data~\citep{journals/jmlr/ZhouCC16} %and below on mutations. 
  For comparison, we trained NMF (from Sci-kit learn~\citep{SKL11})
  using a KL loss and achieved a significantly larger (worse) perplexity of
  34.2$^{+0.3}_{-0.3}$~\footnote{\label{f1} Technically, the perplexity of NMF was infinite because
  zero probability was assigned to non-zero intensity. Samples where NMF incorrectly
  attributed zero probability were removed from the perplexity calculation.}.
  Importantly, the digits highlight the interpretability of the network, which
  hierarchically learns topics from the specific to the general (Fig.~\ref{fig:digits}).
  The sum-of-parts representations enforced by the network cause the lowest layer in the
  network to learn digit patches. These are then combined in higher layers to form
  increasingly general digit representations (Fig.~\ref{fig:digits}).

  \subsection{Mutational signature attribution}

  \begin{table}
  \centering
    \caption{Perplexity of held-out mutations for inferred mutational signature attributions. \rm Signatures were based on COSMIC v3.3 signature weights. Super/subscripts indicate ninety-five per cent confidence intervals, computed by bootstrapping.}
    {
    \begin{tabular}{lll}
      \toprule
      \textbf{Method}     & & \textbf{Hold out perplexity} \\
      \midrule
      SigProfilerExtractor$^{\ref{f1}}$ & & 64.5$^{+0.7}_{-0.7}$     \\
      Zhou--Cong--Chen & (1 layer)     & 62.0$^{+0.7}_{-0.7}$      \\
      & (2 layers)     & 61.9$^{+0.7}_{-0.7}$      \\
      This work & (1 layer) & 62.0$^{+0.7}_{-0.7}$  \\
      & (2 layers) & 61.9$^{+0.7}_{-0.7}$  \\
      \bottomrule \\
    \end{tabular}
    \\}
    \label{tab:mutations}
      {\raggedright  %\\
      %* The perplexity of SigProfilerExtractor was technically infinite due to zero
      %probability being assigned to observed mutations. These samples were removed from
      %the calculations.\par
      }
  \end{table}

  Next, we turned to mutational signatures imprinted in the DNA of cancer cells. Mutational signatures arise from observed patterns of DNA mutations and are not necessarily uniformly randomly distributed across the genome, but influenced by a range of factors including mutagenesis, DNA damage sensing, repair pathways and chromatin context~\citep{SING20}. To compute mutational signatures, observed DNA mutation patterns are typically first summarized into a 96-feature vector of counts $\bm{x}$ of different types of mutations~\cite{ALEX13}, and then approximately factorized into ``mutational signatures'' $\bm{\phi}$ and sample-specific attributions of mutations to each signature $\bm{\theta}$. The signatures $\phi_{vk}^\mathrm{COSM}$ reported in the COSMIC database~\citep{TATE19, COSM23} are the \textit{de facto} standard in the field, relating mutation spectra across $v=1,\dots,96$ possible point mutations types to $k=1,\ldots,78$ signatures named SBS1$,\dots,$SBS94. Attributions based on this curated set of signatures are increasingly used to guide therapeutic decisions in cancer~\citep{BRAD22,PATT23}, underscoring the need for accurate attribution and quantification of uncertainty.

  In short, our goal was to infer for each patient the proportion of mutations,
  $\bm{\theta}$, corresponding to specific signatures ($\bm{\phi}^\mathrm{COSM}$, COSMIC
  v3.3) given their mutation profiles. We tested our model on the mutation dataset
  of~\citet{ALEX20} comprising $\sim\kern-3pt 85$ million mutations from 4,645 patients and
  compared with SigProfilerExtractor, considered the state of art for de novo extraction of mutational signatures \citep{ISL22}, and the Zhou--Cong--Chen
  model. Since we expected around 5--10 signatures to be present per sample, we
  set $\gamma_0=10$ and other hyperparameters $\eta = e_0 = f_0 = 1$ for both the
  Zhou--Cong--Chen and our model and used the greedy layer-wise training procedure
  (Sec.~\ref{supplementary_sec:training_mutational_signatures}, Supplementary Material).
  Although the test-set likelihood indicated that the chains of both models had not yet
  fully converged, we halted computation due to the large computation time (a total of
  77 and 78 GPU days for MBN and PGBN, respectively). Since the chains were initialised
  with overdispersed values (compared to the posterior), pre-mature termination of the
  Markov chains overestimates between-chain variance compared to the ``true'' posterior.
  That is, our uncertainty estimates are conservative. Nevertheless, both the
  Zhou--Cong--Chen and our model more accurately attribute mutations than
  SigProfilerExtractor (Table~\ref{tab:mutations}). As expected, both belief networks
  score comparably with similar architecture.

  \begin{figure*}
      %\figuresize{.3}
      \includegraphics[width=\textwidth]{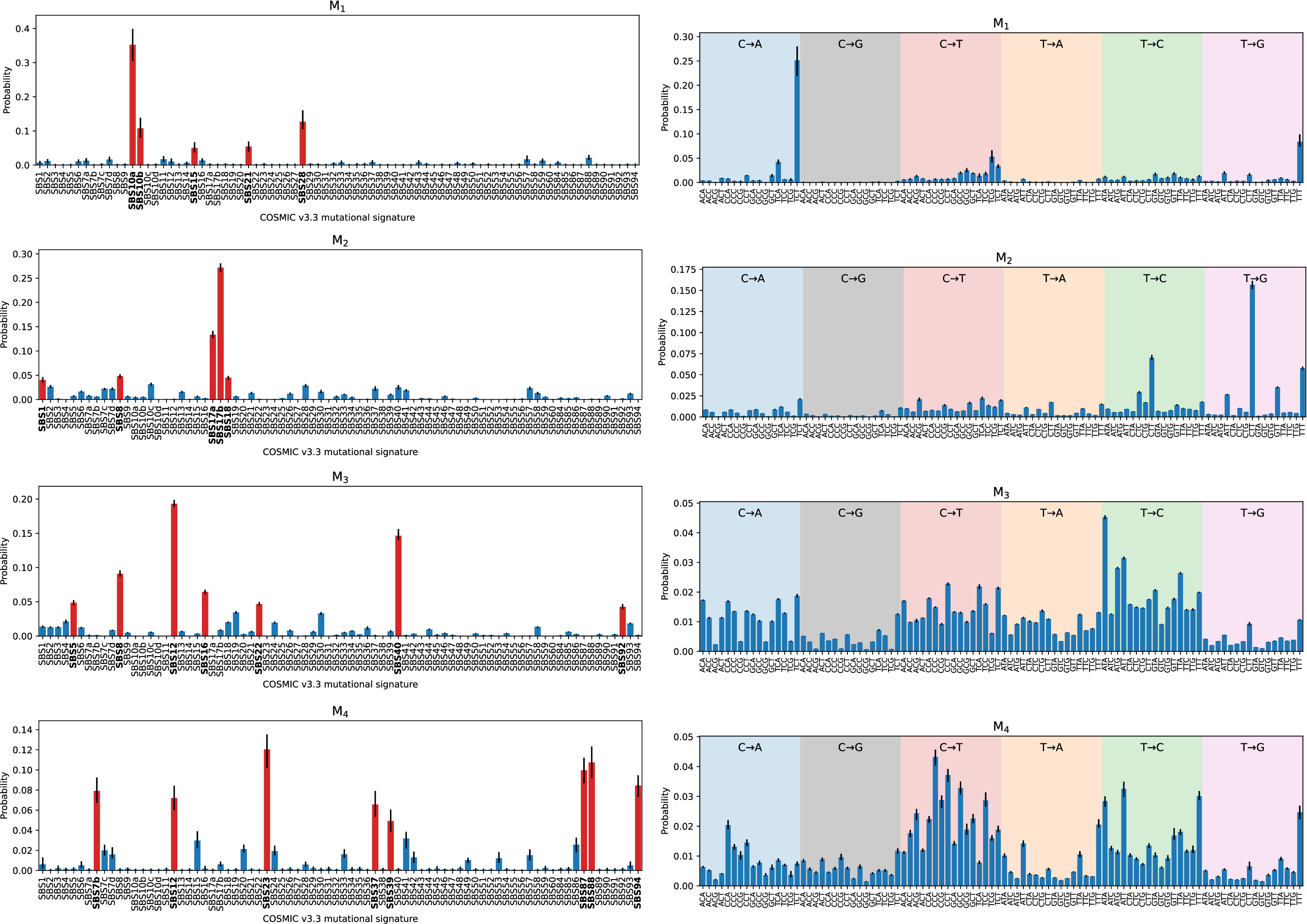}
      \caption{Posterior of four meta-mutational signatures $\phi^{(2)}_{vk}$ (labelled $k=\mathrm{M}_1, \dots, \mathrm{M}_4$) in terms of COSMIC v3.3 mutational signatures $v=\mathrm{SBS}1, \dots, \mathrm{SBS}94$ (left column) and its projection $\sum_{v=\mathrm{SBS}1}^{\mathrm{SBS}94} \phi^{(1)}_{lv} \phi^{(2)}_{vk}$ onto tri-nucleotide single base substitutions $l$ (right column). Bars indicate the average and 95\% quantile range of the posterior samples. On the left, mutational signatures exceeding three times the uniform probability have been marked in bold red.}
      \label{fig:meta-signatures}
  \end{figure*}

  Next, we constructed robust consensus meta-mutational signatures (i.e., topics
  $\phi_{vk}^{(2)}$ from the second layer) from the MBN (Appendix
  \ref{sec:meta_signatures}) that capture co-occurrence of mutational signatures in
  patients. This resulted in four meta-signatures denoted
  $k=\text{M}_1,\dots,\text{M}_4$ (Fig.~\ref{fig:meta-signatures}). 
  
  In brief, the following co-occurrence patterns were identified (an in-depth analysis is provided in the Supplementary Material, Sec.~\ref{sec:meta_signatures_biology}).
  M$_1$ describes the combination of replicative DNA polymerase $\epsilon$ (POLE)
  damage and mismatch-repair deficiency (MMR) (Fig.~\ref{fig:meta-signatures}, first row, left column). Tumours with an ultra-hypermutated phenotype ($\geq$ 100  Mb$^{-1}$) are
  often characterised by these joint disruptions in MMR and POLE~\citep{HODE20}.
  Meta-signature M$_2$ primarily captures, presumably, oxidative stress. To a lesser extent, M$_2$ also captures a signature implicated in BRCA1 and BRCA2 dysfunction in breast cancer~\citep{NIKZ16} which are believed to originate from (uncorrected) replication errors~\citep{SING20}. Meta-signature M$_3$ is marked by components with a pronounced transcriptional strand bias. It contains mutation patterns believed to be related to transcription-coupled
  nucleotide excision repair~\citep{ALEX20}, aging~\citep{ALEX15}, and  aristolochic acid exposure~\citep{HOAN13,POON13,NIK15}. Finally, M$_4$ describes the co-occurrence of several, seemingly disparate, mutational
  signatures. Some are of known aetiology such as ultraviolet light~\citep{NIK15, HAYW17}, thiopurine chemotherapy
  exposure~\citep{LI20}, and damage by the
  Escherichia coli bacterium~\citep{PLEG20,BOOT20}. But M$_4$ also describes the co-occurrence with several signatures of unknown aetiology. 
  
%  Reassuringly, meta-signatures M$_1,\dots,$M$_4$ replicated independently in the PGBN (Fig.~\ref{fig:meta_comparison}, Supplementary Material). 
%To our knowledge, this is the first time a first-principles characterisation of the organising principles of mutagenic processes in cancer has been carried out.
% (Ik heb dit weggehaald; de eerste opmerking 'steals the thunder' van ons model, de tweede opmerking staat al in de discussie en past daar beter)

\section{Discussion \& Conclusion}\label{sec:dicussion_conclusion}
In this paper, we proposed a multi-layer (``deep'') extension of latent Dirichlet
allocation designed to model healthcare data, extending the Zhou--Cong--Chen PGBN model \citep{journals/jmlr/ZhouCC16}. The principal difference
 is that the observed data are multinomial instead of
Poisson distributed: the number of observations per sample is considered
fixed (i.e., conditioned on). In addition to being a practical choice for modelling common data types encountered in healthcare data, it trivially allows for modelling missingness. The model's weights are generated by a Dirichlet and therefore normalized as probabilities, while the dispersion at each layer is controlled by a single concentration parameter to allow sharing of statistical strength between samples, similar to the approach taken in the hierarchical Dirichlet process~\citep{hdp06}. In the
Zhou--Cong--Chen model, Gibbs sampling was achieved by augmentation with Poisson counts
throughout the network; the posterior is then sampled by exploiting an alternative factorization of distributions involving an overdispersed Poisson [negative
binomial distribution, Eq.\ (\ref{eq:factorpoisson})]. In our model, we instead augment with multinomials. Similar to the Zhou--Cong-Chen case, this results in
overdispersed (namely Dirichlet-)multinomials, and we achieve
posterior sampling by developing an analogous factorization, separating the overdispersed distribution's mean and dispersion parameters
and representing their posterior evidence as latent observations from a multinomial and a Chinese restaurant table distribution respectively (Theorem \ref{theorem1}).  In this way, we can treat the
multinomial variable as a (latent) observation generated by the layer above so that the process
can continue upwards, while we obtain the posterior of the CRT governing the dispersion
for this layer using techniques introduced by \citet{ESCO1995} and \citet{hdp06}.

As we show using handwritten digits, the hierarchical setup allows the model to discover different levels of abstractions
in the data. We then applied our model to mutations in
cancer and identified four meta-signatures describing the co-occurrence of
mutagenic processes in cancer. To our knowledge, this is the first time such a
first-principles characterisation of mutagenic processes in cancer has been described.

Our work focuses on the multinomial-distributed outcome variables, given their natural fit for
modelling categorical and count variables frequently encountered in the domain of
healthcare. As a versatile building block, the multinomial can be used to model
real-valued, ordinal, and censored observations using augmentation techniques
(unpublished work), enabling it to accommodate the heterogeneous nature of health data
within a unified framework. By taking a fully Bayesian approach our model is inherently robust
against overfitting, which  is of particular interest for applications in healthcare
where data is often sparse, data collection is expensive, and avoiding 
spurious associations is crucial. 
In complex diseases, like type-1 diabetes and heart disease, patient populations are often highly heterogeneous in clinical presentation and underlying
disease mechanisms~\citep{CORD95, POUL99}. Topic modelling approaches
such as latent Dirichlet allocation have been frequently used to model such complexities~\citep{LU16, LI20b,AHUJ20,BREU21,ZHAN23}, but disregard interactions between topics.  MBNs can model topics and their interactions across multiple layers, capturing more intricate
relationships between patient (sub)populations in terms of their risk factors,
disease mechanisms, and clinical presentation. 
Our approach is also well suited for decomposing 'omics data such as gene expression data, metabolic profiles, and mutation profiles, all representable as counts. While these data are usually analysed using non-negative matrix factorisation (NMF), like in cancer~\citep{HAMA22}, MBNs are able to capture the layered complexity of these datasets.  In addition to avoiding overtraining, inferences under the MBN model come with uncertainty estimates. Initiatives like the 100k genomes project aim to integrate signature analysis into standard patient care~\citep{EVER23}, underscoring the importance of reliable deconvolution and uncertainty quantification to support treatment decisions.

A technical contribution of this paper is the relation between the
Dirichlet-multinomial distribution, the Chinese restaurant table distribution and the
Polya urn scheme presented in Theorem 1, which to the best of our knowledge is novel.
In addition, in the Supplement we present a comprehensive review of the Zhou--Cong--Chen
model, which otherwise is described across several technical papers; we
hope this is useful as an introduction to this elegant model.

\paragraph{Limitations}
Scaling up to large datasets remains challenging using our Gibbs sampling approach,
despite our GPU implementation that can run on multiple accelerators using
JAX~\citep{JAX18}. Approximate Markov chain Monte Carlo~\citep{MA15} and hybrid
approaches~\citep{zhang2018whai,ZHAN20} are an attractive middle-ground between exact
and approximate inference that can scale deep probabilistic models to large datasets. We
leave this problem for future work.

% ACKNOWLEDGEMENTS ONLY GO IN THE CAMERA-READY, NOT THE SUBMISSION
% \acks{Many thanks to all collaborators and funders!}

%Do NOT change font size of references or modify the bibliography style
\bibliography{references}

\begin{thebibliography}{62}
\providecommand{\natexlab}[1]{#1}
\providecommand{\url}[1]{\texttt{#1}}
\expandafter\ifx\csname urlstyle\endcsname\relax
  \providecommand{\doi}[1]{doi: #1}\else
  \providecommand{\doi}{doi: \begingroup \urlstyle{rm}\Url}\fi

\bibitem[Acharya et~al.(2015)Acharya, Teffer, Henderson, Tyler, Zhou, and
  Ghosh]{acharya2015gamma}
Ayan Acharya, Dean Teffer, Jette Henderson, Marcus Tyler, Mingyuan Zhou, and
  Joydeep Ghosh.
\newblock Gamma process poisson factorization for joint modeling of network and
  documents.
\newblock In \emph{Joint European Conference on Machine Learning and Knowledge
  Discovery in Databases}, pages 283--299. Springer, 2015.

\bibitem[Ahuja et~al.(2020)Ahuja, Zhou, He, Sun, Castro, Gainer, Murphy, Hong,
  and Cai]{AHUJ20}
Yuri Ahuja, Doudou Zhou, Zeling He, Jiehuan Sun, Victor~M Castro, Vivian
  Gainer, Shawn~N Murphy, Chuan Hong, and Tianxi Cai.
\newblock surelda: A multidisease automated phenotyping method for the
  electronic health record.
\newblock \emph{J. Am. Med. Inform. Assoc.}, 27\penalty0 (8):\penalty0
  1235--1243, 2020.

\bibitem[Alexandrov et~al.(2013)Alexandrov, Nik-Zainal, Wedge, Aparicio,
  Behjati, Biankin, Bignell, Bolli, Borg, B{\o}rresen-Dale, et~al.]{ALEX13}
Ludmil~B Alexandrov, Serena Nik-Zainal, David~C Wedge, Samuel~AJR Aparicio, Sam
  Behjati, Andrew~V Biankin, Graham~R Bignell, Niccolo Bolli, Ake Borg,
  Anne-Lise B{\o}rresen-Dale, et~al.
\newblock Signatures of mutational processes in human cancer.
\newblock \emph{Nature}, 500\penalty0 (7463):\penalty0 415--421, 2013.

\bibitem[Alexandrov et~al.(2015)Alexandrov, Jones, Wedge, Sale, Campbell,
  Nik-Zainal, and Stratton]{ALEX15}
Ludmil~B Alexandrov, Philip~H Jones, David~C Wedge, Julian~E Sale, Peter~J
  Campbell, Serena Nik-Zainal, and Michael~R Stratton.
\newblock Clock-like mutational processes in human somatic cells.
\newblock \emph{Nat. Genet.}, 47\penalty0 (12):\penalty0 1402--1407, 2015.

\bibitem[Alexandrov et~al.(2020)Alexandrov, Kim, Haradhvala, Huang, Tian~Ng,
  Wu, Boot, Covington, Gordenin, Bergstrom, et~al.]{ALEX20}
Ludmil~B Alexandrov, Jaegil Kim, Nicholas~J Haradhvala, Mi~Ni Huang, Alvin~Wei
  Tian~Ng, Yang Wu, Arnoud Boot, Kyle~R Covington, Dmitry~A Gordenin, Erik~N
  Bergstrom, et~al.
\newblock The repertoire of mutational signatures in human cancer.
\newblock \emph{Nature}, 578\penalty0 (7793):\penalty0 94--101, 2020.

\bibitem[Alpaydin and Kaynak(1998)]{ALPA98}
E.~Alpaydin and C.~Kaynak.
\newblock {Optical Recognition of Handwritten Digits}.
\newblock UCI Machine Learning Repository, 1998.
\newblock {doi}: 10.24432/C50P49.

\bibitem[Alzubaidi et~al.(2023)Alzubaidi, Bai, Al-Sabaawi, Santamaría,
  Albahri, Al-dabbagh, Fadhel, Manoufali, Zhang, Al-Timemy, Duan, Abdullah,
  Farhan, Lu, Gupta, Albu, Abbosh, and Gu]{alzubaidi}
Laith Alzubaidi, Jinshuai Bai, Aiman Al-Sabaawi, Jose Santamaría, A.s Albahri,
  Bashar Al-dabbagh, Mohammed Fadhel, Mohamed Manoufali, Jinglan Zhang, Ali
  Al-Timemy, Ye~Duan, Amjed Abdullah, Laith Farhan, Yi~Lu, Ashish Gupta, Felix
  Albu, Amin Abbosh, and Yuantong Gu.
\newblock A survey on deep learning tools dealing with data scarcity:
  definitions, challenges, solutions, tips, and applications.
\newblock \emph{J. Big Data}, 10, 04 2023.

\bibitem[Beerli(2005)]{beerli06}
Peter Beerli.
\newblock {Comparison of Bayesian and maximum-likelihood inference of
  population genetic parameters}.
\newblock \emph{Bioinformatics}, 22\penalty0 (3):\penalty0 341--345, 11 2005.
\newblock ISSN 1367-4803.

\bibitem[Blei et~al.(2003)Blei, Ng, and Jordan]{blei2003latent}
David~M. Blei, Andrew~Y. Ng, and Michael~I. Jordan.
\newblock {Latent Dirichlet Allocation}.
\newblock \emph{J. Mach. Learn. Res.}, 3:\penalty0 993–1022, mar 2003.

\bibitem[Boot et~al.(2020)Boot, Ng, Chong, Ho, Yu, Tan, Iyer, and
  Rozen]{BOOT20}
Arnoud Boot, Alvin~WT Ng, Fui~Teen Chong, Szu-Chi Ho, Willie Yu, Daniel~SW Tan,
  N~Gopalakrishna Iyer, and Steven~G Rozen.
\newblock Characterization of colibactin-associated mutational signature in an
  asian oral squamous cell carcinoma and in other mucosal tumor types.
\newblock \emph{Genome Res.}, 30\penalty0 (6):\penalty0 803--813, 2020.

\bibitem[Bradbury et~al.(2018)Bradbury, Frostig, Hawkins, Johnson, Leary,
  Maclaurin, Necula, Paszke, Vander{P}las, Wanderman-{M}ilne, and Zhang]{JAX18}
James Bradbury, Roy Frostig, Peter Hawkins, Matthew~James Johnson, Chris Leary,
  Dougal Maclaurin, George Necula, Adam Paszke, Jake Vander{P}las, Skye
  Wanderman-{M}ilne, and Qiao Zhang.
\newblock {JAX}: composable transformations of {P}ython+{N}um{P}y programs,
  2018.

\bibitem[Brady et~al.(2022)Brady, Gout, and Zhang]{BRAD22}
Samuel~W Brady, Alexander~M Gout, and Jinghui Zhang.
\newblock Therapeutic and prognostic insights from the analysis of cancer
  mutational signatures.
\newblock \emph{Trends Genet.}, 38\penalty0 (2):\penalty0 194--208, 2022.

\bibitem[Breuninger et~al.(2021)Breuninger, Wawro, Breuninger, Reitmeier,
  Clavel, Six-Merker, Pestoni, Rohrmann, Rathmann, Peters, et~al.]{BREU21}
Taylor~A Breuninger, Nina Wawro, Jakob Breuninger, Sandra Reitmeier, Thomas
  Clavel, Julia Six-Merker, Giulia Pestoni, Sabine Rohrmann, Wolfgang Rathmann,
  Annette Peters, et~al.
\newblock Associations between habitual diet, metabolic disease, and the gut
  microbiota using latent dirichlet allocation.
\newblock \emph{Microbiome}, 9\penalty0 (1):\penalty0 1--18, 2021.

\bibitem[Cordell and Todd(1995)]{CORD95}
Heather~J Cordell and John~A Todd.
\newblock Multifactorial inheritance in type 1 diabetes.
\newblock \emph{Trends Genet.}, 11\penalty0 (12):\penalty0 499--504, 1995.

\bibitem[Cosmic(2023)]{COSM23}
Cosmic.
\newblock Cosmic - catalogue of somatic mutations in cancer, May 2023.
\newblock URL \url{https://cancer.sanger.ac.uk/cosmic}.
\newblock Accessed: 2023-10-12.

\bibitem[Donker and Groen(2021)]{DONK21}
Hylke~C Donker and Harry~JM Groen.
\newblock Energy-based survival modelling using harmoniums.
\newblock \emph{arXiv preprint arXiv:2110.01960}, 2021.

\bibitem[Emmanuel et~al.(2021)Emmanuel, Maupong, Mpoeleng, Semong, Banyatsang,
  and Tabona]{emmanuel21}
Tlamelo Emmanuel, Thabiso~M. Maupong, Dimane Mpoeleng, Thabo Semong, Mphago
  Banyatsang, and Oteng Tabona.
\newblock A survey on missing data in machine learning.
\newblock \emph{J. Big Data}, 8, 2021.

\bibitem[Escobar and West(1995)]{ESCO1995}
M.~D. Escobar and M.~West.
\newblock {B}ayesian density estimation and inference using mixtures.
\newblock \emph{J. Am. Stat. Assoc.}, 90\penalty0 (430):\penalty0 577--588,
  1995.

\bibitem[Everall et~al.(2023)Everall, Tapinos, Hawari, Cornish, Sud, Chubb,
  Kinnersley, Frangou, Barquin, Jung, et~al.]{EVER23}
Andrew Everall, Avraam Tapinos, Aliah Hawari, Alex Cornish, Amit Sud, Daniel
  Chubb, Ben Kinnersley, Anna Frangou, Miguel Barquin, Josephine Jung, et~al.
\newblock Comprehensive repertoire of the chromosomal alteration and mutational
  signatures across 16 cancer types from 10,983 cancer patients.
\newblock \emph{medRxiv}, pages 2023--06, 2023.

\bibitem[Ferreira et~al.(2022)Ferreira, Kuipers, and Beerenwinkel]{FERR22}
Pedro~F. Ferreira, Jack Kuipers, and Niko Beerenwinkel.
\newblock Deep exponential families for single-cell data analysis.
\newblock \emph{bioRxiv}, 2022.

\bibitem[Gelman et~al.(2013)Gelman, Carlin, Stern, Dunson, Vehtari, and
  Rubin]{GELM13}
Andrew Gelman, John~B Carlin, Hal~S Stern, David~B Dunson, Aki Vehtari, and
  Donald~B Rubin.
\newblock \emph{Bayesian Data Analysis}.
\newblock Chapman and Hall/CRC, New York, 2013.

\bibitem[George and Doss(2017)]{george17}
Clint~P. George and Hani Doss.
\newblock {Principled Selection of Hyperparameters in the Latent Dirichlet
  Allocation Model}.
\newblock \emph{J. Mach. Learn. Res.}, 18:\penalty0 162:1--162:38, 2017.

\bibitem[Griffiths and Steyvers(2004)]{GRIF04}
Thomas~L Griffiths and Mark Steyvers.
\newblock Finding scientific topics.
\newblock \emph{Proc. Natl. Acad. Sci. U.S.A.}, 101\penalty0
  (suppl\_1):\penalty0 5228--5235, 2004.

\bibitem[Hamamoto et~al.(2022)Hamamoto, Takasawa, Machino, Kobayashi,
  Takahashi, Bolatkan, Shinkai, Sakai, Aoyama, Yamada, et~al.]{HAMA22}
Ryuji Hamamoto, Ken Takasawa, Hidenori Machino, Kazuma Kobayashi, Satoshi
  Takahashi, Amina Bolatkan, Norio Shinkai, Akira Sakai, Rina Aoyama, Masayoshi
  Yamada, et~al.
\newblock Application of non-negative matrix factorization in oncology: one
  approach for establishing precision medicine.
\newblock \emph{Brief. Bioinform.}, 23\penalty0 (4):\penalty0 bbac246, 2022.

\bibitem[Hayward et~al.(2017)Hayward, Wilmott, Waddell, Johansson, Field,
  Nones, Patch, Kakavand, Alexandrov, Burke, et~al.]{HAYW17}
Nicholas~K Hayward, James~S Wilmott, Nicola Waddell, Peter~A Johansson,
  Matthew~A Field, Katia Nones, Ann-Marie Patch, Hojabr Kakavand, Ludmil~B
  Alexandrov, Hazel Burke, et~al.
\newblock Whole-genome landscapes of major melanoma subtypes.
\newblock \emph{Nature}, 545\penalty0 (7653):\penalty0 175--180, 2017.

\bibitem[Hinton et~al.(2006)Hinton, Osindero, and Teh]{HintonETAL:06}
Geoffrey Hinton, Simon Osindero, and Yee-Whye Teh.
\newblock A fast learning algorithm for deep belief nets.
\newblock \emph{Neural Comput.}, 18\penalty0 (7):\penalty0 1527--1554, 2006.

\bibitem[Hoang et~al.(2013)Hoang, Chen, Sidorenko, He, Dickman, Yun, Moriya,
  Niknafs, Douville, Karchin, et~al.]{HOAN13}
Margaret~L Hoang, Chung-Hsin Chen, Viktoriya~S Sidorenko, Jian He, Kathleen~G
  Dickman, Byeong~Hwa Yun, Masaaki Moriya, Noushin Niknafs, Christopher
  Douville, Rachel Karchin, et~al.
\newblock Mutational signature of aristolochic acid exposure as revealed by
  whole-exome sequencing.
\newblock \emph{Sci. Transl. Med.}, 5\penalty0 (197):\penalty0 197ra102, 2013.

\bibitem[Hodel et~al.(2020)Hodel, Sun, Ungerleider, Park, Williams, Bauer,
  Immethun, Wang, Suo, Lu, et~al.]{HODE20}
Karl~P Hodel, Meijuan~JS Sun, Nathan Ungerleider, Vivian~S Park, Leonard~G
  Williams, David~L Bauer, Victoria~E Immethun, Jieqiong Wang, Zucai Suo, Hua
  Lu, et~al.
\newblock {POLE} mutation spectra are shaped by the mutant allele identity, its
  abundance, and mismatch repair status.
\newblock \emph{Mol. Cell}, 78\penalty0 (6):\penalty0 1166--1177, 2020.

\bibitem[Islam et~al.(2022)Islam, D{\'\i}az-Gay, Wu, Barnes, Vangara,
  Bergstrom, He, Vella, Wang, Teague, et~al.]{ISL22}
SM~Ashiqul Islam, Marcos D{\'\i}az-Gay, Yang Wu, Mark Barnes, Raviteja Vangara,
  Erik~N Bergstrom, Yudou He, Mike Vella, Jingwei Wang, Jon~W Teague, et~al.
\newblock Uncovering novel mutational signatures by de novo extraction with
  {SigProfilerExtractor}.
\newblock \emph{Cell Genom.}, 2\penalty0 (11), 2022.

\bibitem[Kingma and Welling(2014)]{kingma2013autoencoding}
Diederik~P. Kingma and Max Welling.
\newblock Auto-encoding variational bayes.
\newblock In Yoshua Bengio and Yann LeCun, editors, \emph{2nd International
  Conference on Learning Representations, {ICLR} 2014, Banff, AB, Canada, April
  14-16, 2014, Conference Track Proceedings}, 2014.

\bibitem[Lee and Seung(1999)]{LEE99}
Daniel~D Lee and H~Sebastian Seung.
\newblock Learning the parts of objects by non-negative matrix factorization.
\newblock \emph{Nature}, 401\penalty0 (6755):\penalty0 788--791, 1999.

\bibitem[Li et~al.(2020{\natexlab{a}})Li, Brady, Ma, Shen, Zhang, Li, Szlachta,
  Dong, Liu, Yang, et~al.]{LI20}
Benshang Li, Samuel~W Brady, Xiaotu Ma, Shuhong Shen, Yingchi Zhang, Yongjin
  Li, Karol Szlachta, Li~Dong, Yu~Liu, Fan Yang, et~al.
\newblock Therapy-induced mutations drive the genomic landscape of relapsed
  acute lymphoblastic leukemia.
\newblock \emph{Blood-J. Hematol.}, 135\penalty0 (1):\penalty0 41--55,
  2020{\natexlab{a}}.

\bibitem[Li et~al.(2020{\natexlab{b}})Li, Nair, Lu, Wen, Wang, Dehaghi, Miao,
  Liu, Ordog, Biernacka, et~al.]{LI20b}
Yue Li, Pratheeksha Nair, Xing~Han Lu, Zhi Wen, Yuening Wang, Amir
  Ardalan~Kalantari Dehaghi, Yan Miao, Weiqi Liu, Tamas Ordog, Joanna~M
  Biernacka, et~al.
\newblock Inferring multimodal latent topics from electronic health records.
\newblock \emph{Nat. Commun.}, 11\penalty0 (1):\penalty0 2536,
  2020{\natexlab{b}}.

\bibitem[Lu et~al.(2016)Lu, Wei, and Hsiao]{LU16}
Hsin-Min Lu, Chih-Ping Wei, and Fei-Yuan Hsiao.
\newblock Modeling healthcare data using multiple-channel latent dirichlet
  allocation.
\newblock \emph{J. Biomed. Inform.}, 60:\penalty0 210--223, 2016.

\bibitem[Ma et~al.(2015)Ma, Chen, and Fox]{MA15}
Yi-An Ma, Tianqi Chen, and Emily Fox.
\newblock A complete recipe for stochastic gradient mcmc.
\newblock In C.~Cortes, N.~Lawrence, D.~Lee, M.~Sugiyama, and R.~Garnett,
  editors, \emph{Advances in Neural Information Processing Systems}, volume~28.
  Curran Associates, Inc., 2015.

\bibitem[Minka(2003)]{minka2003}
T.P. Minka.
\newblock Estimating a {Dirichlet} distribution.
\newblock \emph{Ann. Phys.}, 2000\penalty0 (8):\penalty0 1--13, 2003.

\bibitem[Murphy(2023)]{MURP23}
Kevin~P Murphy.
\newblock \emph{{Probabilistic machine learning: Advanced topics}}.
\newblock MIT press, 2023.

\bibitem[Nik-Zainal et~al.(2015)Nik-Zainal, Kucab, Morganella, Glodzik,
  Alexandrov, Arlt, Weninger, Hollstein, Stratton, and Phillips]{NIK15}
Serena Nik-Zainal, Jill~E Kucab, Sandro Morganella, Dominik Glodzik, Ludmil~B
  Alexandrov, Volker~M Arlt, Annette Weninger, Monica Hollstein, Michael~R
  Stratton, and David~H Phillips.
\newblock The genome as a record of environmental exposure.
\newblock \emph{Mutagenesis}, 30\penalty0 (6):\penalty0 763--770, 2015.

\bibitem[Nik-Zainal et~al.(2016)Nik-Zainal, Davies, Staaf, Ramakrishna,
  Glodzik, Zou, Martincorena, Alexandrov, Martin, Wedge, et~al.]{NIKZ16}
Serena Nik-Zainal, Helen Davies, Johan Staaf, Manasa Ramakrishna, Dominik
  Glodzik, Xueqing Zou, Inigo Martincorena, Ludmil~B Alexandrov, Sancha Martin,
  David~C Wedge, et~al.
\newblock Landscape of somatic mutations in 560 breast cancer whole-genome
  sequences.
\newblock \emph{Nature}, 534\penalty0 (7605):\penalty0 47--54, 2016.

\bibitem[Panda et~al.(2019)Panda, Pensia, Mehta, Zhou, and Rai]{PANDA19}
Rajat Panda, Ankit Pensia, Nikhil Mehta, Mingyuan Zhou, and Piyush Rai.
\newblock Deep topic models for multi-label learning.
\newblock In Kamalika Chaudhuri and Masashi Sugiyama, editors,
  \emph{Proceedings of the Twenty-Second International Conference on Artificial
  Intelligence and Statistics}, volume~89 of \emph{Proceedings of Machine
  Learning Research}, pages 2849--2857. PMLR, 16--18 Apr 2019.

\bibitem[Papamarkou et~al.(2024)Papamarkou, Skoularidou, Palla, Aitchison,
  Arbel, Dunson, Filippone, Fortuin, Hennig, Lobato, Hubin, Immer, Karaletsos,
  Khan, Kristiadi, Li, Mandt, Nemeth, Osborne, Rudner, Rügamer, Teh, Welling,
  Wilson, and Zhang]{papamarkou24}
Theodore Papamarkou, Maria Skoularidou, Konstantina Palla, Laurence Aitchison,
  Julyan Arbel, David Dunson, Maurizio Filippone, Vincent Fortuin, Philipp
  Hennig, Jose Miguel~Hernandez Lobato, Aliaksandr Hubin, Alexander Immer,
  Theofanis Karaletsos, Mohammad~Emtiyaz Khan, Agustinus Kristiadi, Yingzhen
  Li, Stephan Mandt, Christopher Nemeth, Michael~A. Osborne, Tim G.~J. Rudner,
  David Rügamer, Yee~Whye Teh, Max Welling, Andrew~Gordon Wilson, and Ruqi
  Zhang.
\newblock Position paper: Bayesian deep learning in the age of large-scale ai,
  2024.

\bibitem[Patterson et~al.(2023)Patterson, Elbasir, Tian, and Auslander]{PATT23}
Andrew Patterson, Abdurrahman Elbasir, Bin Tian, and Noam Auslander.
\newblock Computational methods summarizing mutational patterns in cancer:
  Promise and limitations for clinical applications.
\newblock \emph{Cancers}, 15\penalty0 (7):\penalty0 1958, 2023.

\bibitem[Pedregosa et~al.(2011)Pedregosa, Varoquaux, Gramfort, Michel, Thirion,
  Grisel, Blondel, Prettenhofer, Weiss, Dubourg, Vanderplas, Passos,
  Cournapeau, Brucher, Perrot, and {{\'E}}douard Duchesnay]{SKL11}
Fabian Pedregosa, Ga{{\"e}}l Varoquaux, Alexandre Gramfort, Vincent Michel,
  Bertrand Thirion, Olivier Grisel, Mathieu Blondel, Peter Prettenhofer, Ron
  Weiss, Vincent Dubourg, Jake Vanderplas, Alexandre Passos, David Cournapeau,
  Matthieu Brucher, Matthieu Perrot, and {{\'E}}douard Duchesnay.
\newblock {Scikit-learn: Machine Learning in Python}.
\newblock \emph{J. Mach. Learn. Res.}, 12\penalty0 (85):\penalty0 2825--2830,
  2011.

\bibitem[Pleguezuelos-Manzano et~al.(2020)Pleguezuelos-Manzano, Puschhof,
  Rosendahl~Huber, van Hoeck, Wood, Nomburg, Gurjao, Manders, Dalmasso, Stege,
  et~al.]{PLEG20}
Cayetano Pleguezuelos-Manzano, Jens Puschhof, Axel Rosendahl~Huber, Arne van
  Hoeck, Henry~M Wood, Jason Nomburg, Carino Gurjao, Freek Manders, Guillaume
  Dalmasso, Paul~B Stege, et~al.
\newblock Mutational signature in colorectal cancer caused by genotoxic
  pks${}^+$ {E}. coli.
\newblock \emph{Nature}, 580\penalty0 (7802):\penalty0 269--273, 2020.

\bibitem[Poon et~al.(2013)Poon, Pang, McPherson, Yu, Huang, Guan, Weng, Siew,
  Liu, Heng, et~al.]{POON13}
Song~Ling Poon, See-Tong Pang, John~R McPherson, Willie Yu, Kie~Kyon Huang,
  Peiyong Guan, Wen-Hui Weng, Ee~Yan Siew, Yujing Liu, Hong~Lee Heng, et~al.
\newblock Genome-wide mutational signatures of aristolochic acid and its
  application as a screening tool.
\newblock \emph{Science Transl. Med.}, 5\penalty0 (197):\penalty0 197ra101,
  2013.

\bibitem[Poulter(1999)]{POUL99}
N~Poulter.
\newblock Coronary heart disease is a multifactorial disease.
\newblock \emph{Am. J. Hypertens.}, 12\penalty0 (S6):\penalty0 92S--95S, 1999.

\bibitem[Ranganath et~al.(2015)Ranganath, Tang, Charlin, and Blei]{RANG15}
Rajesh Ranganath, Linpeng Tang, Laurent Charlin, and David Blei.
\newblock {Deep Exponential Families}.
\newblock In Guy Lebanon and S.~V.~N. Vishwanathan, editors, \emph{Proceedings
  of the Eighteenth International Conference on Artificial Intelligence and
  Statistics}, volume~38 of \emph{Proceedings of Machine Learning Research},
  pages 762--771, San Diego, California, USA, 09--12 May 2015. PMLR.

\bibitem[Ranganath et~al.(2016)Ranganath, Tran, and Blei]{RANG16}
Rajesh Ranganath, Dustin Tran, and David Blei.
\newblock Hierarchical variational models.
\newblock In Maria~Florina Balcan and Kilian~Q. Weinberger, editors,
  \emph{Proceedings of The 33rd International Conference on Machine Learning},
  volume~48 of \emph{Proceedings of Machine Learning Research}, pages 324--333,
  New York, New York, USA, 20--22 Jun 2016. PMLR.

\bibitem[Singh et~al.(2020)Singh, Rastogi, Hu, Wang, and De]{SING20}
Vinod~Kumar Singh, Arnav Rastogi, Xiaoju Hu, Yaqun Wang, and Subhajyoti De.
\newblock Mutational signature {SBS8} predominantly arises due to late
  replication errors in cancer.
\newblock \emph{Commun. Biol.}, 3\penalty0 (1):\penalty0 421, 2020.

\bibitem[Smith and Naylor(1987)]{smith87}
Richard~L. Smith and J.~C. Naylor.
\newblock A comparison of maximum likelihood and bayesian estimators for the
  three- parameter weibull distribution.
\newblock \emph{J. R. Stat. Soc. C: Appl. Stat.)}, 36\penalty0 (3):\penalty0
  358--369, 1987.
\newblock ISSN 00359254, 14679876.

\bibitem[Soleimani et~al.(2017)Soleimani, Hensman, and
  Saria]{soleimani2017scalable}
Hossein Soleimani, James Hensman, and Suchi Saria.
\newblock Scalable joint models for reliable uncertainty-aware event
  prediction.
\newblock \emph{IEEE T. Pattern Anal.}, 40\penalty0 (8):\penalty0 1948--1963,
  2017.

\bibitem[Talts et~al.(2018)Talts, Betancourt, Simpson, Vehtari, and
  Gelman]{TALT18}
Sean Talts, Michael Betancourt, Daniel Simpson, Aki Vehtari, and Andrew Gelman.
\newblock Validating bayesian inference algorithms with simulation-based
  calibration.
\newblock \emph{arXiv preprint arXiv:1804.06788}, 2018.

\bibitem[Tate et~al.(2019)Tate, Bamford, Jubb, Sondka, Beare, Bindal,
  Boutselakis, Cole, Creatore, Dawson, et~al.]{TATE19}
John~G Tate, Sally Bamford, Harry~C Jubb, Zbyslaw Sondka, David~M Beare, Nidhi
  Bindal, Harry Boutselakis, Charlotte~G Cole, Celestino Creatore, Elisabeth
  Dawson, et~al.
\newblock Cosmic: the catalogue of somatic mutations in cancer.
\newblock \emph{Nucleic Acids Res.}, 47\penalty0 (D1):\penalty0 D941--D947,
  2019.

\bibitem[Teh et~al.(2006)Teh, Jordan, Beal, and Blei]{hdp06}
Yee~Whye Teh, Michael~I. Jordan, Matthew~J. Beal, and David~M. Blei.
\newblock Hierarchical dirichlet processes.
\newblock \emph{J. Am. Stat. Assoc.}, 101\penalty0 (476):\penalty0 1566--1581,
  2006.

\bibitem[Welling et~al.(2004)Welling, Rosen-Zvi, and Hinton]{WELL04}
Max Welling, Michal Rosen-Zvi, and Geoffrey~E Hinton.
\newblock Exponential family harmoniums with an application to information
  retrieval.
\newblock \emph{Adv. Neural Inf. Process. Syst.}, 17, 2004.

\bibitem[Zhang et~al.(2018)Zhang, Chen, Guo, and Zhou]{zhang2018whai}
Hao Zhang, Bo~Chen, Dandan Guo, and Mingyuan Zhou.
\newblock Whai: Weibull hybrid autoencoding inference for deep topic modeling.
\newblock In \emph{ICLR}, 2018.

\bibitem[Zhang et~al.(2020)Zhang, Chen, Cong, Guo, Liu, and Zhou]{ZHAN20}
Hao Zhang, Bo~Chen, Yulai Cong, Dandan Guo, Hongwei Liu, and Mingyuan Zhou.
\newblock Deep autoencoding topic model with scalable hybrid bayesian
  inference.
\newblock \emph{IEEE T. Pattern Anal.}, 43\penalty0 (12):\penalty0 4306--4322,
  2020.

\bibitem[Zhang et~al.(2023)Zhang, Jiang, Mentzer, McVean, and Lunter]{ZHAN23}
Yidong Zhang, Xilin Jiang, Alexander~J Mentzer, Gil McVean, and Gerton Lunter.
\newblock Topic modeling identifies novel genetic loci associated with
  multimorbidities in uk biobank.
\newblock \emph{Cell Genom.}, 3\penalty0 (8), 2023.

\bibitem[Zhao et~al.(2018)Zhao, Du, Buntine, and Zhou]{zhao2018dirichlet}
He~Zhao, Lan Du, Wray Buntine, and Mingyuan Zhou.
\newblock Dirichlet belief networks for topic structure learning.
\newblock In S.~Bengio, H.~Wallach, H.~Larochelle, K.~Grauman, N.~Cesa-Bianchi,
  and R.~Garnett, editors, \emph{Advances in Neural Information Processing
  Systems}, volume~31. Curran Associates, Inc., 2018.

\bibitem[Zhou and Carin(2015)]{journals/pami/ZhouC15}
Mingyuan Zhou and Lawrence Carin.
\newblock Negative binomial process count and mixture modeling.
\newblock \emph{IEEE T. Pattern Anal.}, 37\penalty0 (2):\penalty0 307--320,
  2015.

\bibitem[Zhou et~al.(2012)Zhou, Hannah, Dunson, and Carin]{zhou2012beta}
Mingyuan Zhou, Lauren Hannah, David Dunson, and Lawrence Carin.
\newblock Beta-negative binomial process and poisson factor analysis.
\newblock In Neil~D. Lawrence and Mark Girolami, editors, \emph{Proceedings of
  the Fifteenth International Conference on Artificial Intelligence and
  Statistics}, volume~22 of \emph{Proceedings of Machine Learning Research},
  pages 1462--1471, La Palma, Canary Islands, 21--23 Apr 2012. PMLR.

\bibitem[Zhou et~al.(2016)Zhou, Cong, and Chen]{journals/jmlr/ZhouCC16}
Mingyuan Zhou, Yulai Cong, and Bo~Chen.
\newblock Augmentable gamma belief networks.
\newblock \emph{J. Mach. Learn. Res.}, 17\penalty0 (163):\penalty0 1--44, 2016.

\end{thebibliography}


\begin{thebibliography}{40}
        \providecommand{\natexlab}[1]{#1}
        \providecommand{\url}[1]{\texttt{#1}}
        \expandafter\ifx\csname urlstyle\endcsname\relax
          \providecommand{\doi}[1]{doi: #1}\else
          \providecommand{\doi}{doi: \begingroup \urlstyle{rm}\Url}\fi
        
        \bibitem[Alexandrov et~al.(2015)Alexandrov, Jones, Wedge, Sale, Campbell,
          Nik-Zainal, and Stratton]{S_ALEX15}
        Ludmil~B Alexandrov, Philip~H Jones, David~C Wedge, Julian~E Sale, Peter~J
          Campbell, Serena Nik-Zainal, and Michael~R Stratton.
        \newblock Clock-like mutational processes in human somatic cells.
        \newblock \emph{Nat. Genet.}, 47\penalty0 (12):\penalty0 1402--1407, 2015.
        
        \bibitem[Alexandrov et~al.(2020)Alexandrov, Kim, Haradhvala, Huang, Tian~Ng,
          Wu, Boot, Covington, Gordenin, Bergstrom, et~al.]{S_ALEX20}
        Ludmil~B Alexandrov, Jaegil Kim, Nicholas~J Haradhvala, Mi~Ni Huang, Alvin~Wei
          Tian~Ng, Yang Wu, Arnoud Boot, Kyle~R Covington, Dmitry~A Gordenin, Erik~N
          Bergstrom, et~al.
        \newblock The repertoire of mutational signatures in human cancer.
        \newblock \emph{Nature}, 578\penalty0 (7793):\penalty0 94--101, 2020.
        
        \bibitem[Antoniak(1974)]{S_antoniak1974}
        Charles~E. Antoniak.
        \newblock {Mixtures of Dirichlet Processes with Applications to Bayesian
          Nonparametric Problems}.
        \newblock \emph{Ann. Stat.}, 2\penalty0 (6):\penalty0 1152 -- 1174, 1974.
        
        \bibitem[Blei et~al.(2003)Blei, Ng, and Jordan]{S_blei2003latent}
        David~M. Blei, Andrew~Y. Ng, and Michael~I. Jordan.
        \newblock {Latent Dirichlet Allocation}.
        \newblock \emph{J. Mach. Learn. Res.}, 3:\penalty0 993–1022, mar 2003.
        
        \bibitem[Boot et~al.(2020)Boot, Ng, Chong, Ho, Yu, Tan, Iyer, and
          Rozen]{S_BOOT20}
        Arnoud Boot, Alvin~WT Ng, Fui~Teen Chong, Szu-Chi Ho, Willie Yu, Daniel~SW Tan,
          N~Gopalakrishna Iyer, and Steven~G Rozen.
        \newblock Characterization of colibactin-associated mutational signature in an
          asian oral squamous cell carcinoma and in other mucosal tumor types.
        \newblock \emph{Genome Res.}, 30\penalty0 (6):\penalty0 803--813, 2020.
        
        \bibitem[Christensen et~al.(2019)Christensen, Van~der Roest, Besselink,
          Janssen, Boymans, Martens, Yaspo, Priestley, Kuijk, Cuppen, et~al.]{S_CHRI19}
        Sharon Christensen, Bastiaan Van~der Roest, Nicolle Besselink, Roel Janssen,
          Sander Boymans, John~WM Martens, Marie-Laure Yaspo, Peter Priestley, Ewart
          Kuijk, Edwin Cuppen, et~al.
        \newblock {5-Fluorouracil treatment induces characteristic T>G mutations in
          human cancer}.
        \newblock \emph{Nature Commun.}, 10\penalty0 (1):\penalty0 4571, 2019.
        
        \bibitem[Cosmic(2023)]{S_COSM23}
        Cosmic.
        \newblock Cosmic - catalogue of somatic mutations in cancer, May 2023.
        \newblock URL \url{https://cancer.sanger.ac.uk/cosmic}.
        \newblock Accessed: 2023-10-12.
        
        \bibitem[Crouse(2016)]{S_CROU16}
        David~F Crouse.
        \newblock On implementing 2d rectangular assignment algorithms.
        \newblock \emph{IEEE T. Aero. Elec. Sys.}, 52\penalty0 (4):\penalty0
          1679--1696, 2016.
        
        \bibitem[Donker et~al.(2023)Donker, van Es, Tamminga, Lunter, van Kempen,
          Schuuring, Hiltermann, and Groen]{S_DONK23}
        H.~C. Donker, B~van Es, M~Tamminga, G.~A. Lunter, L.~C. L.~T. van Kempen,
          E~Schuuring, T.~J.~N. Hiltermann, and H.~J.~M. Groen.
        \newblock Using genomic scars to select immunotherapy beneficiaries in advanced
          non-small cell lung cancer.
        \newblock \emph{Sci. Rep.}, 13\penalty0 (1):\penalty0 6581, 2023.
        
        \bibitem[Escobar and West(1995)]{S_ESCO1995}
        M.~D. Escobar and M.~West.
        \newblock {B}ayesian density estimation and inference using mixtures.
        \newblock \emph{J. Am. Stat. Assoc.}, 90\penalty0 (430):\penalty0 577--588,
          1995.
        
        \bibitem[Hayward et~al.(2017)Hayward, Wilmott, Waddell, Johansson, Field,
          Nones, Patch, Kakavand, Alexandrov, Burke, et~al.]{S_HAYW17}
        Nicholas~K Hayward, James~S Wilmott, Nicola Waddell, Peter~A Johansson,
          Matthew~A Field, Katia Nones, Ann-Marie Patch, Hojabr Kakavand, Ludmil~B
          Alexandrov, Hazel Burke, et~al.
        \newblock Whole-genome landscapes of major melanoma subtypes.
        \newblock \emph{Nature}, 545\penalty0 (7653):\penalty0 175--180, 2017.
        
        \bibitem[Hoang et~al.(2013)Hoang, Chen, Sidorenko, He, Dickman, Yun, Moriya,
          Niknafs, Douville, Karchin, et~al.]{S_HOAN13}
        Margaret~L Hoang, Chung-Hsin Chen, Viktoriya~S Sidorenko, Jian He, Kathleen~G
          Dickman, Byeong~Hwa Yun, Masaaki Moriya, Noushin Niknafs, Christopher
          Douville, Rachel Karchin, et~al.
        \newblock Mutational signature of aristolochic acid exposure as revealed by
          whole-exome sequencing.
        \newblock \emph{Sci. Transl. Med.}, 5\penalty0 (197):\penalty0 197ra102, 2013.
        
        \bibitem[Hodel et~al.(2020)Hodel, Sun, Ungerleider, Park, Williams, Bauer,
          Immethun, Wang, Suo, Lu, et~al.]{S_HODE20}
        Karl~P Hodel, Meijuan~JS Sun, Nathan Ungerleider, Vivian~S Park, Leonard~G
          Williams, David~L Bauer, Victoria~E Immethun, Jieqiong Wang, Zucai Suo, Hua
          Lu, et~al.
        \newblock {POLE} mutation spectra are shaped by the mutant allele identity, its
          abundance, and mismatch repair status.
        \newblock \emph{Mol. Cell}, 78\penalty0 (6):\penalty0 1166--1177, 2020.
        
        \bibitem[Islam et~al.(2022)Islam, D{\'\i}az-Gay, Wu, Barnes, Vangara,
          Bergstrom, He, Vella, Wang, Teague, et~al.]{S_ISL22}
        SM~Ashiqul Islam, Marcos D{\'\i}az-Gay, Yang Wu, Mark Barnes, Raviteja Vangara,
          Erik~N Bergstrom, Yudou He, Mike Vella, Jingwei Wang, Jon~W Teague, et~al.
        \newblock Uncovering novel mutational signatures by de novo extraction with
          {SigProfilerExtractor}.
        \newblock \emph{Cell Genom.}, 2\penalty0 (11), 2022.
        
        \bibitem[Kucab et~al.(2019)Kucab, Zou, Morganella, Joel, Nanda, Nagy, Gomez,
          Degasperi, Harris, Jackson, et~al.]{S_KUCA19}
        Jill~E Kucab, Xueqing Zou, Sandro Morganella, Madeleine Joel, A~Scott Nanda,
          Eszter Nagy, Celine Gomez, Andrea Degasperi, Rebecca Harris, Stephen~P
          Jackson, et~al.
        \newblock A compendium of mutational signatures of environmental agents.
        \newblock \emph{Cell}, 177\penalty0 (4):\penalty0 821--836, 2019.
        
        \bibitem[Lawson et~al.(2020)Lawson, Abascal, Coorens, Hooks, O’Neill,
          Latimer, Raine, Sanders, Warren, Mahbubani, et~al.]{S_LAWS20}
        Andrew~RJ Lawson, Federico Abascal, Tim~HH Coorens, Yvette Hooks, Laura
          O’Neill, Calli Latimer, Keiran Raine, Mathijs~A Sanders, Anne~Y Warren,
          Krishnaa~TA Mahbubani, et~al.
        \newblock Extensive heterogeneity in somatic mutation and selection in the
          human bladder.
        \newblock \emph{Science}, 370\penalty0 (6512):\penalty0 75--82, 2020.
        
        \bibitem[Lee-Six et~al.(2019)Lee-Six, Olafsson, Ellis, Osborne, Sanders, Moore,
          Georgakopoulos, Torrente, Noorani, Goddard, et~al.]{S_LEE19}
        Henry Lee-Six, Sigurgeir Olafsson, Peter Ellis, Robert~J Osborne, Mathijs~A
          Sanders, Luiza Moore, Nikitas Georgakopoulos, Franco Torrente, Ayesha
          Noorani, Martin Goddard, et~al.
        \newblock The landscape of somatic mutation in normal colorectal epithelial
          cells.
        \newblock \emph{Nature}, 574\penalty0 (7779):\penalty0 532--537, 2019.
        
        \bibitem[Li et~al.(2020)Li, Brady, Ma, Shen, Zhang, Li, Szlachta, Dong, Liu,
          Yang, et~al.]{S_LI20}
        Benshang Li, Samuel~W Brady, Xiaotu Ma, Shuhong Shen, Yingchi Zhang, Yongjin
          Li, Karol Szlachta, Li~Dong, Yu~Liu, Fan Yang, et~al.
        \newblock Therapy-induced mutations drive the genomic landscape of relapsed
          acute lymphoblastic leukemia.
        \newblock \emph{Blood-J. Hematol.}, 135\penalty0 (1):\penalty0 41--55, 2020.
        
        \bibitem[Li et~al.(2018)Li, Cuevas, Zhang, Lu, Alam, Fu, You, Akbay, Zhang,
          Castrillon, et~al.]{S_LI18}
        Hao-Dong Li, Ileana Cuevas, Musi Zhang, Changzheng Lu, Md~Maksudul Alam,
          Yang-Xin Fu, M~James You, Esra~A Akbay, He~Zhang, Diego~H Castrillon, et~al.
        \newblock Polymerase-mediated ultramutagenesis in mice produces diverse cancers
          with high mutational load.
        \newblock \emph{J. Clin. Invest.}, 128\penalty0 (9):\penalty0 4179--4191, 2018.
        
        \bibitem[Meier et~al.(2018)Meier, Volkova, Hong, Schofield, Campbell, Gerstung,
          and Gartner]{S_MEIE18}
        Bettina Meier, Nadezda~V Volkova, Ye~Hong, Pieta Schofield, Peter~J Campbell,
          Moritz Gerstung, and Anton Gartner.
        \newblock {Mutational signatures of DNA mismatch repair deficiency in C.
          elegans and human cancers}.
        \newblock \emph{Genome Res.}, 28\penalty0 (5):\penalty0 666--675, 2018.
        
        \bibitem[Minka and Lafferty(2002)]{S_minka2002}
        T.~Minka and J.~Lafferty.
        \newblock Expectation-{P}ropagation for the generative aspect model.
        \newblock In \emph{Proceedings of the 18th {C}onference on {U}ncertainty in
          {A}rtificial {I}ntelligence {(UAI)}}, San Francisco, CA, 2002.
        
        \bibitem[Murphy(2023)]{S_MURP23}
        Kevin~P Murphy.
        \newblock \emph{{Probabilistic machine learning: Advanced topics}}.
        \newblock MIT press, 2023.
        
        \bibitem[Nie et~al.(2013)Nie, Gan, Shi, Hu, Chen, Hayakawa, Sekiguchi, Cai,
          et~al.]{S_NIE13}
        Ben Nie, Wei Gan, Fei Shi, Guo-Xin Hu, Lian-Guo Chen, Hiroshi Hayakawa, Mutsuo
          Sekiguchi, Jian-Ping Cai, et~al.
        \newblock Age-dependent accumulation of 8-oxoguanine in the dna and rna in
          various rat tissues.
        \newblock \emph{Oxid. Med. Cell. Longev.}, 2013, 2013.
        
        \bibitem[Nik-Zainal et~al.(2012)Nik-Zainal, Alexandrov, Wedge, Van~Loo,
          Greenman, Raine, Jones, Hinton, Marshall, Stebbings, et~al.]{S_NIK12}
        Serena Nik-Zainal, Ludmil~B Alexandrov, David~C Wedge, Peter Van~Loo,
          Christopher~D Greenman, Keiran Raine, David Jones, Jonathan Hinton, John
          Marshall, Lucy~A Stebbings, et~al.
        \newblock Mutational processes molding the genomes of 21 breast cancers.
        \newblock \emph{Cell}, 149\penalty0 (5):\penalty0 979--993, 2012.
        
        \bibitem[Nik-Zainal et~al.(2015)Nik-Zainal, Kucab, Morganella, Glodzik,
          Alexandrov, Arlt, Weninger, Hollstein, Stratton, and Phillips]{S_NIK15}
        Serena Nik-Zainal, Jill~E Kucab, Sandro Morganella, Dominik Glodzik, Ludmil~B
          Alexandrov, Volker~M Arlt, Annette Weninger, Monica Hollstein, Michael~R
          Stratton, and David~H Phillips.
        \newblock The genome as a record of environmental exposure.
        \newblock \emph{Mutagenesis}, 30\penalty0 (6):\penalty0 763--770, 2015.
        
        \bibitem[Nik-Zainal et~al.(2016)Nik-Zainal, Davies, Staaf, Ramakrishna,
          Glodzik, Zou, Martincorena, Alexandrov, Martin, Wedge, et~al.]{S_NIKZ16}
        Serena Nik-Zainal, Helen Davies, Johan Staaf, Manasa Ramakrishna, Dominik
          Glodzik, Xueqing Zou, Inigo Martincorena, Ludmil~B Alexandrov, Sancha Martin,
          David~C Wedge, et~al.
        \newblock Landscape of somatic mutations in 560 breast cancer whole-genome
          sequences.
        \newblock \emph{Nature}, 534\penalty0 (7605):\penalty0 47--54, 2016.
        
        \bibitem[Nones et~al.(2014)Nones, Waddell, Wayte, Patch, Bailey, Newell,
          Holmes, Fink, Quinn, Tang, et~al.]{S_NONE14}
        Katia Nones, Nicola Waddell, Nicci Wayte, Ann-Marie Patch, Peter Bailey,
          Felicity Newell, Oliver Holmes, J~Lynn Fink, Michael~CJ Quinn, Yue~Hang Tang,
          et~al.
        \newblock Genomic catastrophes frequently arise in esophageal adenocarcinoma
          and drive tumorigenesis.
        \newblock \emph{Nat. Commun.}, 5\penalty0 (1):\penalty0 5224, 2014.
        
        \bibitem[Pleguezuelos-Manzano et~al.(2020)Pleguezuelos-Manzano, Puschhof,
          Rosendahl~Huber, van Hoeck, Wood, Nomburg, Gurjao, Manders, Dalmasso, Stege,
          et~al.]{S_PLEG20}
        Cayetano Pleguezuelos-Manzano, Jens Puschhof, Axel Rosendahl~Huber, Arne van
          Hoeck, Henry~M Wood, Jason Nomburg, Carino Gurjao, Freek Manders, Guillaume
          Dalmasso, Paul~B Stege, et~al.
        \newblock Mutational signature in colorectal cancer caused by genotoxic
          pks${}^+$ {E}. coli.
        \newblock \emph{Nature}, 580\penalty0 (7802):\penalty0 269--273, 2020.
        
        \bibitem[Poetsch et~al.(2018)Poetsch, Boulton, and Luscombe]{S_POET18}
        Anna~R Poetsch, Simon~J Boulton, and Nicholas~M Luscombe.
        \newblock Genomic landscape of oxidative dna damage and repair reveals
          regioselective protection from mutagenesis.
        \newblock \emph{Genome Biol.}, 19\penalty0 (1):\penalty0 1--23, 2018.
        
        \bibitem[Poon et~al.(2013)Poon, Pang, McPherson, Yu, Huang, Guan, Weng, Siew,
          Liu, Heng, et~al.]{S_POON13}
        Song~Ling Poon, See-Tong Pang, John~R McPherson, Willie Yu, Kie~Kyon Huang,
          Peiyong Guan, Wen-Hui Weng, Ee~Yan Siew, Yujing Liu, Hong~Lee Heng, et~al.
        \newblock Genome-wide mutational signatures of aristolochic acid and its
          application as a screening tool.
        \newblock \emph{Science Transl. Med.}, 5\penalty0 (197):\penalty0 197ra101,
          2013.
        
        \bibitem[Rousseeuw(1987)]{S_ROUS87}
        Peter~J Rousseeuw.
        \newblock Silhouettes: a graphical aid to the interpretation and validation of
          cluster analysis.
        \newblock \emph{J. Comput. Appl. Math.}, 20:\penalty0 53--65, 1987.
        
        \bibitem[Secrier et~al.(2016)Secrier, Li, De~Silva, Eldridge, Contino,
          Bornschein, MacRae, Grehan, O'Donovan, Miremadi, et~al.]{S_SECR16}
        Maria Secrier, Xiaodun Li, Nadeera De~Silva, Matthew~D Eldridge, Gianmarco
          Contino, Jan Bornschein, Shona MacRae, Nicola Grehan, Maria O'Donovan, Ahmad
          Miremadi, et~al.
        \newblock Mutational signatures in esophageal adenocarcinoma define
          etiologically distinct subgroups with therapeutic relevance.
        \newblock \emph{Nat. Genet.}, 48\penalty0 (10):\penalty0 1131--1141, 2016.
        
        \bibitem[Singh et~al.(2020)Singh, Rastogi, Hu, Wang, and De]{S_SING20}
        Vinod~Kumar Singh, Arnav Rastogi, Xiaoju Hu, Yaqun Wang, and Subhajyoti De.
        \newblock Mutational signature {SBS8} predominantly arises due to late
          replication errors in cancer.
        \newblock \emph{Commun. Biol.}, 3\penalty0 (1):\penalty0 421, 2020.
        
        \bibitem[Tate et~al.(2019)Tate, Bamford, Jubb, Sondka, Beare, Bindal,
          Boutselakis, Cole, Creatore, Dawson, et~al.]{S_TATE19}
        John~G Tate, Sally Bamford, Harry~C Jubb, Zbyslaw Sondka, David~M Beare, Nidhi
          Bindal, Harry Boutselakis, Charlotte~G Cole, Celestino Creatore, Elisabeth
          Dawson, et~al.
        \newblock Cosmic: the catalogue of somatic mutations in cancer.
        \newblock \emph{Nucleic Acids Res.}, 47\penalty0 (D1):\penalty0 D941--D947,
          2019.
        
        \bibitem[Teh et~al.(2006)Teh, Jordan, Beal, and Blei]{S_hdp06}
        Yee~Whye Teh, Michael~I. Jordan, Matthew~J. Beal, and David~M. Blei.
        \newblock Hierarchical dirichlet processes.
        \newblock \emph{J. Am. Stat. Assoc.}, 101\penalty0 (476):\penalty0 1566--1581,
          2006.
        
        \bibitem[Tomkova et~al.(2018)Tomkova, Tomek, Kriaucionis, and
          Schuster-B{\"o}ckler]{S_TOMK18}
        Marketa Tomkova, Jakub Tomek, Skirmantas Kriaucionis, and Benjamin
          Schuster-B{\"o}ckler.
        \newblock Mutational signature distribution varies with dna replication timing
          and strand asymmetry.
        \newblock \emph{Genome Biol.}, 19\penalty0 (1):\penalty0 1--12, 2018.
        
        \bibitem[Vehtari et~al.(2021)Vehtari, Gelman, Simpson, Carpenter, and
          B{\"u}rkner]{S_VEHT21}
        Aki Vehtari, Andrew Gelman, Daniel Simpson, Bob Carpenter, and Paul-Christian
          B{\"u}rkner.
        \newblock Rank-normalization, folding, and localization: An improved r-hat for
          assessing convergence of mcmc (with discussion).
        \newblock \emph{Bayesian Anal.}, 16\penalty0 (2):\penalty0 667--718, 2021.
        
        \bibitem[Zhou and Carin(2015)]{S_journals/pami/ZhouC15}
        Mingyuan Zhou and Lawrence Carin.
        \newblock Negative binomial process count and mixture modeling.
        \newblock \emph{IEEE T. Pattern Anal.}, 37\penalty0 (2):\penalty0 307--320,
          2015.
        \newblock \doi{10.1109/TPAMI.2013.211}.
        
        \bibitem[Zhou et~al.(2012)Zhou, Hannah, Dunson, and Carin]{S_zhou2012beta}
        Mingyuan Zhou, Lauren Hannah, David Dunson, and Lawrence Carin.
        \newblock Beta-negative binomial process and poisson factor analysis.
        \newblock In Neil~D. Lawrence and Mark Girolami, editors, \emph{Proceedings of
          the Fifteenth International Conference on Artificial Intelligence and
          Statistics}, volume~22 of \emph{Proceedings of Machine Learning Research},
          pages 1462--1471, La Palma, Canary Islands, 21--23 Apr 2012. PMLR.
        
        \bibitem[Zhou et~al.(2016)Zhou, Cong, and Chen]{S_journals/jmlr/ZhouCC16}
        Mingyuan Zhou, Yulai Cong, and Bo~Chen.
        \newblock Augmentable gamma belief networks.
        \newblock \emph{J. Mach. Learn. Res.}, 17\penalty0 (163):\penalty0 1--44, 2016.
        
    \end{thebibliography}

%%%%%%%%%% Merge with supplemental materials %%%%%%%%%%
%\widetext
\clearpage
\onecolumn  % Reset to single column mode.
\begin{center}
    \textbf{\large Supplemental Materials for Multinomial belief networks}
\end{center}
%%%%%%%%%% Merge with supplemental materials %%%%%%%%%%
%%%%%%%%%% Prefix a "S" to all equations, figures, tables and reset the counter %%%%%%%%%%
\setcounter{equation}{0}
\setcounter{figure}{0}
\setcounter{table}{0}
\setcounter{page}{1}
\setcounter{section}{0}
\makeatletter
\renewcommand{\theequation}{S\arabic{equation}}
\renewcommand{\thefigure}{S\arabic{figure}}
\renewcommand{\thetable}{S\arabic{table}}
\renewcommand{\thesection}{S\arabic{section}}
%\renewcommand{\bibnumfmt}[1]{[S#1]}
%\renewcommand{\citenumfont}[1]{S#1}

%%%%%%%%%% Prefix a "S" to all equations, figures, tables and reset the counter %%%%%%%%%%
\section{Preliminaries}
A sampling strategy for a Bayesian network involves a series of marginalization and augmentation steps, with relations between distributions that can be summarized by factorizations such as
\begin{equation}
    p(x)p(y|x) = p(y)p(x|y),
    \label{eq:fact}
\end{equation}
which implies $p(x) = \int p(y)p(x|y)\d y$, a relation that can be used either to marginalize $y$ or to augment with $y$.
The first factorization we use involves the Poisson distribution.  Let
\begin{equation}
    x_j \sim \Pois(\lambda_j); \qquad  y=x_\marg{j},
    \label{eq:augpois0}
\end{equation}
where underlined indices denote summation, $x_\marg{j} := \sum_j x_j$, and we write vectors as $\{x_j\}_j$, dropping the outer index $j$ when there is no ambiguity.
Then $y$ is also Poisson distributed, and conditional on $y$ the $x_j$ have a multinomial distribution:
\begin{equation}
    y\sim\Pois(\lambda_\marg{j}); \qquad
    \{x_j\} \sim
    \Mult(y, \{\lambda_j / \lambda_\marg{j}\}).
\label{eq:augpois}
\end{equation}
This is an instance of \refeq{eq:fact} if the deterministic relationship $y=x_\marg{j}$ is interpreted as the degenerate distribution $p(y|\{x_j\})=\delta_{x_\marg{j},y}$.
Distributions hold conditional on fixed values of variables that appear on the right-hand side; for instance the distribution of $\{x_j\}$ in \refeq{eq:augpois} is conditional on both $\{\lambda_j\}$ and $y$, while  in \refeq{eq:augpois0} $x_j$ is conditioned on $\lambda_j$ only.
%Other variables are not conditioned on -- these are either conditionally independent given the fixed values, or they are marginalized.

The negative binomial distribution can be seen as an overdispersed version of the Poisson distribution, in two ways.  First, we can write it as a gamma-Poisson mixture. The joint distribution defined by
\begin{equation}
\lambda\sim\Gam(a,c);\qquad
x\sim\Pois(q \lambda),
    \label{eq:augnb1}
\end{equation}
is the same as the joint distribution defined by
\begin{equation}
    x \sim \NB(a, {q \over q+c});
    \qquad
    \lambda \sim \Gam(a+x, c+q),
    \label{eq:augnb2}
\end{equation}
also showing that the gamma distribution is a conjugate prior for the Poisson distribution. Note that we use the shape-and-rate parameterization of the gamma distribution.

The negative binomial can also be written as a Poisson-Logarithmic mixture \cite{S_journals/pami/ZhouC15}. Let $\Log(p)$ be the distribution with probability mass function
\begin{equation*}
  \Log(k;p) =  {-1\over \ln(1-p)}{p^k\over k},
\end{equation*}
where $0<p<1$, and define $n\sim\SumLog(l,p)$ by
$u_i \sim \Log(p)$ for $i=1,\ldots,l$, and $n = \sum_{i=1}^l u_i$.
Then, the joint distribution over $l$ and $n$ defined by
\begin{equation}
    n \sim \NB(a,p);\qquad
    l \sim \CRT(n,a),
        \label{eq:augnb3}
\end{equation}
is the same as
\begin{equation}
    l \sim \Pois(-a \ln(1-p));\qquad
    n \sim \SumLog(l,p),
        \label{eq:augnb4}
\end{equation}
where $\CRT$ is the Chinese restaurant table distribution
\cite{S_antoniak1974}.
This factorization allows augmenting a gamma-Poisson mixture (the negative binomial $n$) with a pure Poisson variate $l$, which is a crucial step in the deep Poisson factor analysis model. For an extension of the model we will need a similar augmentation of a Dirichlet-multinomial mixture with a pure multinomial.
It can be shown (see \ref{sec:proof} below) that
the joint distribution over $\{x_k\}$ and $\{y_k\}$ defined by
\begin{equation}
    \{x_k\}\sim\DirMult(n,\{\lambda_k\});\qquad
    y_k \sim \CRT(x_k,\lambda_k);\qquad m = y_\marg{k}
    \label{eq:dirmult1}
\end{equation}
is the same as the joint distribution over $\{x_k\}$ and $\{y_k\}$ defined by
\begin{equation}
   m\sim\CRT(n,\lambda_\marg{k});\qquad
   \{y_k\}\sim\Mult(m,\{\lambda_k\});\qquad
   \{x_k\}\sim \Polya(n, \{y_k\}),
    \label{eq:dirmult2}
\end{equation}
Here $\Polya(n, \{y_k\})$ is the distribution of the contents of an urn after running a Polya scheme (drawing a ball, returning the drawn ball and a new identically colored one each time, until the urn contains $n$ balls), where the urn initially contains $y_k$ balls of color $k$.  It is straightforward to see that
$\Polya(n, \{y_k\}) = \{y_k\} + \DirMult(n-m,\{y_k\}$).

\subsection{Proof of Dirichlet-multinomial-CRT factorization (Theorem 1; eqs.\ \refeq{eq:dirmult1}-\refeq{eq:dirmult2})}
\label{sec:proof}

A draw  from a Dirichlet-multinomial is defined by
\begin{equation*}
    \{p_j\}\sim\Dir(\{\lambda_j\});\qquad \{x_j\}\sim\Mult(n,\{p_j\});\qquad
    \{x_j\}\sim\DirMult(n,\{\lambda_j\})
\end{equation*}
By building up a draw from the multinomial as $n$ draws from a categorical distribution and using Dirichlet-multinomial conjugacy we get the Polya urn scheme,
\begin{equation*}
    \{x^{(1)}_j\}\sim\Mult(1,\{\lambda_j\});\qquad
    \{x^{(i+1)}_j\}\sim\{x^{(i)}_j\}+\Mult(1,\{\lambda_j+x^{(i)}_j\}),
\end{equation*}
where $x_j=x_j^{(n)}$ and to simplify notation we dropped the normalization of the multinomial's probability parameter.
This scheme highlights the overdispersed or "rich get richer" character of the Dirichlet-multinomial mixture distribution.

For the proof of \refeq{eq:dirmult1}-\refeq{eq:dirmult2}, recall that a draw from the Chinese Restaurant Table distribution $t\sim \CRT(n,\lambda)$ is generated by a similar scheme.  Starting with an urn containing a single special ball with weight $\lambda$, balls are drawn $n$ times, and each time the drawn ball is returned together with a new, ordinary ball of weight $1$.  The outcome $t$ is the number of times the special ball was drawn.

Now return to the Polya urn scheme above and let the initial $j$-colored balls of weight $\lambda_j$ be made of iron, let
$j$-colored balls that are added because an iron $j$-colored ball was drawn be made of oak, and let other balls be made of pine. Wooden balls have weight $1$ and all balls are drawn with probability proportional to their weight. Let $x_j$ be the final number of $j$-colored wooden balls in the urn, let $y_j$ be the final number of $j$-colored oak balls, and $m$ the final number of oak balls of any color.

Using the equivalence between the Polya urn scheme and the Dirichlet multinomial we see that $\{x_k\}$ follow a Dirichlet multinomial distribution with parameters $n$ and $\{\lambda_j\}$.  By focusing on material and ignoring color, we see that $m$ follows a CRT distribution with parameters $n$ and $\lambda_\marg{j}$. Similarly, focusing only on $j$-colored balls shows that conditional on $x_j$, $y_j$ again follows a CRT distribution, with parameters $x_j$ and $\lambda_j$, since the only events of interests are drawing a $j$-colored iron or wooden ball, which have probabilities proportional to $\lambda_j$ and $1$ respectively. Since iron balls are drawn with probability proportional to their weight, conditional on $m$ the distribution over colors among the $m$ oak balls is multinomial with parameters $m$ and $\{\lambda_j\}$. Finally, conditional on knowing the number and color of the oak balls $\{y_j\}$, the process of inserting the remaining pine balls is still a Polya process except that events involving drawing iron balls are now forbidden, so that the distribution of pine balls $\{x_j-y_j\}$ is again a Dirichlet-multinomial but with parameters $n-m$ and $\{y_j\}$.  This proves \refeq{eq:dirmult1} and \refeq{eq:dirmult2}.

\subsection{Sampling the concentration parameters of a Dirichlet distribution}
In models similar to the one considered here, the concentration parameters of a Dirichlet distribution are often kept fixed \cite{S_blei2003latent,S_zhou2012beta,S_journals/jmlr/ZhouCC16} or inferred by maximum likelihood \cite{S_minka2002}. The factorization above makes it
possible to efficiently generate posterior samples from the concentration parameters, under an appropriate prior and given multinomial observations driven by draws from the Dirichlet.  The setup is
\begin{align}
\alpha&\sim\Gam(a,b);\quad
\{\eta_k\}\sim\Dir(\{\eta^0_k\});\quad\nonumber \\
\{x_{jk}\}_k&\sim\DirMult(n_j,\{\alpha \eta_k\}_k);\quad
    y_{jk} \sim \CRT(x_{jk},\alpha \eta_k);\quad m_j = y_{j\marg{k}}.
    \label{eq:dirmult1b}
\end{align}
where we have written the concentration parameters as the product of
a probability vector $\{\eta_k\}$
and a scalar $\alpha$; these will be given Dirichlet and Gamma priors respectively.
By the factorization above this is the same joint distribution as
\begin{align}
\alpha&\sim\Gam(a,b);\quad
\{\eta_k\}\sim\Dir(\{\eta^0_k\});\quad \nonumber\\
   m_{j}&\sim\CRT(n_j,\alpha);\quad
   \{y_{jk}\}_k\sim\Mult(m_j,\{\eta_k\});\quad
   \{x_{jk}\}_k\sim\Polya(n_j, \{y_{jk}\}_k).
    \label{eq:dirmult2b}
\end{align}
The numbers $m_j$ represent the total number of distinct groups in a draw from a Dirichlet process, given the concentration parameter $\alpha$ \cite{S_hdp06}. Evidence for the value of $\alpha$ is encoded in the (unobserved) $m_j$, which in turn are determined by the (unobserved) $y_{jk}$ which sort the unobserved groups into $K$ separate subgroups conditional on the observed counts $x_{jk}$. The likelihood of the number of distinct groups $m$ in a draw from a Dirichlet process given the concentration parameter $\alpha$ and total number of draws $n$ is the probability mass function of the CRT distribution,
\begin{equation}
    p(m|\alpha,n) =
    s(n,m)\alpha^m { \Gamma(\alpha) \over \Gamma(\alpha+n)},
    \label{eq:crt1}
\end{equation}
where $s(n,m)$ are unsigned Stirling numbers of the first kind \cite{S_antoniak1974,S_hdp06}.  By multiplying
over observations $j$ a similar likelihood
is obtained for multiple observations, together with a gamma prior on $\alpha$ results in a posterior distribution that we refer to as the CRT-gamma posterior:
\begin{equation}
    \label{eq:pgcrt}
    \alpha \sim \Gam(a, b); \quad
    m_j \sim \CRT(n_j, \alpha); \quad
    \alpha \sim \GCRTP(m_\marg{j}, \{n_j\}_j, a, b),
\end{equation}
where $\GCRTP(\alpha | m, \{n_j\}_j, a, b) \propto \Gam(\alpha|a, b) \alpha^m \prod_j \frac{\Gamma(\alpha)}{\Gamma(\alpha + n_j)}$  \cite{S_hdp06}.
A sampling scheme for $\alpha$ for the likelihood \refeq{eq:crt1} and a Gamma prior was devised by \cite{S_ESCO1995}, and was extended by \cite{S_hdp06} to the case of multiple observations \refeq{eq:pgcrt}.
Finally, the multinomial distribution of $\{y_{jk}\}_k$ is conjugate to the Dirichlet prior on $\{\eta_k\}$ leading to
\begin{equation}
\{\eta_k\}\sim\Dir(\{\eta^0_k\});\qquad
   \{y_{jk}\}_k\sim\Mult(m_j,\{\eta_k\}_k);\qquad
    \{\eta_k\}\sim\Dir(\{\eta^0_k + y_{\marg{j}k}\}).
\end{equation}

\subsection{Sampling parameters of the gamma distribution}
If a Poisson-distributed observation with rate proportional to a gamma-distributed variable is available, we can use conditional conjugacy to sample the posterior of the gamma parameters.  Suppose that
\begin{equation*}
    \alpha\sim\Gam(a_0,b_0);\quad
    \beta\sim\Gam(e_0,f_0);\quad
    \theta\sim\Gam(\alpha,\beta);\quad
    m\sim\Pois(q\theta),
\end{equation*}
and that $\theta$ is not observed, but
the count $m$ is. Marginalizing $\theta$ we get $m\sim\NB(\alpha,q/(q+\beta))$.  Augmenting with
$x\sim\CRT(m,\alpha)$ and using
\refeq{eq:augnb3} and \refeq{eq:augnb4} we find that
$x\sim\Pois[\alpha\ln(1+q/\beta)]$. Using gamma-Poisson conjugacy gives the mutually dependent update equations
\begin{eqnarray}
    x\sim\CRT(m,\alpha);\quad
    \alpha\sim\Gam(a_0+x,b_0+\ln{(1+q/\beta)}).
\end{eqnarray}
As posterior for $\theta$ given $m$ we get $\theta\sim\Gam(\alpha+m,\beta+q)$; augmenting with $\theta$ and using the gamma-gamma conjugacy
\begin{equation}
    \beta\sim\Gam(e_0,f_0);\qquad
    \theta\sim\Gam(\alpha,\beta);\qquad
    \beta \sim \Gam(e_0+\alpha,f_0+\theta).
    \label{eq:gamgam}
\end{equation}
results in the mutually dependent update equations
\begin{eqnarray}
    \theta\sim\Gam(\alpha+m,\beta+q);\quad
    \beta\sim\Gam(e_0+\alpha,f_0+\theta).
\end{eqnarray}

\section{Gamma belief network}
\label{supplementary_sec:pgbn}

Since many of the techniques of the Gamma belief network of Zhou et al. \cite{S_journals/jmlr/ZhouCC16} apply to the multinomial belief network, we start by summarising and reviewing their method in some detail.  We stay close to their notation, but have made some modifications where this simplifies the future connection to the multinomial belief network.

\subsection{Backbone of feature activations}

The backbone of the model is a stack of Gamma-distributed hidden units $\theta^{(t)}_{vj}$, where the last unit parameterizes observed counts $x_{vj}$ following the Poisson distribution, one for each sample $j$ and feature $v$. The generative model is
\begin{align}
a_{vj}^{(T+1)} &= r_v, %\label{eq:thetaprior1},
 \\
\theta^{(t)}_{vj} &\sim \Gam(a_{vj}^{(t+1)}, {c_j^{(t+1)}}), && t=T,\ldots,1
\label{eq:thetaprior0}
\\
a_{vj}^{(t)} &= \sum_{k=1}^{K_{t}} \phi^{(t)}_{vk}\theta^{(t)}_{kj}, \label{eq:updateaS} && t=T,\ldots,1 \\
x_{vj} &\sim \Pois( a_{vj}^{(1)} ).
%\label{eq:poislikel}
\end{align}
For $T=1$ we only have one layer, and the model reduces to $x_{vj} = \Pois([\pmb{\phi\theta}]_{vj})$, called Poisson Factor Analysis \cite{S_zhou2012beta}. For multiple layers, the features $\pmb{\theta}^{(t+1)}$ on layer $t+1$ determine the shape parameters of the gamma distributions on layer $t$ through a connection weight matrix $\pmb{\phi}^{(t+1)}\in\mR^{K_{t}\times K_{t+1}}$, so that
$\pmb{\phi}^{(t+1)}$ induces correlations between features on level $t$.
The lowest-level activations $\pmb{a}^{(1)}$ are used to parameterize a Poisson distribution, which generates the observed count variables $x_{vj}$ for individual (document, observation) $j$.
Below we will treat $r_v$ and $c_j^{(t)}$ as random variables and targets for inference, but for now, we consider them as fixed parameters and focus on inference of $\pmb{\phi}^{(t)}$ and $\pmb{\theta}^{(t)}_j$.  We will assume that $\sum_v \phi^{(t)}_{vk}=1$, which later on is enforced by Dirichlet priors on $\pmb{\phi}_k$.

This model architecture is similar to a $T$-layer neural network, with $\pmb{a}^{(t)}$ playing the role of activations that represent the activity of features (topics, factors) of increasing complexity as $t$ increases. In the remainder, we use the language of topic models, so that $x_{vj}$ is the number of times word $v$ is used in document $j$, and $\phi^{(1)}_{vk}$ is the probability that word $v$ occurs in topic $k$. This is for the lowest level $1$; we will similarly refer to level-$t$ topics and level-$t$ ``words'', the latter representing the activity of corresponding topics on level $t-1$.

Different from \cite{S_journals/jmlr/ZhouCC16} we use a Gamma-distributed variate $c_j^{(2)}$ as rate parameter of the gamma distribution for $\theta^{(1)}$, instead of $p_j^{(2)}/1-p_j^{(2)}$ where $p_j^{(2)}$ has a Beta distribution; we will come back to this choice below.

\subsection{Augmentation with latent counts}
\label{sec:dpfa}

We review Zhou's augmentation and marginalization scheme that enables efficient inference for this model.
First, introduce new variables
\begin{equation}
    x^{(t)}_{vj}\sim \Pois(q_j^{(t)}a^{(t)}_{vj}),
    \qquad t=1,\ldots,T+1
    \label{eq:q}
\end{equation}
where we set $q_j^{(1)}=1$ so that we can identify $x^{(1)}_{vj}$ with the observed counts $x_{vj}$; the $q_j^{(t)}$ for $t>1$ will be defined below. Using \refeq{eq:augpois0}-\refeq{eq:augpois} we can augment $x^{(t)}_{vj}$ as
\begin{align}
\label{eq:samp30}
  y^{(t)}_{vjk} &\sim \Pois(q^{(t)}_j \phi^{(t)}_{vk}\theta^{(t)}_{kj});\\
x^{(t)}_{vj} &= y^{(t)}_{vj\underline{k}}. \label{eq:samp3}
\end{align}
The counts $y_{vjk}^{(t)}$ represent a possible assignment of level-$t$ words to level-$t$ topics.
Marginalizing over $v$ and using \refeq{eq:augpois0}-\refeq{eq:augpois} again we get the augmentation
\begin{align}
    m_{jk}^{(t)}&:=y_{\underline{v}jk}^{(t)}
    \sim \Pois(q_j^{(t)}\theta^{(t)}_{kj});
    \label{eq:mjkt} \\
    y_{vjk} &= \Mult(m^{(t)}_{jk}, \{\phi^{(t)}_{vk}\}_v ). \label{eq:samp2}
\end{align}
since $\phi^{(t)}_{\underline{v}k}=1$.
These counts represent level-$t$ topic usage in document $j$.
Now, marginalizing $\pmb{\theta}^{(t)}$ turns $m_{jk}^{(t)}$ into an overdispersed Poisson distribution; from \refeq{eq:thetaprior0} we see that $\pmb{\theta}^{(t)}$ is Gamma distributed, so it becomes a negative binomial:
\begin{align}
    m_{jk}^{(t)}&\sim \NB(a^{(t+1)}_{kj},q^{(t)}_j/(q^{(t)}_j+c^{(t+1)}_j)),
    \label{eq:sampnb}
\end{align}
using \refeq{eq:augnb1} and \refeq{eq:augnb2}.
This gives us a count variable that is parameterized by the activation of the layer above $t$, but one which follows a negative binomial distribution rather than a Poisson distribution \refeq{eq:q}.  However, using \refeq{eq:augnb3} and \refeq{eq:augnb4} we can augment once more to write the negative binomial as a Poisson-Logarithmic mixture:
\begin{equation}
    x_{kj}^{(t+1)} \sim \Pois(a_{kj}^{(t+1)} \ln {q^{(t)}_j+c_j^{(t+1)} \over c_j^{(t+1)}});\qquad
    m_{jk}^{(t)} \sim \SumLog(x_{kj}^{(t+1)}, {q^{(t)}_j \over q^{(t)}_j+c_j^{(t+1)}}),
    \label{eq:samp1}
\end{equation}
so that $x^{(t+1)}_{kj}$ agrees with \refeq{eq:q} if we choose
\begin{equation}
    q_j^{(t+1)} := \ln
    {q_j^{(t)} + c_j^{(t+1)} \over c_j^{(t+1)}}.
    \label{eq:defq}
\end{equation}
This allows us to continue the procedure for layer $y+1$, and so on until $t=T$, sampling augmented variables $y^{(t)}_{vjk}$, $m^{(t)}_{jk}$ and $x^{(t+1)}_{kj}$ for $t=1,\ldots,T$.

\subsection{Alternative representation as Deep Poisson Factor model}
\label{sec:altdpfa}

The procedure described in section \ref{sec:dpfa} not only augments the model with new counts but also integrates out $\pmb{\theta}^{(t)}$.  This provides an alternative and equivalent representation as a generative model. Starting from $a^{(t+1)}_{kj}$ we can
use \refeq{eq:samp1}, \refeq{eq:samp2} and \refeq{eq:samp3} to sample $x^{(t+1)}_{kj}$,
$m^{(t)}_{jk}$, $y_{vjk}^{(t)}$,
and finally $x_{vj}^{(t)}$. Continuing downwards this shows how to eventually generate the observed counts $x^{(1)}_{vj}$ using count variables, while $\pmb{\theta}^{(t)}$ is integrated out. Explicitly, the generative model becomes
\begin{align*}
&x_{kj}^{(t+1)}\sim\Pois(q_j^{(t+1)}a_{kj}^{(t+1)});\\ &m_{jk}^{(t)}\sim\SumLog(x_{kj}^{(t+1)},1-e^{-q_j^{(t+1)}});\qquad
\{y_{vjk}^{(t)}\}_v\sim\Mult(m_{jk}^{(t)},\{\phi_{vk}^{(t)}\}_v);\qquad
x_{vj}^{(t)}:=y_{vj\marg{k}}^{(t)}.
\end{align*}
(Note that throughout we condition on $q^{(t)}_j$ for all $t$, and therefore on all $c_j^{(t)}$ as well; we also haven't specified how to sample $\phi^{(t)}_{vk}$ yet.) This equivalent generative process motivates the name Deep Poisson Factor Analysis. The two alternative schemes are shown graphically in figure \ref{supplementary_fig:equiv}.

\begin{figure}
\begin{tikzpicture}[node distance=1.5cm,>=latex]

% Nodes for a^3, theta^2, a^2, theta^1, a^1, and phi^1
\node (a3) {$\pmb{a}^{(3)}$};
\node (theta2) [below of=a3] {$\pmb{\theta}^{(2)}$};
\draw [->] (a3) -- (theta2);
\node (a2) [below of=theta2] {$\pmb{a}^{(2)}$};
\draw [-|] (theta2) -- (a2);
\node (theta1) [below of=a2] {$\pmb{\theta}^{(1)}$};
\draw [->] (a2) -- (theta1);
\node (a1) [below of=theta1] {$\pmb{a}^{(1)}$};
\draw [-|] (theta1) -- (a1);
\node (phi1) [left of=a1] {$\pmb{\phi}^{(1)}$};
\draw [-|] (phi1) -- (a1);

% Node for c^3 and arrow to theta^2
\node (c3) [left of=theta2] {$\pmb{c}^{(3)}$};
\draw [->] (c3) -- (theta2);

% Node for c^2 and arrow to theta^1
\node (c2) [left of=theta1] {$\pmb{c}^{(2)}$};
\draw [->] (c2) -- (theta1);

% Nodes for x^i, q^i, and arrows to them
\node (xi3) [text=gray,right of=a3] {$\pmb{x}^{(3)}$};
\node (q3) [text=gray,right of=xi3] {$\pmb{q}^{(3)}$};
\node (xi2) [text=gray,right of=a2] {$\pmb{x}^{(2)}$};
\node (q2) [text=gray,right of=xi2] {$\pmb{q}^{(2)}$};
\node (xi1) [right of=a1] {$\pmb{x}^{(1)}$};
\node (q1) [right of=xi1] {$\pmb{q}^{(1)}$};
\draw [->,gray] (a3) -- (xi3);
\draw [->,gray] (a2) -- (xi2);
\draw [->] (a1) -- (xi1);
\draw [-|,gray] (q1) -- (q2);
\draw [-|,gray] (c2) -- (q2);
\draw [-|,gray] (q2) -- (q3);
\draw [-|,gray] (c3) -- (q3);

\draw [->,gray] (q3) -- (xi3);
\draw [->,gray] (q2) -- (xi2);
\draw [->] (q1) -- (xi1);

% Node for phi^2 and arrow to a^2
\node (phi2) [left of=a2] {$\pmb{\phi}^{(2)}$};
\draw [-|] (phi2) -- (a2);

% Node for phi^3 and arrow to a^3
\node (phi3) [left of=a3] {};

\node (labelA) [left of=phi3] {\bf a.};
\node (labelB) [right of=q3] {\bf b.};
\node (cempty) [right of=labelB] {};
\node (ca3) [right=0.25 cm of cempty] {$\pmb{a}^{(3)}$};
\node (cx3) [right of=ca3] {$\pmb{x}^{(3)}$};
\node (cq3) [right of=cx3] {$\pmb{q}^{(3)}$};
\node (cm2) [below=0.42 cm of cx3] {$\pmb{m}^{(2)}$};
\node (cy2) [below=0.42 cm of cm2] {$\pmb{y}^{(2)}$};
\node (cx2) [below=0.42 cm of cy2] {$\pmb{x}^{(2)}$};
\node (cm1) [below=0.42 cm of cx2] {$\pmb{m}^{(1)}$};
\node (cy1) [below=0.42 cm of cm1] {$\pmb{y}^{(1)}$};
\node (cx1) [below=0.42 cm of cy1] {$\pmb{x}^{(1)}$};
\node (cc3) at  (cm2 -| cempty) {$\pmb{c}^{(3)}$};
\node (cc2) at  (cm1 -| cempty) {$\pmb{c}^{(2)}$};
\node (cphi2) at  (cy2 -| cempty) {$\pmb{\phi}^{(2)}$};
\node (cphi1) at  (cy1 -| cempty) {$\pmb{\phi}^{(1)}$};
\node (cq2) at  (q2 -| cq3) {$\pmb{q}^{(2)}$};
\node (cq1) at  (q1 -| cq3) {$\pmb{q}^{(1)}$};
\draw [->] (ca3) -- (cx3);
\draw [->] (cq3) -- (cx3);
\draw [->] (cx3) -- (cm2);
\draw [->] (cm2) -- (cy2);
\draw [-|] (cy2) -- (cx2);
\draw [->] (cx2) -- (cm1);
\draw [->] (cm1) -- (cy1);
\draw [-|] (cy1) -- (cx1);
\draw [-|] (cc3) -- (cq3);
\draw [-|] (cc2) -- (cq2);
\draw [->] (cphi2) -- (cy2);
\draw [->] (cphi1) -- (cy1);
\draw [-|] (cq1) -- (cq2);
\draw [-|] (cq2) -- (cq3);
\draw [->] (cq3) -- (cm2);
\draw [->] (cq2) -- (cm1);
\end{tikzpicture}
\caption{Two equivalent generative models for a count variable $\pmb{x}^{(1)}$ from the Poisson gamma belief network, using ({\bf a}) a tower of real-valued latent variables $\pmb{\theta}$, $\pmb{a}$,  or ({\bf b}) latent counts $\pmb{m}$, $\pmb{y}$, $\pmb{x}$.  Blunt arrows indicate deterministic relationships.  The variable $\pmb{q}^{(1)}$ is a dummy and has a fixed value $1$.  The counts $\pmb{x}^{(t)}$ and variables $\pmb{q}^{(t)}$, $t>1$, in the left representation, are included for clarity (and have the same distribution as the variables in the right model) but are not used to generate the outcome $\pmb{x}^{(1)}$, and so can be marginalized out.}
\label{supplementary_fig:equiv}
\end{figure}

\subsection{Sampling per-document latent variables}

The derivation above can be used to sample the latent counts conditional on observations (and parameters $\pmb{\phi}$, $\pmb{\theta}$ and $\pmb{r}$), from layer 1 upwards.  These steps are,
\begin{align}
\label{eq:cond1}
y_{vjk}^{(t)} & \sim \Mult(x^{(t)}_{vj},\{\phi^{(t)}_{vk}\theta_{kj}^{(t)}
%/ \textstyle\sum_k\phi^{(t)}_{vk}\theta_{kj}^{(t)}  % Ik gebruik de conventie dat de parameters van \Mult niet genormaliseerd hoeven te worden - dat scheelt een hoop gepriegel in de formules
\}_k);
\\
\label{eq:cond2}
m^{(t)}_{jk} &= y^{(t)}_{\marg{v}jk}; \\
\label{eq:cond3}
x_{kj}^{(t+1)} &\sim \CRT(m_{jk}^{(t)}, a_{kj}^{(t+1)}),
\end{align}
where for \refeq{eq:cond1} we used \refeq{eq:augpois0}-\refeq{eq:augpois} and \refeq{eq:samp30}-\refeq{eq:samp3}; and for
\refeq{eq:cond3} we used \refeq{eq:augnb3}-\refeq{eq:augnb4} and \refeq{eq:sampnb}-\refeq{eq:samp1}.  Note that
after the last step $x_{kj}^{(t+1)}$ is no longer conditioned on $\theta^{(t)}_{kj}$ because it is integrated out, however, explicit values for
$\theta^{(t)}_{kj}$ \emph{are} used in \refeq{eq:cond1}.
To sample new values for $\theta^{(t)}_{kj}$, we use Gamma-Poisson conjugacy \refeq{eq:augnb1}-\refeq{eq:augnb2} on
\refeq{eq:thetaprior0} and \refeq{eq:mjkt} to get
\begin{equation}
    \theta^{(t)}_{kj}\sim
    \Gam(a^{(t+1)}_{kj} + m^{(t)}_{kj},
    c^{(t+1)}_j + q_j^{(t)}).
    \label{eq:theta}
\end{equation}
In order to sample the final set of per-document variables, the inverse scaling parameters $c_j^{(t)}$,
we first need to integrate out $q_j^{(t)}$ since it
depends on $c_j^{(t)}$ via \refeq{eq:defq}. That
means we also need to integrate out $x_{vj}^{(t)}$ and
$m^{(t-1)}_{jk}$ which both depend on $q_j^{(t)}$, as well as $y^{(t-1)}_{vjk}$  because of the deterministic relationship
\refeq{eq:mjkt}.  We do not need to marginalize other variables as $\pmb{x}^{(t-1)}$ and its dependents
are conditionally independent of $\pmb{x}^{(t)}$ given $\pmb{\theta}^{(t-1)}$,
as can be seen from figure \ref{fig:equiv}a.
Once $\pmb{x}^{(t)}$, $\pmb{m}^{(t-1)}$ and $\pmb{y}^{(t-1)}$ are
integrated out, $\theta_{vj}^{(t-1)}$ is related to $c_j^{(t)}$ solely through \refeq{eq:thetaprior0}, and marginalizing this over $v$ we get
\begin{equation*}
    \theta^{(t-1)}_{\marg{v}j}\sim
    \Gam(a^{(t)}_{\marg{v}j}, c_j^{(t)}),\qquad
    (t=2,\ldots,T+1).
\end{equation*}
where $a^{(t)}_{\marg{v}j}=\theta_{\marg{k}j}^{(t)}$ for $t=2,\ldots,T$, and $a^{(T+1)}_{\marg{v}j} = r_{\marg{v}}$.
The conjugate prior for a gamma likelihood with fixed shape parameter is a gamma distribution again.  This gives the following prior and posterior distributions for $c_j^{(t)}$:
\begin{align}
    c_j^{(t)}\sim \Gam(e_0,f_0);\qquad
    c_j^{(t)}\sim \Gam(e_0+a^{(t)}_{\marg{v}j}, f_0 + \theta^{(t-1)}_{\marg{k}j}). \qquad
    (t=2,\ldots,T+1)
    \label{eq:sampc}
\end{align}
As an aside, note that it is possible to integrate out $\pmb{\phi}^{(1)}$ and $\pmb{\theta}^{(1)}$, in the same way as is done in the collapsed Gibbs sampler for the LDA model.  This is done in \cite{S_journals/jmlr/ZhouCC16} and may lead to faster mixing. However, if we do that no $\pmb{\theta}^{(1)}$ is available to construct a posterior for $\pmb{c}^{(2)}$ as in \refeq{eq:sampc}.  Instead, \refeq{eq:sampnb} can be used:
\begin{equation*}
    m_{jk}^{(1)}\sim \NB(a^{(2)}_{kj},1/(1+c_j^{(2)})),
\end{equation*}
using that $q_j^{(1)}=1$. Using the beta-negative binomial conjugacy, we can write
\begin{align*}
    p^{(2)} \sim\Beta(a_0,b_0); \qquad p^{(2)} \sim \Beta(a_0 +  a^{(2)}_{\marg{k}j}, b_0 + m_{j\marg{k}}^{(1)}),\end{align*}
where we defined
\begin{equation*}
    p^{(2)} := (1+c_j^{(2)})^{-1},    \qquad\text{so that}\qquad c_j^{(2)} = (1-p^{(2)})/p^{(2)},
\end{equation*}
which gives $\pmb{c}^{(2)}$ a Beta distribution of the second kind. We cannot similarly integrate out $\pmb{\theta}^{(t)}$ and $\pmb{\phi}^{(t)}$ for $t>1$ as already for collapsed Gibbs sampling, values for $\pmb{\theta}^{(2)}$ and $\pmb{\phi}^{(2)}$ are necessary to define the prior on $\pmb{\theta}^{(1)}$.
%{\bf Could we not do something similar to HDP here?}

\subsection{Sampling
\mathinhead{\phi}{phi}
and
\mathinhead{r}{r}}

It remains to sample the model-level parameters $\phi^{(t)}_{vk}$, $\eta^{(t)}_v$, and $r_v$.
For $\pmb{\phi}$ we marginalize \refeq{eq:samp2} over $j$ and, noting that the multinomial is conjugate to a Dirichlet, we use a Dirichlet prior for $\pmb{\phi}$ to obtain the update equations
\begin{equation}
\{\phi^{(t)}_{vk}\}_v \sim \Dir(\{\eta^{(t)}_v\}_v);\qquad
    \{y_{v\marg{j}k}^{(t)}\}_v\sim
\Mult(m_{\marg{j}k}^{(t)}, \{\phi^{(t)}_{vk}\}_v);
\qquad
\{\phi^{(t)}_{vk}\}_v \sim \Dir(\{\eta^{(t)}_v
+y^{(t)}_{v\marg{j}k}\}_v).
\label{eq:sampphi}
\end{equation}

To sample $\{\eta^{(t)}_v\}$, we integrate out $\phi^{(t)}_{vk}$ and consider $x_{v\marg{j}k}^{(t)}$ as a draw from a Dirichlet-multinomial distribution with parameters $\alpha^{(t)}\eta_v^{(t)}$ where the factors $\alpha^{(t)}$ and $\{\eta_v^{(t)}\}$ have Gamma and Dirichlet priors respectively, as in \refeq{eq:dirmult1b}.  This results in the following  update equations,
\begin{align}
    \alpha^{(t)}&\sim\Gam(a,b);\quad
    \{\eta^{(t)}_v\}_v \sim\Dir(\{\eta_{v,0}\}_v);\quad
    \{y^{(t)}_{v\marg{j}k}\}_v \sim \DirMult(m_{\marg{j}k}^{(t)}, \{\alpha^{(t)} \eta_v^{t}\}_v);\quad \nonumber \\
    z^{(t)}_{vk}&\sim\CRT(y^{(t)}_{v\marg{j}k},\alpha^{(t)}\eta^{(t)}_v);\quad
    \alpha\sim\GCRTP(z^{(t)}_{\marg{v}\marg{k}},
    \{m_{\marg{j}k}^{(t)}\}_k,a,b);\quad
    \{\eta^{(t)}_{v}\}_v \sim\Dir(\{\eta_{v,0}+z^{(t)}_{v\marg{k}}\}_v).
    \label{eq:phigcrtp}
\end{align}
%Note that if we choose $s_{v,0}^{(t)}=b$ in \refeq{eq:gscp} and $\eta_{v,0}=1$ and $a=\eta_{\marg{v},0}=V$ in \refeq{eq:phigcrtp} then the priors on $\phi_{vk}^{(t)}$ coincide and the two approaches are equivalent, although the collapsed sampler \refeq{eq:phigcrtp} will probably mix better since $\phi_{vk}^{(t)}$ is integrated out.

For $r_v$ we use \refeq{eq:samp1} and \refeq{eq:defq} for $t=T$, marginalize  $j$, and use gamma-Poisson conjugacy to get update equations
\begin{equation}
    r_v\sim\Gam(\gamma_0/K_T,c_0);\qquad
    x^{(T+1)}_{v\marg{j}} \sim\Pois(r_vq_{\marg{j}}^{(T+1)});\qquad
   r_v\sim\Gam(\gamma_0/K_T+x^{(T+1)}_{v\marg{j}},
   c_0+q_{\marg{j}}^{(T+1)})
   \label{eq:r}
\end{equation}

\subsection{Sampling strategy}

It is helpful to think of the model as organised as an alternating stack of layers, one taking inputs $\pmb{a}$ and $\pmb{c}$ and using the gamma distribution to produce an output $\pmb{\theta}$; and one taking $\pmb{\theta}$ and edge weights $\pmb{\phi}$ to produce an activation $\pmb{a}$.  During inference, the model also uses latent variables $x_{vj}$, $y_{vjk}$, $m_{jk}$ and $q_j$.  Inference proceeds in two main stages.  First, the $\pmb{q}^{(t)}$ are calculated, followed by augmentation with the $\pmb{y}$, $\pmb{m}$ and $\pmb{x}^{(t)}$ $(t\geq 2)$ count variables going up the stack, while marginalising $\pmb{\phi}$, and also updating the $\pmb{\phi}$ variables.
After updating the parameter $\pmb{r}$, the second stage involves updating $\pmb{c}$ and $\pmb{\theta}$ going down the stack, while the augmented count variables are dropped again.  Table \ref{tbl:sampling} provides a detailed overview.

\begin{table}
    \centering
    \resizebox{\columnwidth}{!}{
    \def\sa{\hspace{1.1cm}}
    \def\sb{\hspace{0.8cm}}
    \def\S{\bf S}
    \def\U{\bf U}
    \def\bx{{\fboxsep=.1cm\fbox{x}}}
    \def\bone{{\fboxsep=.1cm\fbox{\rlap{1}\phantom{x}}}}
    \def\bo{{\fboxsep=.05cm\fbox{O}}}
    \def\bmin{{\fboxsep=.15cm\fbox{\rlap{-}\phantom{x}}}}
    \setlength{\tabcolsep}{3pt}
    \begin{tabular}{l|l|l|ccccccccccccccccccc}
    &&& \multicolumn{7}{c}{ \ovalbox{\sa  factor-layer $1$ \sa}} &
    & \multicolumn{7}{c}{\ovalbox{\sa factor-layer $2$ \sa}} \\
    &&& \multicolumn{5}{c}{} &
      \multicolumn{6}{c}{\ovalbox{\sb gamma layer $1$ \sb}} &&&
            \multicolumn{6}{c}{\ovalbox{\sb gamma layer $2$ \sb}} \\
      \hline
Stage & Eq. & (Eq.) & $x^{1}$ & $q^1$ & $a^1$ & $\phi^1$ & $y^1$ & $\theta^1$ & $m^1$ & $c^2$ & $x^2$ & $q^2$ & $a^2$ & $\phi^2$ & $y^2$ & $\theta^2$ & $m^2$ & $c^3$ & $x^3$ & $q^3$ & $r$ \\
 \hline & & & \\
 ~~  &    &                & O   & 1     & x     & x   & -   & x   & -     & x   & -     & -   & x     & x   & -     & x   & -     & x   & -     & -     & x \\
  0 (all $t$)  & \refeq{eq:defq} & \refeq{eq:n_upward}                   & O   & \bone     & x     & x   & -   & x   & -     & \bx   & -     & \U   & x     & x   & -     & x   & -     & \bx   & -     & \U     & x \\
 1 ($t$=1)  & \refeq{eq:cond1} & \refeq{eq:sample_my}  & \bo & 1     & x     & \bx & \S   & \bx & \bmin & x   & -     & x   & x     & x   & -     & x   & -     & x   & -     & x     & x \\
 2 ($t$=1)  & \refeq{eq:sampphi} & \refeq{eq:sampphi} & O   & 1     & \bmin & \S   & \bx & x   & -     & x   & -     & x   & x     & x   & -     & x   & -     & x   & -     & x     & x \\
 3 ($t$=1)  & \refeq{eq:cond2} & \refeq{eq:mm}   & O   & 1     & -     & x   & \bx & x   & \U     & x   & -     & x   & x     & x   & -     & x   & -     & x   & -     & x     & x \\
 4 ($t$=1)  & \refeq{eq:cond3} & \refeq{eq:x_upward}   & O   & 1     & -     & x   & x   & \bmin   & \bx   & x   & \S     & x   & \bx   & x   & \bmin & x   & -     & x   & -     & x     & x \\
 1 ($t$=2)  & \refeq{eq:cond1} & \refeq{eq:sample_my}  & O   & 1     & -     & x   & x   & -   & x     & x   & \bx   & x   & x     & \bx & \S     & \bx & \bmin & x   & -     & x     & x \\
 2 ($t$=2)  & \refeq{eq:sampphi} & \refeq{eq:sampphi} & O   & 1     & -     & x   & x   & -   & x     & x   & x     & x   & \bmin & \S   & \bx   & x   & -     & x   & -     & x     & x \\
 3 ($t$=2)  & \refeq{eq:cond2} & \refeq{eq:mm}   & O   & 1     & -     & x   & x   & -   & x     & x   & x     & x   & -     & x   & \bx   & x   & \U     & x   & -     & x     & x \\
 4 ($t$=2)  & \refeq{eq:cond3} & \refeq{eq:x_upward}   & O   & 1     & -     & x   & x   & -   & x     & x   & x     & x   & -     & x   & x     & \bmin   & \bx   & x   & \S     & x     & \bx \\
 r        & \refeq{eq:r} & \refeq{eq:sample_mr}       & O   & 1     & -     & x   & x   & -   & x     & x   & x     & x   & -     & x   & x     & -   & x     & x   & \bx   & \bx   & \S \\
 5 ($t$=2)  & \refeq{eq:theta} & \refeq{eq:sample_theta}   & O   & 1     & -     & x   & x   & -   & x     & x   & x     & \bx & \bmin     & x   & x     & \S   & \bx   & \bx & x     & x     & \bx \\
 6 ($t$=2)  & \refeq{eq:sampc} & \refeq{eq:sample_c}   & O   & 1     & -     & x   & x   & -   & x     & x   & x     & x   & -     & x   & \bmin & \bx & \bmin & \S   & \bmin & \bmin & \bx \\
 7 ($t$=2)  & \refeq{eq:updateaS} & \refeq{eq:dmfaa2}   & O   & 1     & -     & x   & x   & -   & x     & x   & x     & x   & \U    & \bx   & - & \bx & - & x   & - & - & x \\

 %7 (t=2)  & \refeq{eq:defq}    & O   & 1     & -     & x   & x   & -   & x     & x   & x     & \bx & x     & x   & x     & x   & x     & \bx & -     & \U     & x \\
 5 ($t$=1)  & \refeq{eq:theta} & \refeq{eq:sample_theta}   & O   & \bone & \bmin     & x   & x   & \S   & \bx   & \bx & x     & x   & \bx   & x   & -     & x   & -     & x   & -     & -     & x \\
 6 ($t$=1)  & \refeq{eq:sampc} & \refeq{eq:sample_c}  & O   & 1     & -     & x   & \bmin   & \bx & \bmin     & \S   & \bmin & \bmin   & \bx   & x   & \bmin & x   & \bmin & x   & \bmin & \bmin     & x \\
 7 ($t$=1)  & \refeq{eq:updateaS} & \refeq{eq:dmfaa2}   & O   & 1     & \U     & \bx   & -   & \bx & -     & x   & - & -   & x   & x   & - & x   & - & x   & - & -     & x \\
    \end{tabular}
    }  % Closing bracket resize box.
    \caption{Variable instantiation and marginalisation during inference. \rm Upper indices denote layer number (without parentheses). Lower indices are suppressed.  All variables include the observation index $j$ except for $r$ and $\phi$.  The symbols $1$, $-$, O, x, S, and U denote the value 1; marginalized; observed; instantiated; sampled; and deterministically updated respectively. Boxed symbols denote dependencies (either on an instantiated value, or on the corresponding variable having been marginalized). Inference of $\eta^{(t)}$, the parameter of the Dirichlet prior for $\phi^{(t)}$,  not shown.  Equation numbers refer to Gamma-Poisson model (left column) and multinomial belief network (right column; replace columns labeled "$q$" with "$n$"). }
    \label{tbl:sampling}
    \end{table}

\section{Multinomial observables}
\label{supplementary_sec:multinomial}

\subsection{Deep Multinomial Factor Analysis}\label{supplementary_sec:multinomial_factorisation}
To model multinomial observations, we replace the Poisson observables with a multinomial and, for each sample $j$, we swap out the gamma-distributed hidden activations for Dirichlet samples $\{ \theta^{(t)}_{vj} \}_v$.
The generative model is
\begin{align}
    a_{vj}^{(T+1)} &= r_v, \\
    \{\theta^{(t)}_{vj}\}_v &\sim \Dir(\{c^{(t+1)} a_{vj}^{(t+1)}\}_v), && t=T,\ldots,1 \label{eq:dmfatheta} \\
%    c^{(t)} & \sim  \Gam(e_0, f_0), && t=T,\ldots,2\\
    a_{vj}^{(t)} &= \sum_{k=1}^{K_{t}} \phi^{(t)}_{vk}\theta^{(t)}_{kj}, && t=T,\ldots,1 \label{eq:dmfaa2} \\
    \{x_{vj}\}_v &\sim \Mult( n_j, \{a_{vj}^{(1)}\}_v ).
    %\label{eq:dmfamult}
\end{align}

%In \refeq{eq:dmfamult} we implicitly normalize the probability parameters of the multinomial.  Because of this, the scale of $a^{(1)}_{vj}$ is irrelevant, so that $c^{(1)}$ is also irrelevant and is arbitrarily fixed to $1$.
Missing from this model definition are the specification of the prior distributions of $r_v$ and $c^{(t)}$; these are introduced in section \ref{sec:multsamp} but are considered fixed in this section.
The variables $c^{(t)}$, $t=2,\ldots,T+1$ set the scale of the Dirichlet's concentration parameters which modulate the variance of $\theta^{(t-1)}_{vj}$ across documents $j$. Different from the PGBN model we choose one $c^{(t)}$ per dataset instead of one per sample $j$, reducing the number of free parameters per sample, and allowing the variance across samples to inform the $c^{(t)}$.
Similar to the PGBN, the first step towards a posterior sampling procedure involves augmentation and marginalization.  First, we introduce new variables
\begin{align}
\{x^{(t)}_{vj}\}_v&\sim \Mult(n^{(t)}_j,\{a^{(t)}_{vj}\}_v),\qquad t=1,\ldots,T+1
\label{eq:defx}
\end{align}
where we set $n^{(1)}_j:=n_j$ and we identify $x_{vj}$ with $x^{(1)}_{vj}$; below we define $n^{(t)}_j$ for $t>1$.  We can augment $x_{vj}^{(t)}$ as
\begin{align}
    \{y_{vjk}^{(t)}\}_{vk}&\sim\Mult(n^{(t)}_j, \{\phi^{(t)}_{vk}\theta^{(t)}_{kj}\}_{vk});\qquad
    x_{vj}^{(t)}=y^{(t)}_{vj\marg{k}};
    \label{eq:chi_upward}
\end{align}
%so that conditional on $x_{vj}^{(t)}$ we have
%\begin{equation}
%$    \{y_{vjk}^{(t)}\}_{k}\sim\Mult(x^{(t)}_{vj},\{ \phi^{(t)}_{vk}\theta^{(t)}_{kj}\}_{k}  \}$.
%\end{equation}
Marginalizing $y^{(t)}_{vjk}$ over $v$ results in the augmentation
\begin{align}
    \{m_{jk}^{(t)} \}_k &:=y_{\marg{v}jk}^{(t)}\sim
    \Mult(n_j^{(t)}, \{\theta_{kj}^{(t)}\}_k);
        \label{eq:multm}
    \\
    \{y_{vjk}^{(t)}\}_v &\sim
    \Mult(m_{jk}^{(t)}, \{\phi_{vk}^{(t)}\}_v).
    \label{eq:multphi}
\end{align}
where \refeq{eq:multm} holds since $\phi_{\marg{v}k}^{(t)}=1$.
%so that conditional on $m_{jk}^{(t)}$ we have
Now marginalizing over $\theta^{(t)}_{kj}$ in \refeq{eq:multm} results in a
Dirichlet-multinomial,
an overdispersed multinomial that plays a role similar to the negative binomial as an overdispersed Poisson for the PGBN:
\begin{equation}
    \{m_{jk}^{(t)}\}_k \sim \DirMult(n^{(t)}_j, \{c^{(t+1)}a^{(t+1)}_{kj}\}_k).
    \label{eq:augmult1}
\end{equation}
To augment this overdispersed multinomial with a pure multinomial, so that we can continue the augmentation in the layer above, we use \refeq{eq:dirmult1}--\refeq{eq:dirmult2}:
\begin{align}
    n^{(t+1)}_j&\sim \CRT(n^{(t)}_j,c^{(t+1)});\qquad
     \label{eq:ntplus1}
   \\
    \{x^{(t+1)}_{kj}\}_k&\sim \Mult(n^{(t+1)}_j, \{a^{(t+1)}_{kj}\}_{k});\qquad
    \\
    \{m^{(t)}_{jk}\}_{k}&\sim
    \Polya(n_j^{(t)}, \{x^{(t+1)}_{kj}\}_k),
    \label{eq:ntplus3}
\end{align}
where in the first line we used that
$a^{(t+1)}_{\marg{k}j}=1$, because $\phi_{\marg{v}k}^{(t+1)} =\theta_{\marg{k}j}^{(t+1)}=1$.  Equation \refeq{eq:ntplus1} recursively defines the distribution of the scale counts $n_j^{(t)}$ for $t>1$, which play a role analogous to the $q_j^{(t)}$ in the PGBN; these counts depend only on $n_j$ and are independent of the observations $x^{(1)}_{vj}$ that we condition on, depending only on the $c^{(t')}$, $t'\leq t$. Continuing this procedure results in augmented counts $y^{(t)}_{vjk}$, $m^{(t)}_{jk}$ and $x^{(t+1)}_{kj}$ for $t=1,\ldots,T$.

With $\theta^{(t)}$ integrated out, the emerging alternative generative model representation is a deep multinomial factor model, as follows:
\begin{align}
    n_j^{(t+1)}&\sim\CRT(n_j^{(t)},c^{(t+1)});\qquad
    \{x_{kj}^{(t+1)}\}_k \sim\Mult(n_j^{(t+1)}, \{a^{(t+1)}_{kj}\}_k);\\
    \{m^{(t)}_{kj}\}_k &\sim \Polya(n_j^{(t)}, \{x_{kj}^{(t+1)}\}_k);\qquad
    \{y^{(t)}_{vjk}\} \sim\Mult(m_{jk}^{(t)},\{\phi^{(t)}_{vk}\}_v);\qquad
    x^{(t)}_{vj} = y^{(t)}_{vj\marg{k}}.
\end{align}
The two representations of the model are structurally identical to the two representations of the PGBN shown in figure \ref{supplementary_fig:equiv}, except that $q^{(t)}$ are replaced by $n^{(t)}$.

\subsection{Sampling posterior variables}
\label{sec:multsamp}

To sample counts conditional on observations we use
\begin{align}
%\label{eq:cond1m}
\{y_{vjk}^{(t)}\}_k & \sim \Mult(x^{(t)}_{vj},\{\phi^{(t)}_{vk}\theta_{kj}^{(t)}  \}_k);
\label{eq:sample_my}
\\
%\label{eq:cond2m}
m^{(t)}_{jk} &= y^{(t)}_{\marg{v}jk};
\label{eq:mm}
\\
x_{kj}^{(t+1)} &\sim \CRT(m_{jk}^{(t)}, c^{(t+1)} a_{kj}^{(t+1)});
\label{eq:x_upward} \\
n^{(t+1)}_j &= x_{\marg{k}j}^{(t+1)}
\label{eq:n_upward}
\end{align}
similar to \refeq{eq:cond1}-\refeq{eq:cond2}; for \refeq{eq:x_upward}  we used \refeq{eq:augmult1}-\refeq{eq:ntplus3}
and \refeq{eq:dirmult1}--\refeq{eq:dirmult2}.
To sample $\theta$, use \refeq{eq:multm}, \refeq{eq:dmfatheta} and Dirichlet-multinomial conjugacy to get
\begin{equation}
    \{\theta_{kj}^{(t)}\}_k \sim \Dir(\{c^{(t+1)}a_{kj}^{(t+1)}+m_{jk}^{(t)}\}_k),
    \label{eq:sample_theta}
\end{equation}
To sample the scaling factor $c^{(t)}$, we use the Chinese restaurant representation of $n_j^{(t)}$ together with \refeq{eq:pgcrt}:
\begin{align}
     c^{(t)} \sim  \Gam(e_0, f_0); \quad
     n^{(t)}_j \sim \CRT(n^{(t-1)}_j,c^{(t)}); \quad
     c^{(t)} \sim \GCRTP (n^{(t)}_{\marg{j}}, \{n^{(t-1)}_{j}\}_j, e_0, f_0).
     \label{eq:sample_c}
\end{align}
where $n_j^{(t)} = x_{\marg{k}j}^{(t)}$.
Because the relationship \refeq{eq:multphi} between $y_{vjk}^{(t)}$ and $\phi^{(t)}_{vk}$ is as in the PGBN model, we sample $\phi^{(t)}$ and its prior parameters using \refeq{eq:sampphi}-\refeq{eq:phigcrtp} as before, using a Dirichlet prior on $\phi^{(t)}$, and
%either a $\GSCP$ or
a gamma-Dirichlet prior on its concentration parameters. Finally, using a Dirichlet prior for $r_v$ we have the update equations
\begin{align}
    \{r_v\}_v \sim\Dir(\{\gamma_0/K_T\}_v);\qquad
    \{x_{vj}^{(T+1)}\}_v\sim\Mult(n_j^{(T+1)},\{r_v\}_v);\qquad
    \{r_v\}_v\sim\Dir(\{\gamma_0/K_t+x_{v\marg{j}}^{(T+1)}\}_v).
    \label{eq:sample_mr}
\end{align}

\section{Experiments}
% To visualise the topic hierarchies learned by the MBN, we computed the projections from higher-level topics onto pixel activations. For each of the four Markov chains, Fig.~\ref{fig:digits} illustrates the projection from the top layer $\pmb{\phi}^{(3)} \pmb{\phi}^{(2)} \pmb{\phi}^{(1)}$, the middle layer $\pmb{\phi}^{(2)} \pmb{\phi}^{(1)}$, and the bottom layer $\pmb{\phi}^{(1)}$ onto the pixels.

\subsection{Greedy layer-wise training on mutational signatures}
\label{supplementary_sec:training_mutational_signatures}
For both the single layer PGBN and MBN, four chains were run for 1700 Gibbs steps each. Samples from the last 250 iterations, thinned every fifth sample, were collected for analysis (leaving 50 samples per chain). Thereafter, an additional $K_2=78$ latent component layer was added on top of each respective model and the chains were run for 500 additional steps. For the PGBN we inferred 38 latent components on the second layer (that is, out of all four chains, the smallest number of empty signatures $m_{\marg{j}k}^{(2)}=0$). The top layer was subsequently pruned back to 38 latent components and the chains were run for an additional 550 steps collecting the last 250 samples thinned to 50 samples. The MBN, was (accidentally) run slightly longer, for 750 steps, and we inferred 41 latent components. After pruning the empty topics, 250 additional steps were collected and thinned for analysis. Overall, a total of 77 days (78 days) of GPU time---divided across four nVidia A40 GPU devices---were used to execute 2700 Markov steps per chain for the MBN (2800 steps for the PGBN).

\subsection{Meta-signature construction}
\label{sec:meta_signatures}
Consensus meta-signatures were determined by matching the topics of different chains to its' centroid by repeatedly solving the optimal transport problem~\cite{S_MURP23} for the Jensen-Shannon distance (JSD)~\cite{S_MURP23} using the Hungarian algorithm~\cite{S_CROU16} until the centroid converged in terms of silhouette score~\cite{S_ROUS87}, similar to Ref.~\cite{S_ALEX20}. The centroid was initialised with restarting points coming from different chains and the consensus meta-signatures that gave the best silhouette score were selected. Finally, we selected robust meta-signatures by choosing those centroids where the JSD between the closest signature was no less than 0.25 across all chains, leaving four meta-signatures in total (named, M$_1$ through M$_4$).
For completeness, we list all 37 other meta signatures in Figs.~\ref{fig:meta_signatures_rest_a}-~\ref{fig:meta_signatures_rest_c}. While completely inactive meta signatures were pruned, seven out of the 41 signatures remained with a very small topic activity throughout the dataset (to wit, M$_{23}$, M$_{26}$, M$_{35}$, M$_{37}$-M$_{40}$).

\subsection{Interpretation meta signatures $\mathrm{M}_1,\dots,\mathrm{M}_4$}
\label{sec:meta_signatures_biology}
Here, we characterise the four meta signatures named M$_1$ through M$_4$. Summarising the
  meta-signatures by entropy $s(k) = -\sum_{v=1}^{78} \phi^{(2)}_{vk} \ln
  \phi^{(2)}_{vk}$, we found that the posterior coverage was low with an entropy-based
  effective sample size~\citep{S_VEHT21} of 11, 5, 6, and 6, respectively. 

 Next, we describe, per meta-signature, the ($K_1=78$) mutational signatures, $v$, exceeding three times uniform probability (i.e., $\phi_{vk}^{(2)} \geq 3/K_1$, analogous
  to~\citet{S_journals/jmlr/ZhouCC16}) and their biological interpretation.
  
  M$_1$ describes the co-occurrence of replicative DNA polymerase $\epsilon$ (POLE)
  damage (SBS10a, SBS10b, and SBS28~\citep{S_LI18,S_HODE20}, but \textit{not} POLE
  associated SBS14~\citep{S_HODE20}) and mismatch-repair deficiency (MMR, SBS15 and
  SSB21~\citep{S_MEIE18}) (Fig.~\ref{fig:meta-signatures}, first row, left column).
  Tumours with an ultra-hypermutated phenotype ($\geq$ 100  Mb$^{-1}$) are
  often characterised by these joint disruptions in MMR and POLE~\citep{S_HODE20}.
  Combined, M$_1$ describes a preference for altering C$\rightarrow$A and
  T$\rightarrow$G when flanked by a T on either side (Fig.~\ref{fig:meta-signatures},
  first row, right column). 

  Meta-signature M$_2$ primarily captures, presumably, oxidative stress. Its
  constituents SBS17a/b~\citep{S_SECR16} and SBS18 are thought to be related to guanine
  oxidation, resulting in the formation of 8-Oxo-2'-deoxyguanosine
  (8-oxo-dG)~\citep{S_NONE14,S_TOMK18,S_POET18,S_CHRI19}; SBS18 is additionally linked to
  hydroxyl radicals in culture~\citep{S_KUCA19}. Similar to clock-like signature SBS1
  (describing spontaneous deamination of 5-methylcytosine~\citep{S_NIK12,S_ALEX15}), damage
  due to 8-oxo-dG accumulates in the course of life~\citep{S_NIE13}. To a lesser extent,
  M$_2$ also captures SBS8, which is implicated in BRCA1 and BRCA2 dysfunction in breast
  cancer~\citep{S_NIKZ16} and believed to be (uncorrected) replication
  errors~\citep{S_SING20}. Characteristically, M$_2$ prefers T$\rightarrow$G and
  T$\rightarrow$A singlets with a contextual T on the right-hand side
  (Fig.~\ref{fig:meta-signatures}, second row, right column).

  Meta-signature M$_3$ is marked by a pronounced transcriptional strand bias, including
  signatures such as SBS5, SBS8, SBS12, SBS16~\citep{S_ALEX20}, along with
  SBS92~\citep{S_LAWS20} and SBS22~\citep{S_COSM23}, with SBS40 being the exception. Its
  primary constituent SBS12 is believed to be related to transcription-coupled
  nucleotide excision repair~\citep{S_ALEX20}. The second largest contributor, SBS40, is a
  spectrally flat, late-replicating~\citep{S_SING20}, signature with spectral similarities
  to SBS5 (both are related to age~\citep{S_ALEX15}) and is believed to be linked to
  SBS8~\citep{S_SING20}. According to COSMIC, some contamination between SBS5 and SBS16
  may be present~\citep{S_TATE19,S_COSM23}; M$_3$ is consistent with this observation.
  Finally, M$_3$ also captures the co-occurrence with SBS22, which is canonically
  attributed to aristolochic acid exposure~\citep{S_HOAN13,S_POON13,S_NIK15}. Overall, M$_3$
  gives rise to a dense spectrum and inherits the quintessential depletion of C
  substitutions when right-flanked by a G from SBS40 (Fig.~\ref{fig:meta-signatures},
  third row, right column).

\begin{figure}
    \centering
    \includegraphics[width=\textwidth]{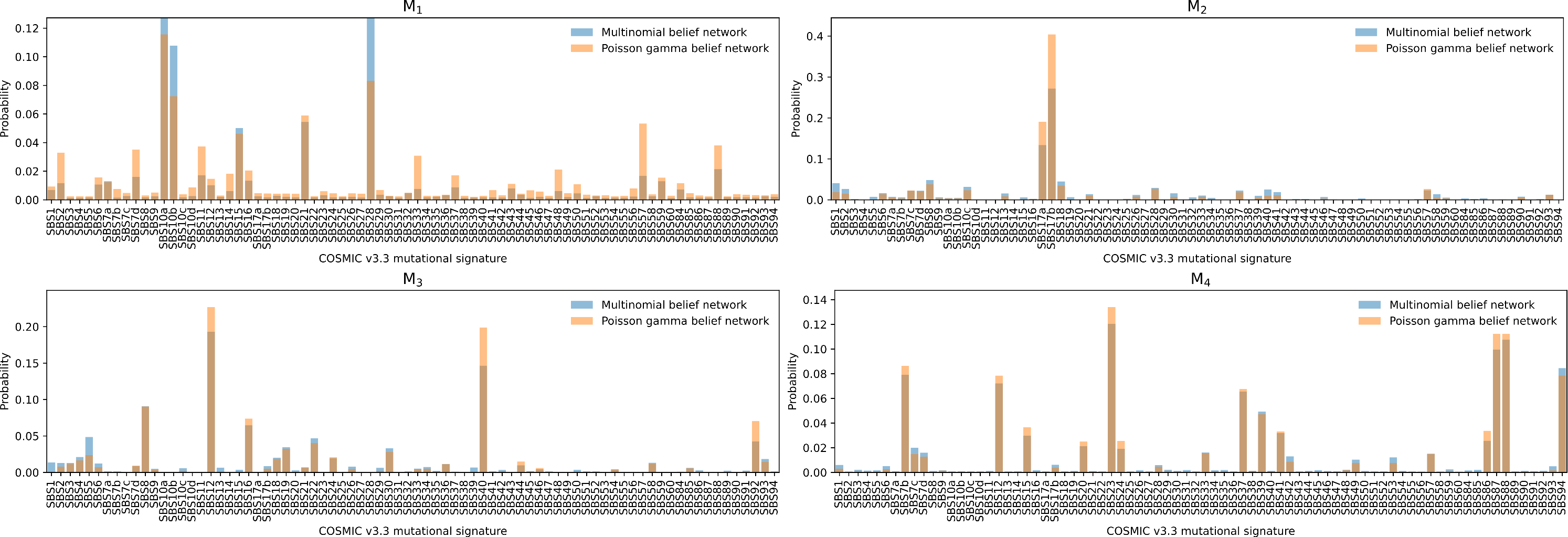}
    \caption{Meta signatures M$_1$-M$_4$ replicate in the Poisson gamma belief network (PGBN).
    The data shows the posterior average of the closest matching meta signatures extracted from the PGBN.}
    \label{fig:meta_comparison}
\end{figure}

  Finally, M$_4$ describes the co-occurrence of several, seemingly disparate, mutational
  signatures of known and unknown aetiology. Of known cause are, SBS7b, linked to
  ultraviolet light~\citep{S_NIK15, S_HAYW17}, SBS87 to thiopurine chemotherapy
  exposure~\citep{S_LI20} (although its presence has been reported in a thiopurine-naive
  population~\citep{S_DONK23}) and SBS88, related to colibactin-induced damage from the
  Escherichia coli bacterium~\citep{S_PLEG20,S_BOOT20} (found in various tissue
  types~\citep{S_BOOT20,S_LEE19,S_PLEG20}). Concurrently, M$_4$ comprises
  SBS12~\citep{S_ALEX20}, SBS23~\citep{S_ALEX15,S_NIKZ16}, SBS37~\citep{S_ALEX20},
  SBS39~\citep{S_ALEX20} and SBS94~\citep{S_ISL22}, all of unknown cause. Jointly, these
  signatures describe a dense spectrum with a slight tendency for C$\rightarrow$T
  substitutions. Reassuringly, meta-signatures M$_1,\dots,$M$_4$ replicated
  independently in the PGBN (Fig.~\ref{fig:meta_comparison}, Supplementary Material). To
  our knowledge, this is the first time a first-principles characterisation of the
  organising principles of mutagenic processes in cancer has been carried out.

\begin{figure}
    \centering
    \begin{tabular}{cc}
        \includegraphics[width=0.5\textwidth]{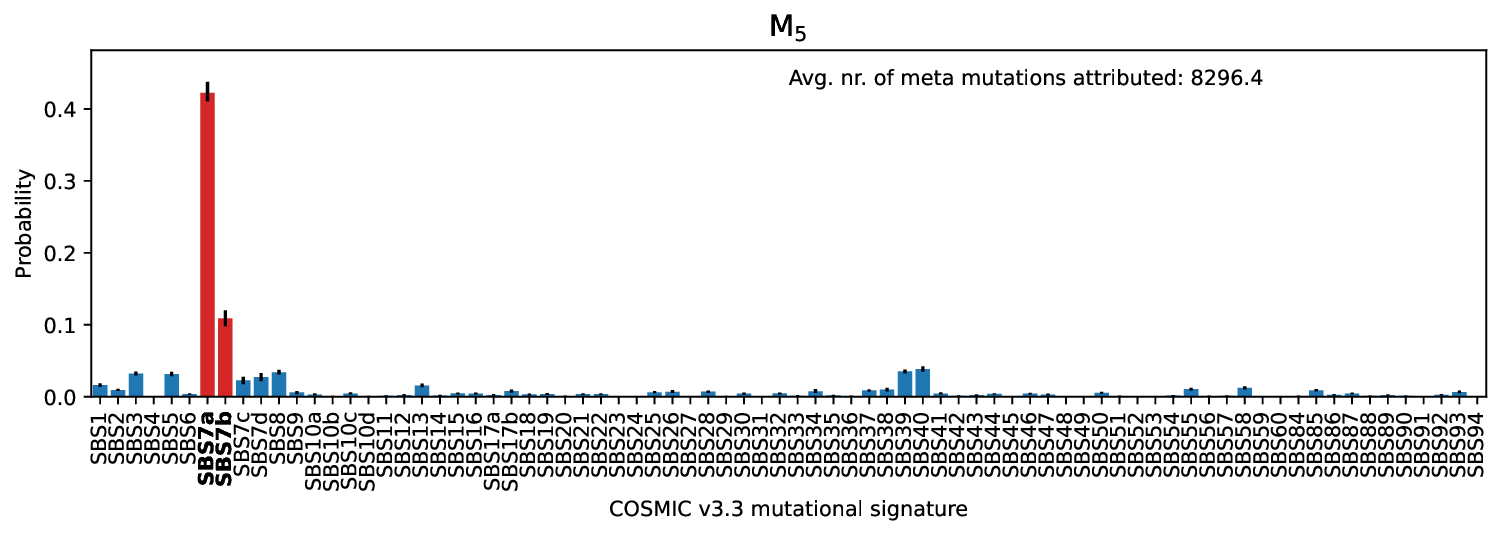}
        &
        \includegraphics[width=0.5\textwidth]{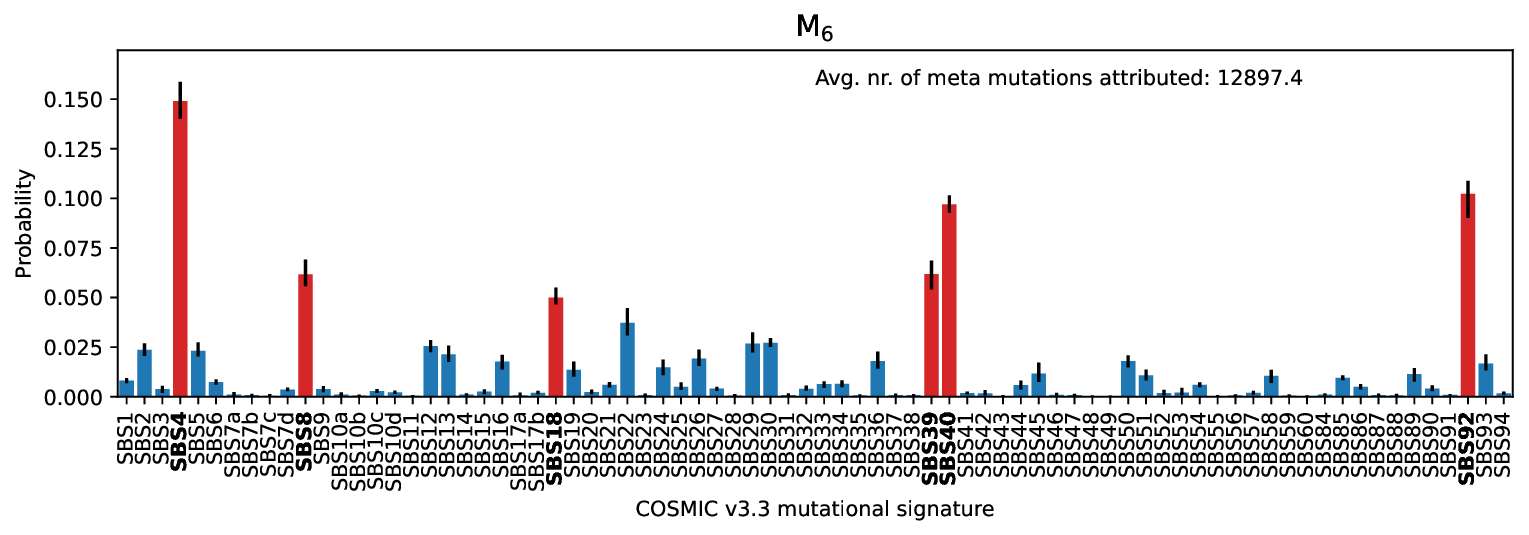}
        \\
        \includegraphics[width=0.5\textwidth]{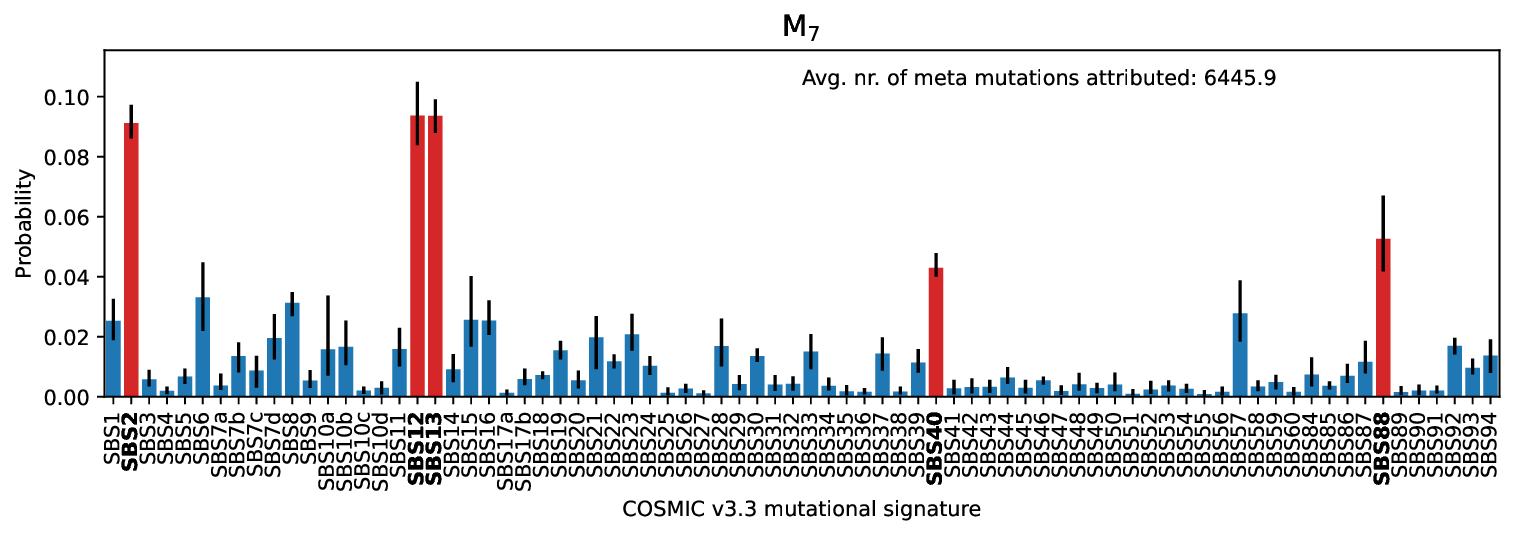}
        &
        \includegraphics[width=0.5\textwidth]{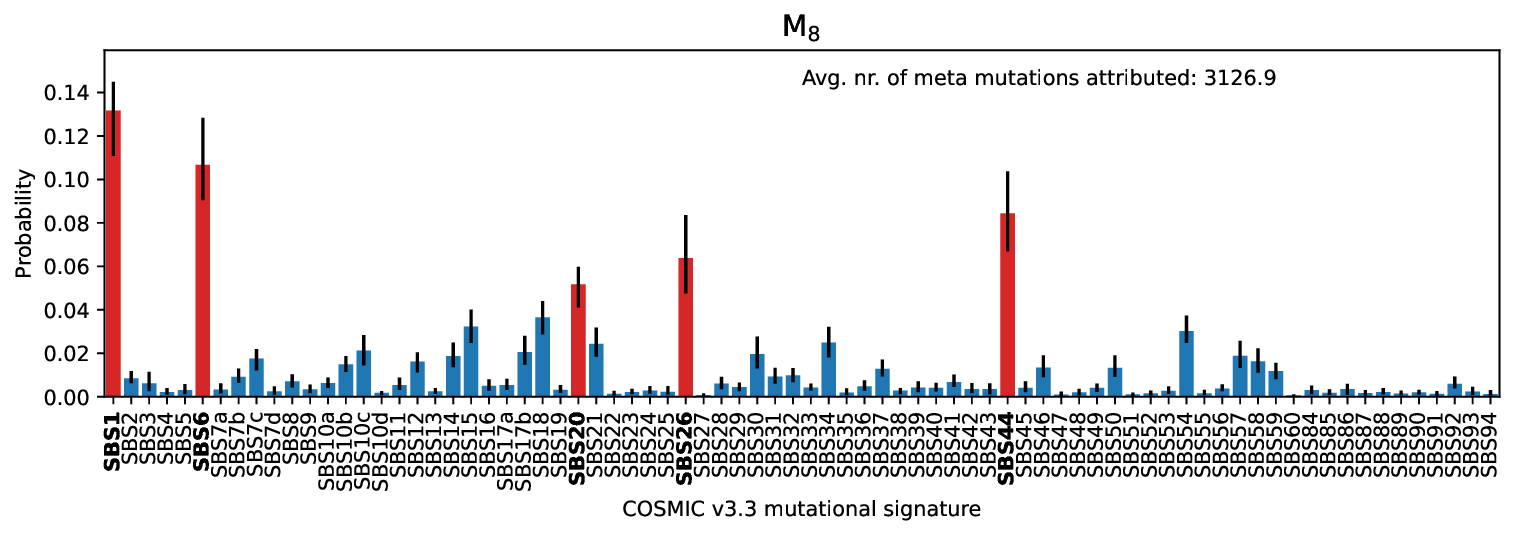}
        \\
        \includegraphics[width=0.5\textwidth]{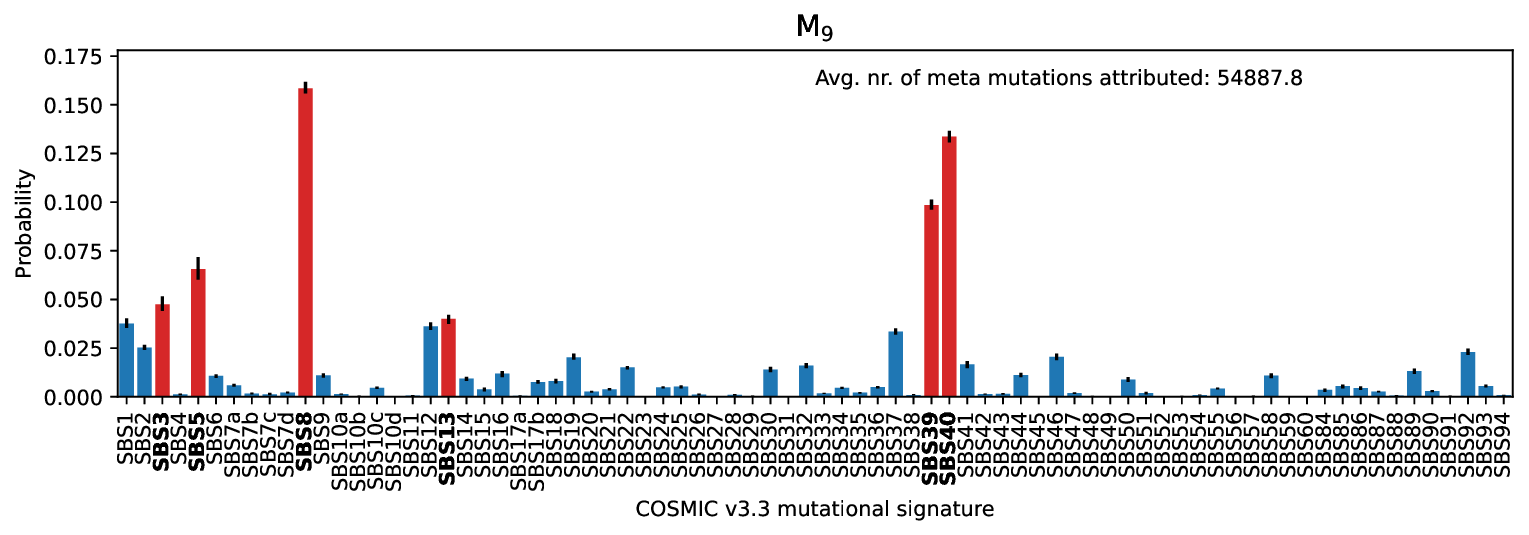}
        &
        \includegraphics[width=0.5\textwidth]{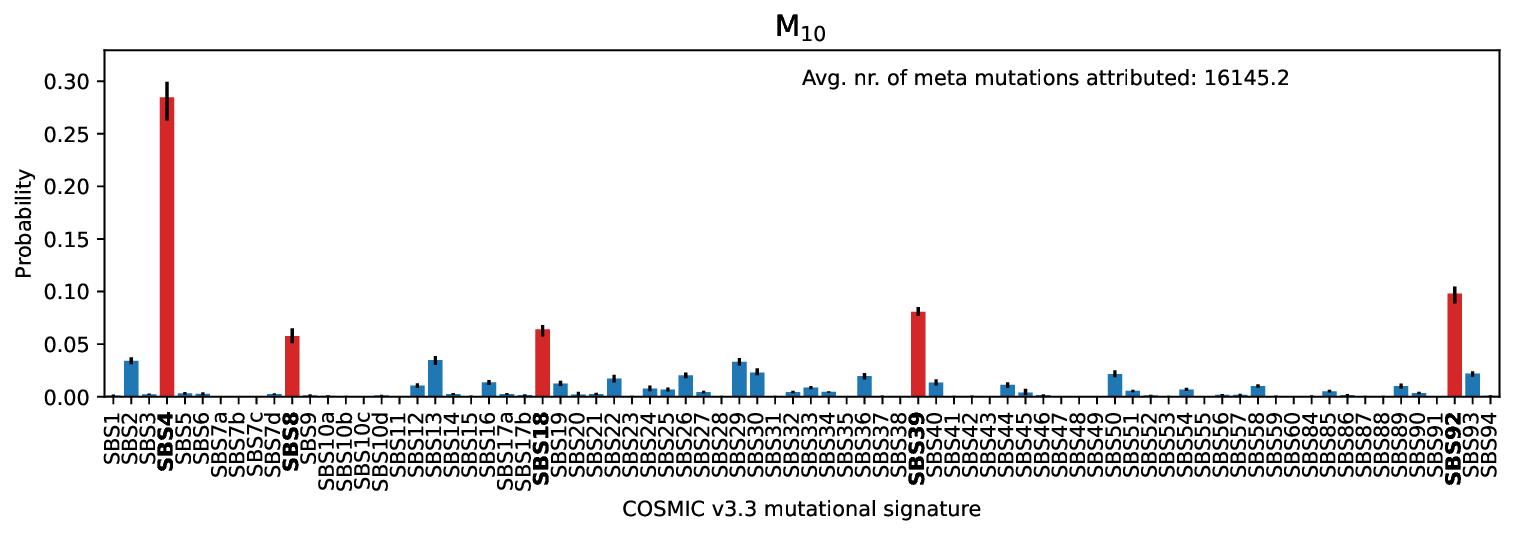}
        \\
        \includegraphics[width=0.5\textwidth]{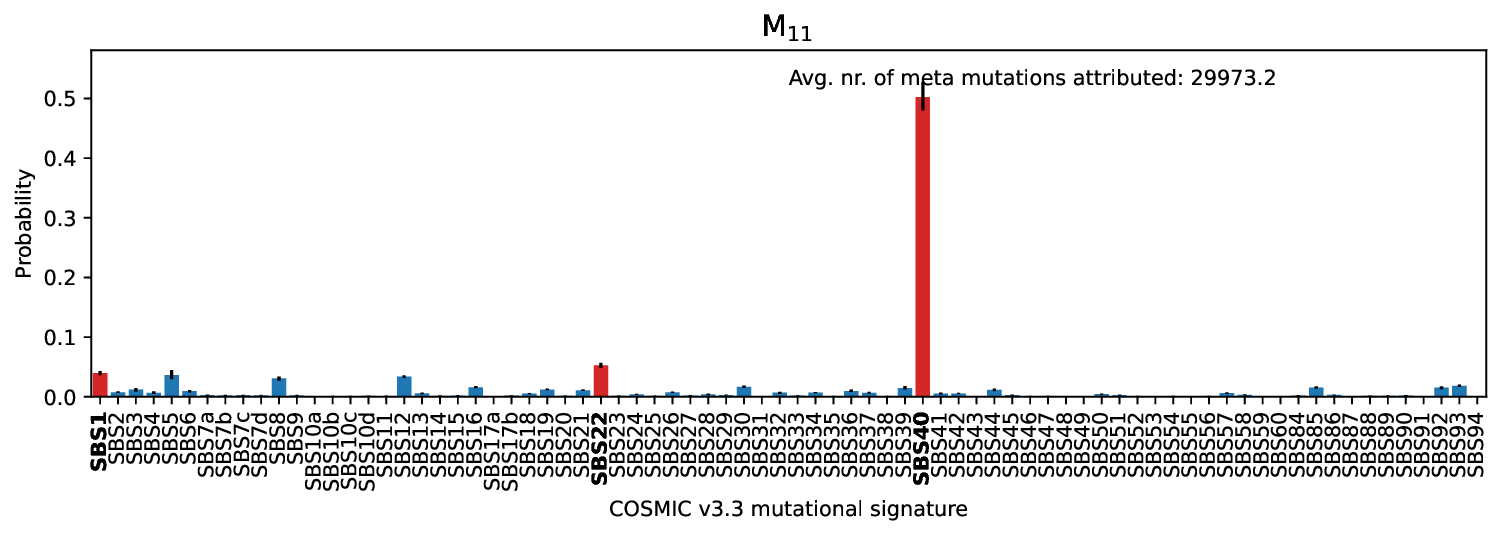}
        &
        \includegraphics[width=0.5\textwidth]{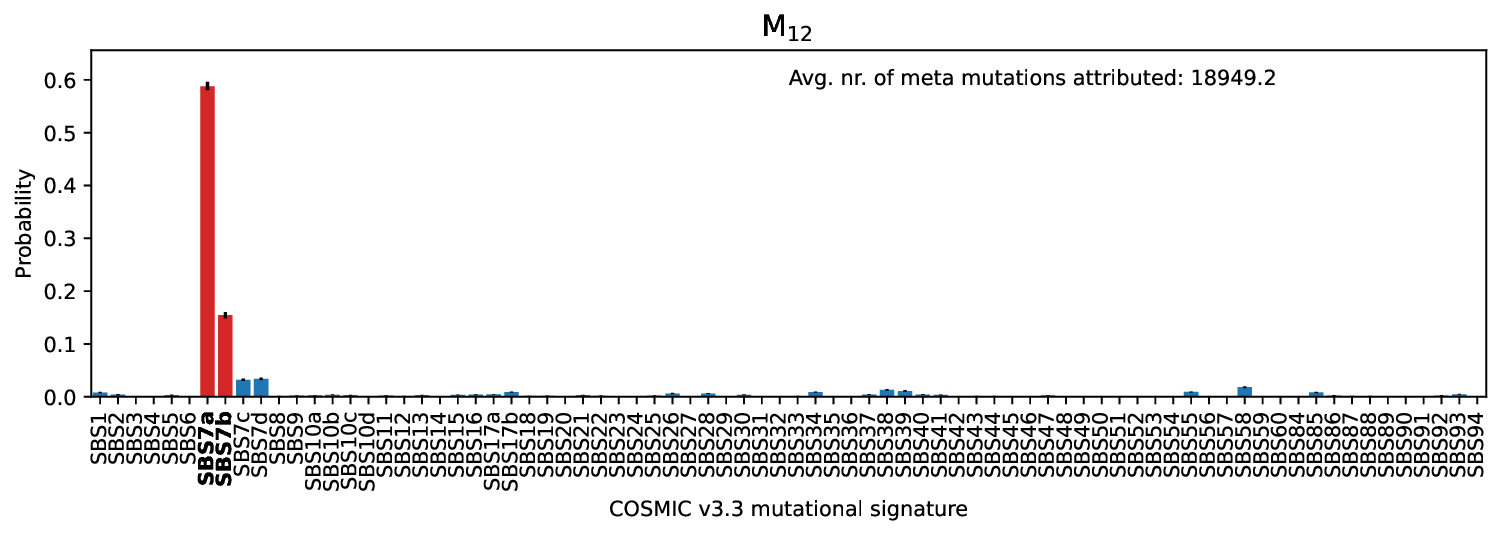}
        \\
        \includegraphics[width=0.5\textwidth]{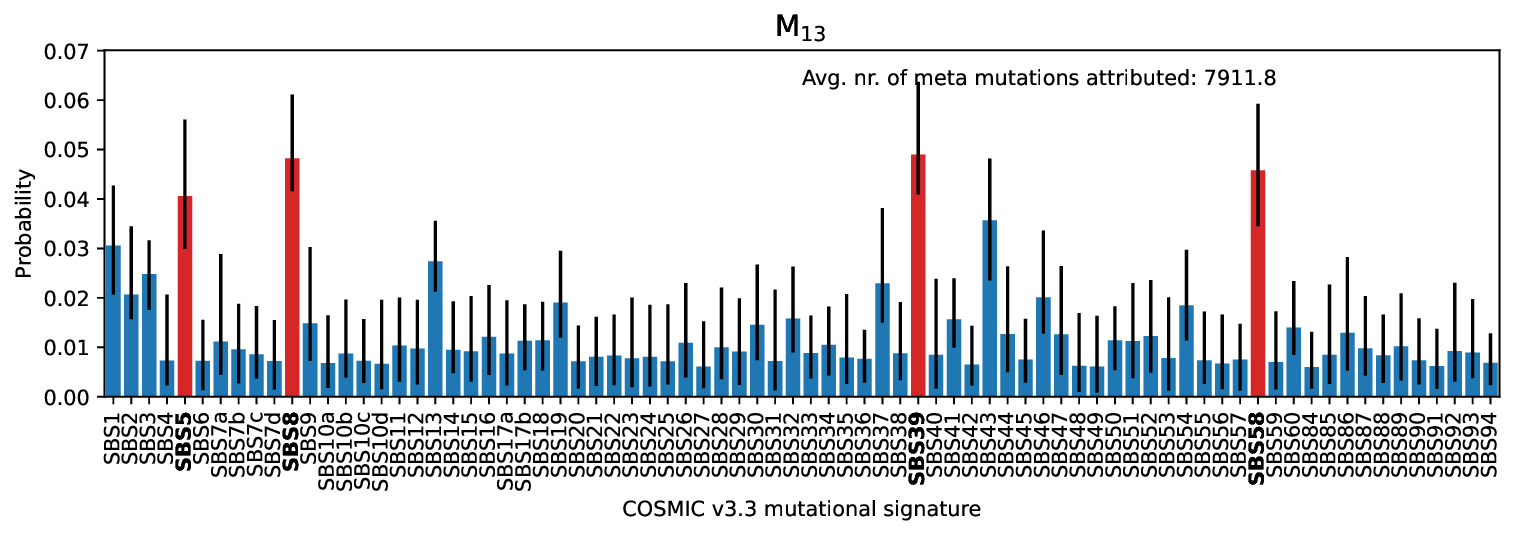}
        &
        \includegraphics[width=0.5\textwidth]{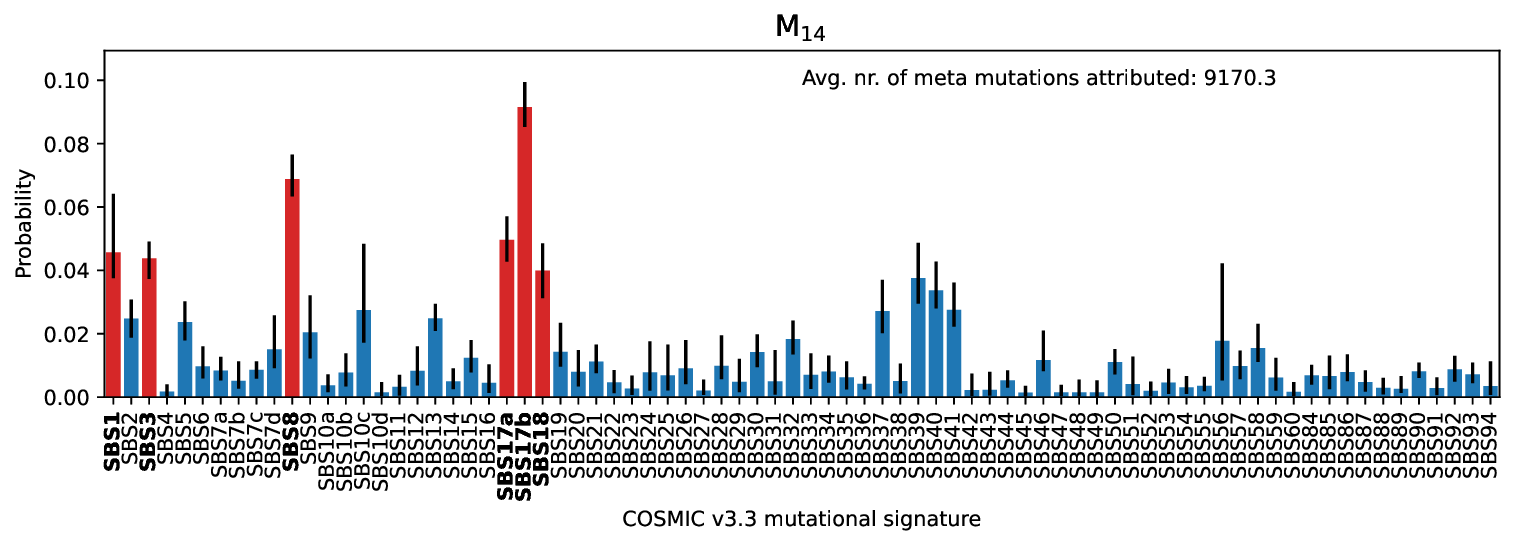}
        \\
        \includegraphics[width=0.5\textwidth]{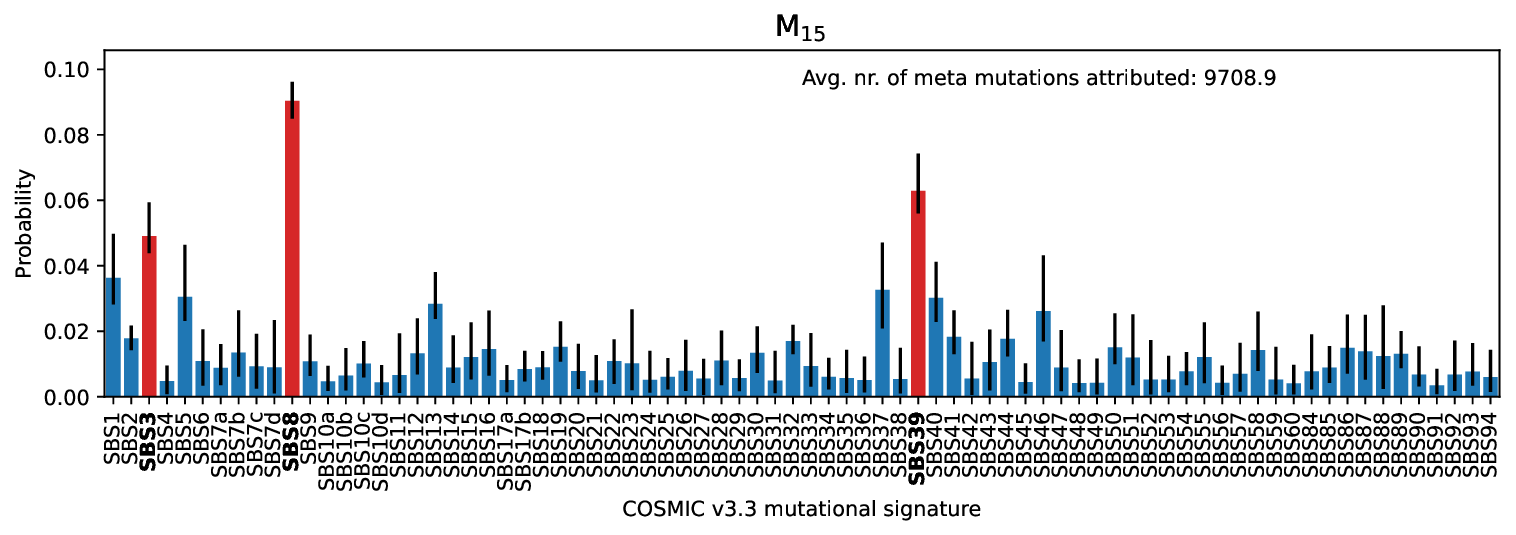}
        &
        \includegraphics[width=0.5\textwidth]{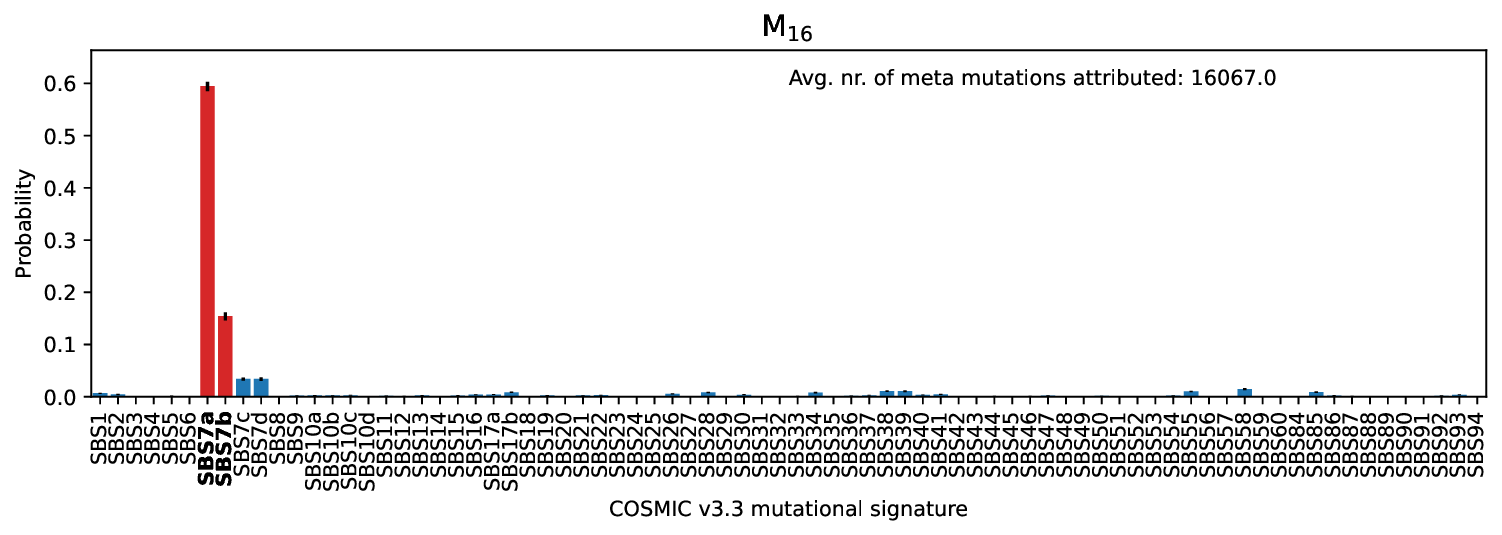}
        \\
        \includegraphics[width=0.5\textwidth]{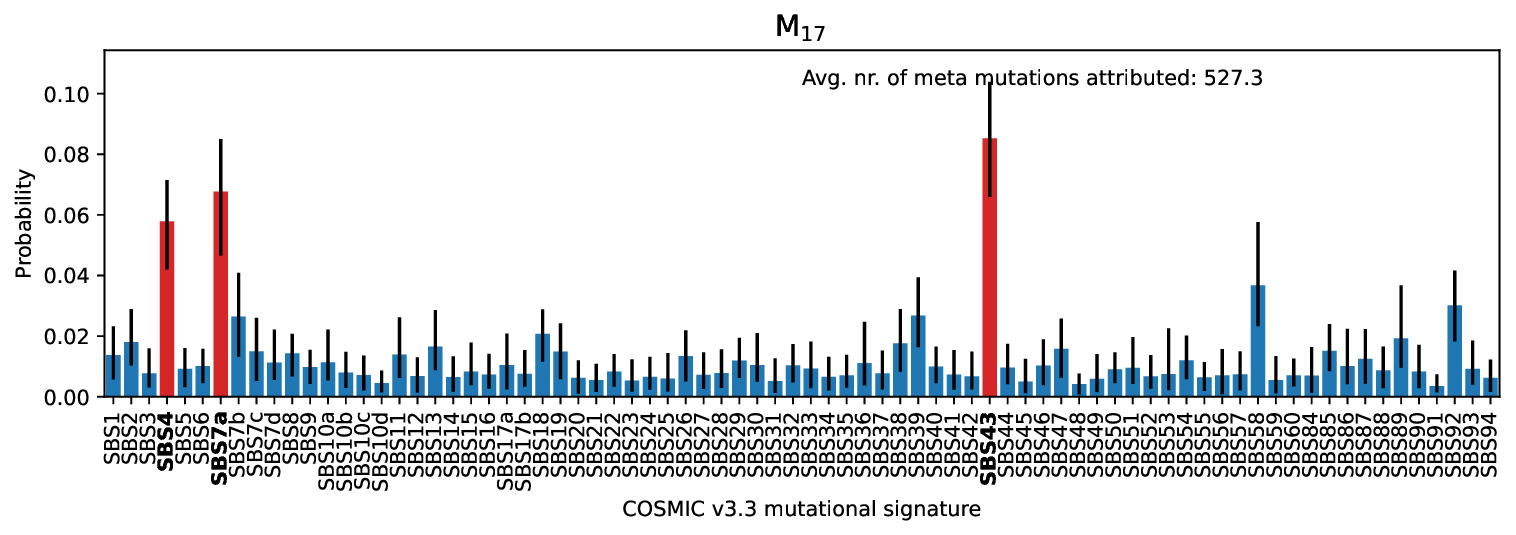}
        &
        \includegraphics[width=0.5\textwidth]{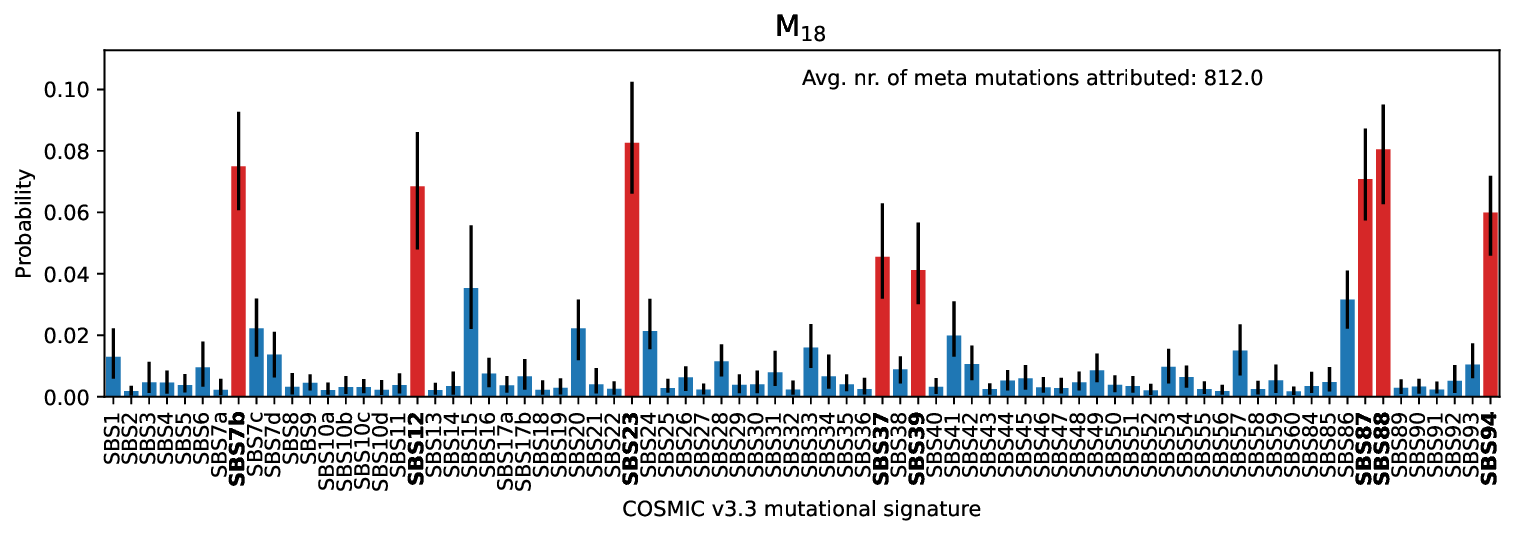}
    \end{tabular}
    \caption{Posterior of meta-mutational signatures $k = \text{M}_5,\dots, \text{M}_{18}$ for the multinomial belief network (meta signatures $k = \text{M}_{19},\dots, \text{M}_{41}$ are listed in subsequent figures). In each panel, the total number of meta signature $k$ counts $m^{(2)}_{k\marg{j}}$ (averaged over the posterior samples) is indicated as a measure of topic loading.}
    \label{fig:meta_signatures_rest_a}
\end{figure}

\begin{figure}
    \centering
    \begin{tabular}{cc}
        \includegraphics[width=0.5\textwidth]{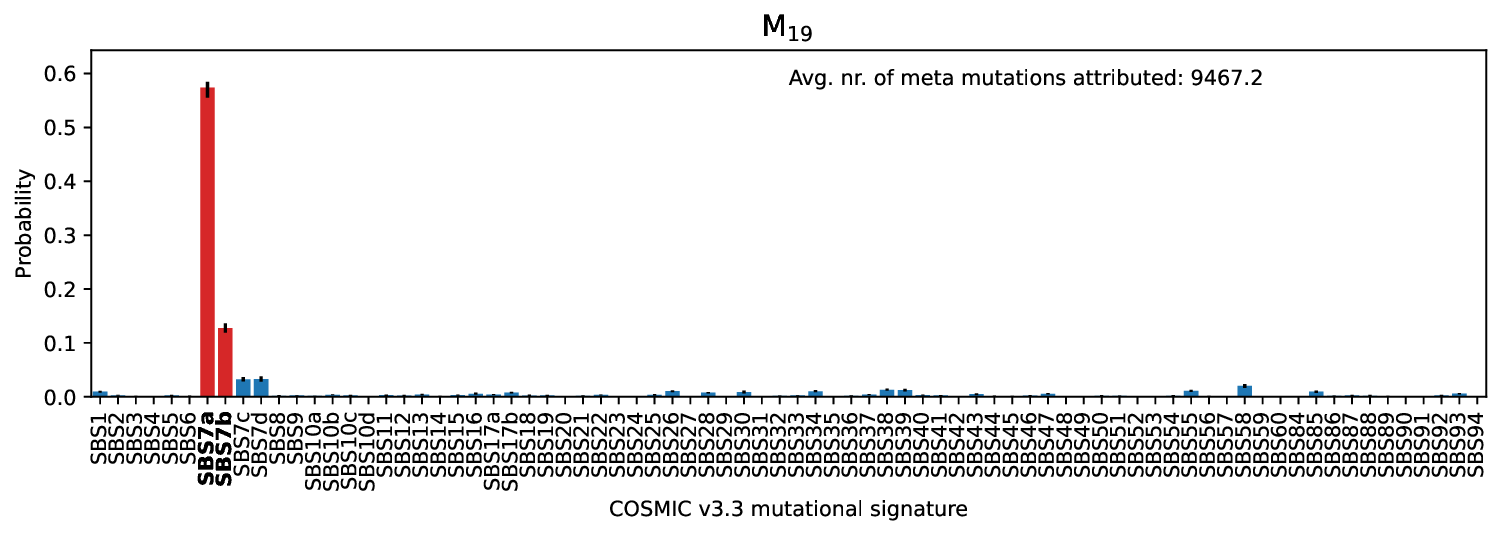}
        &
        \includegraphics[width=0.5\textwidth]{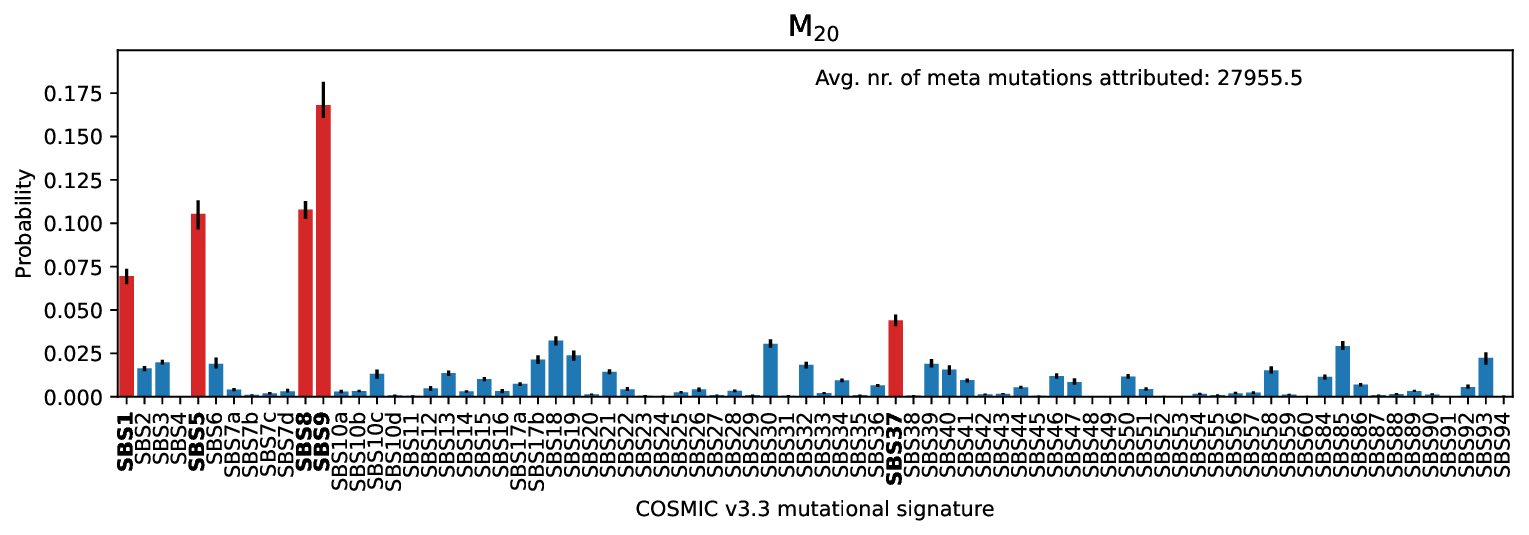}
        \\
        \includegraphics[width=0.5\textwidth]{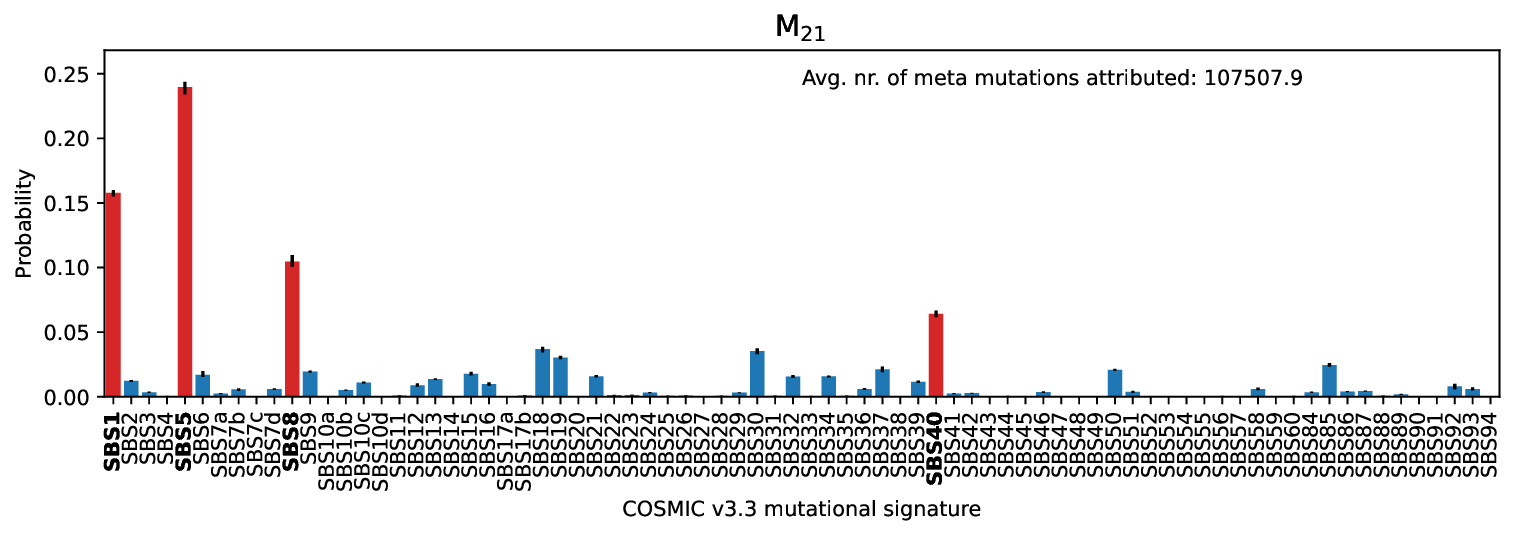}
        &
        \includegraphics[width=0.5\textwidth]{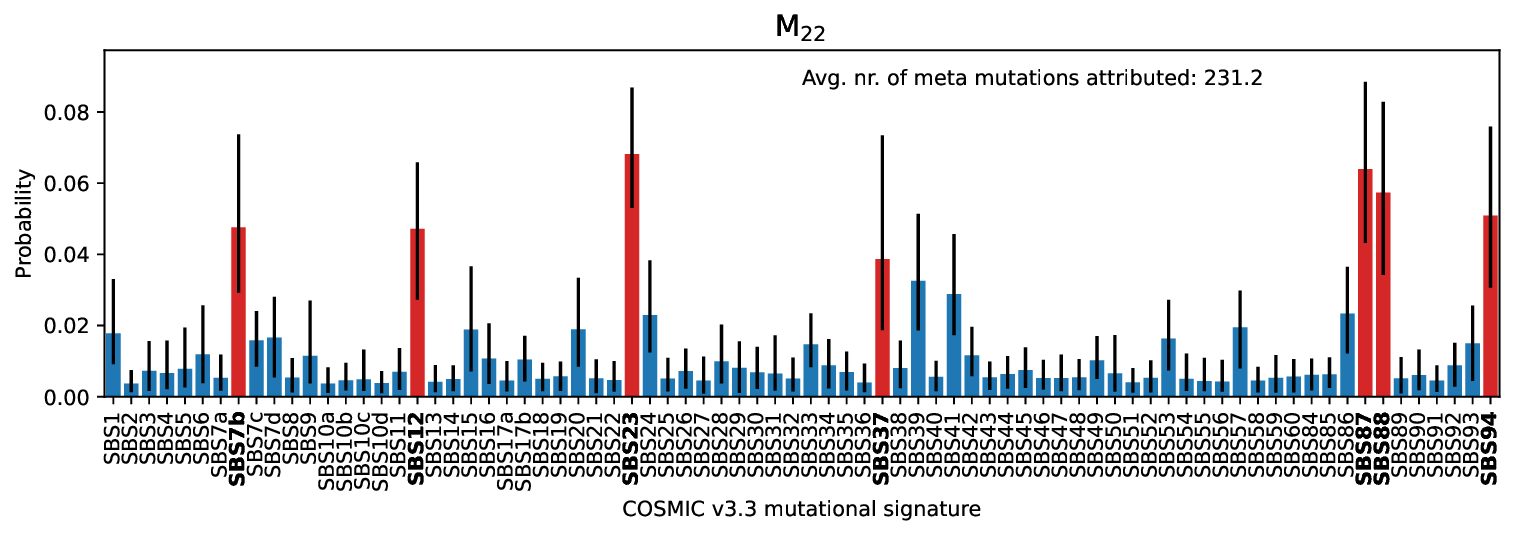}
        \\
        \includegraphics[width=0.5\textwidth]{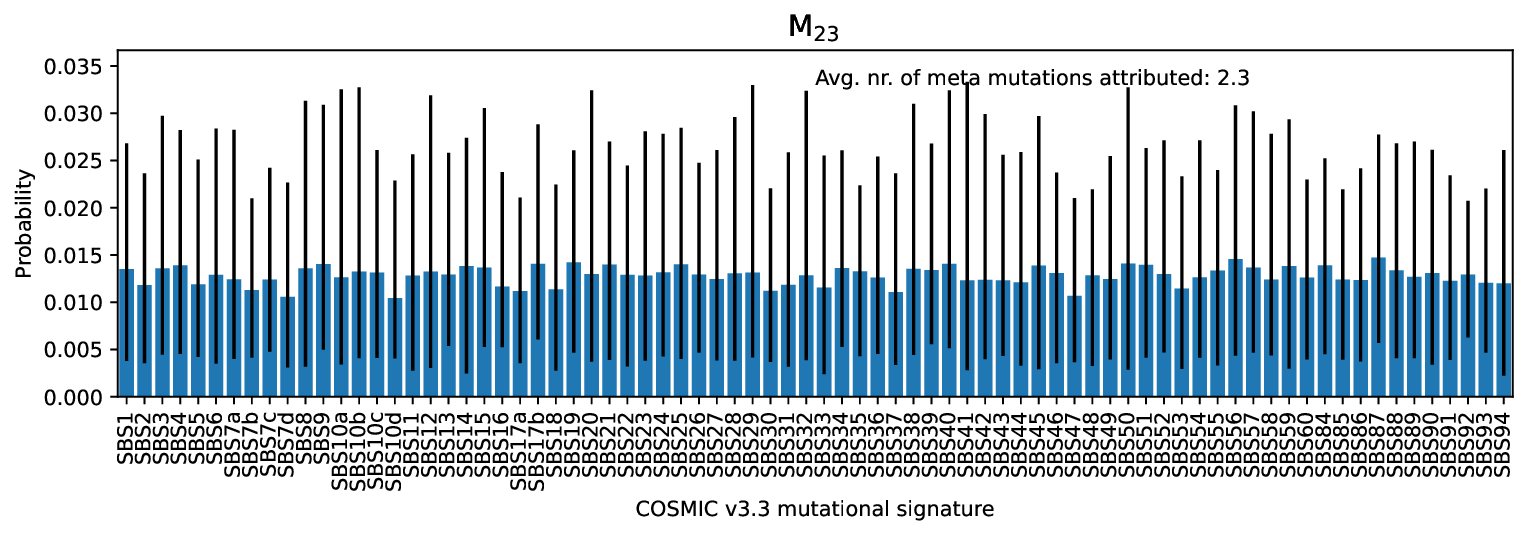}
        &
        \includegraphics[width=0.5\textwidth]{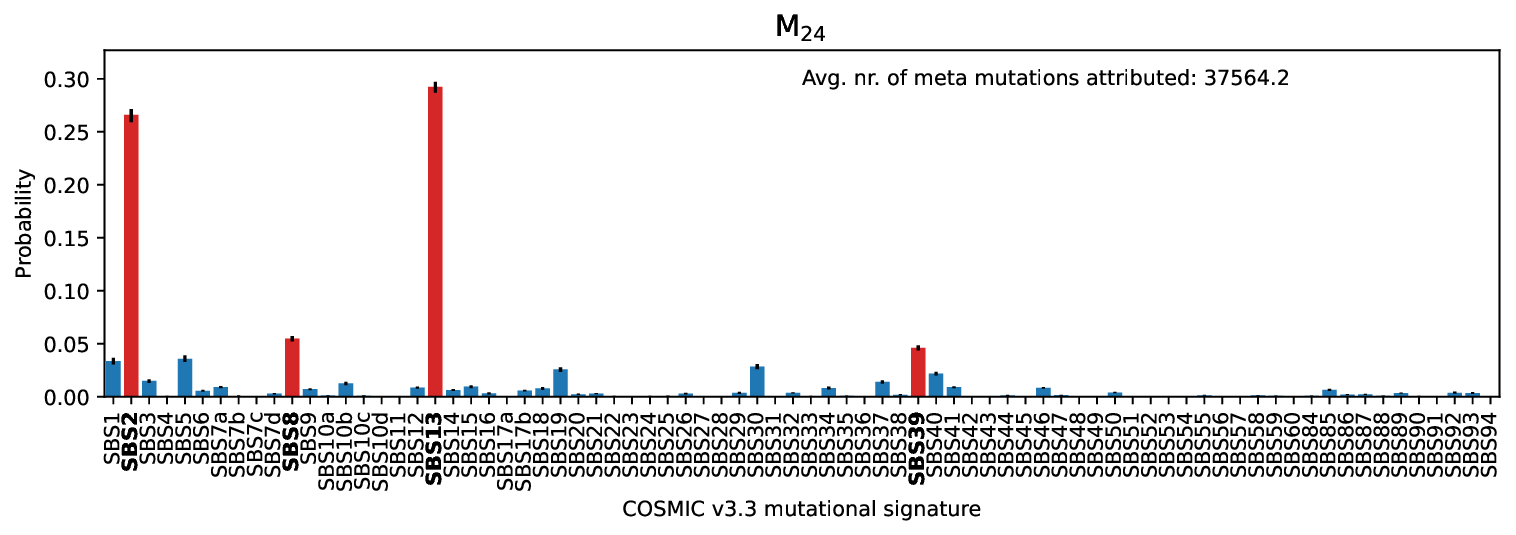}
        \\
        \includegraphics[width=0.5\textwidth]{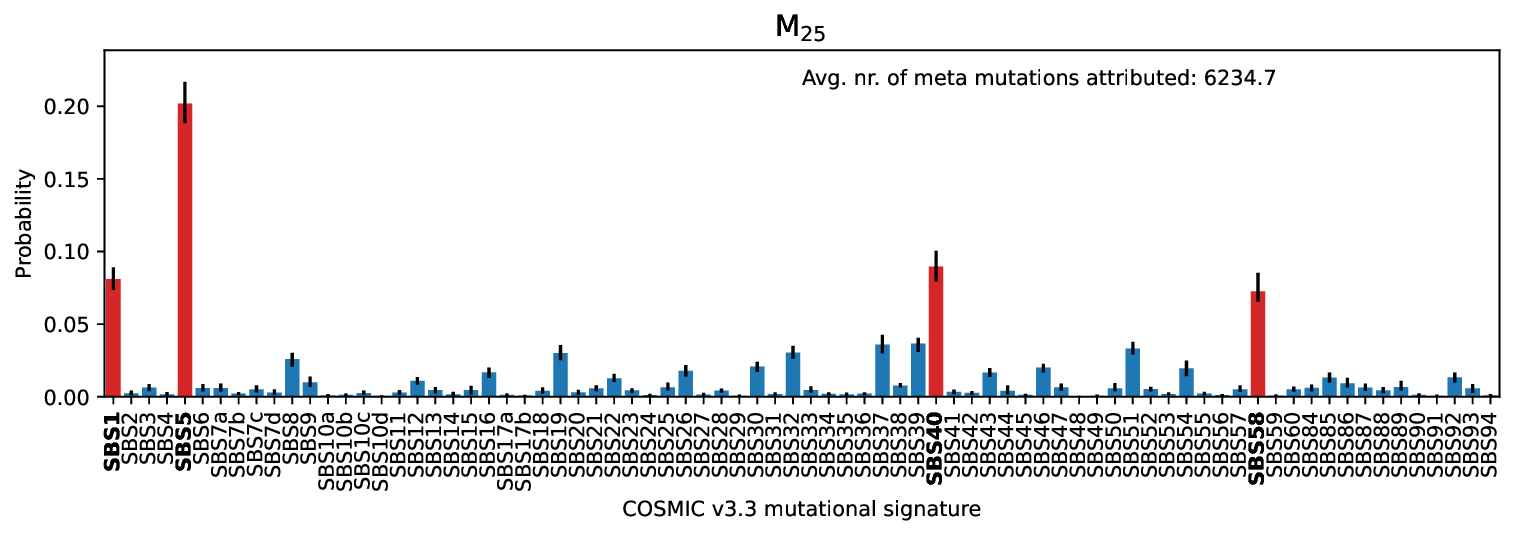}
        &
        \includegraphics[width=0.5\textwidth]{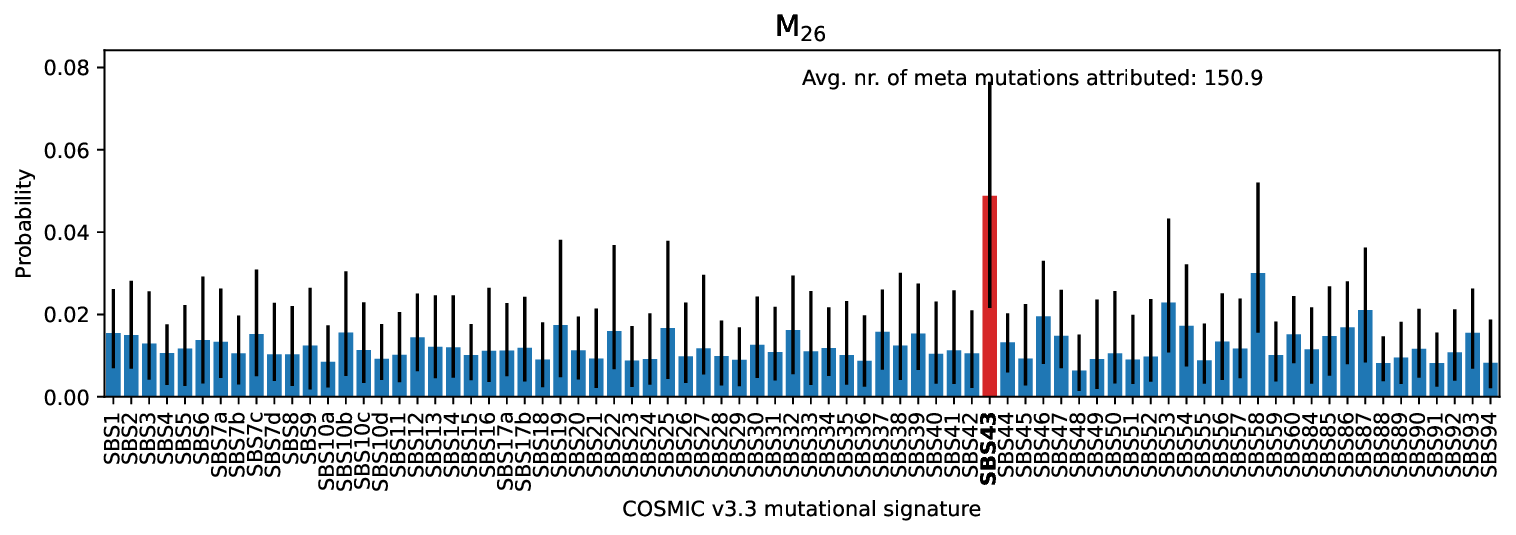}
        \\
        \includegraphics[width=0.5\textwidth]{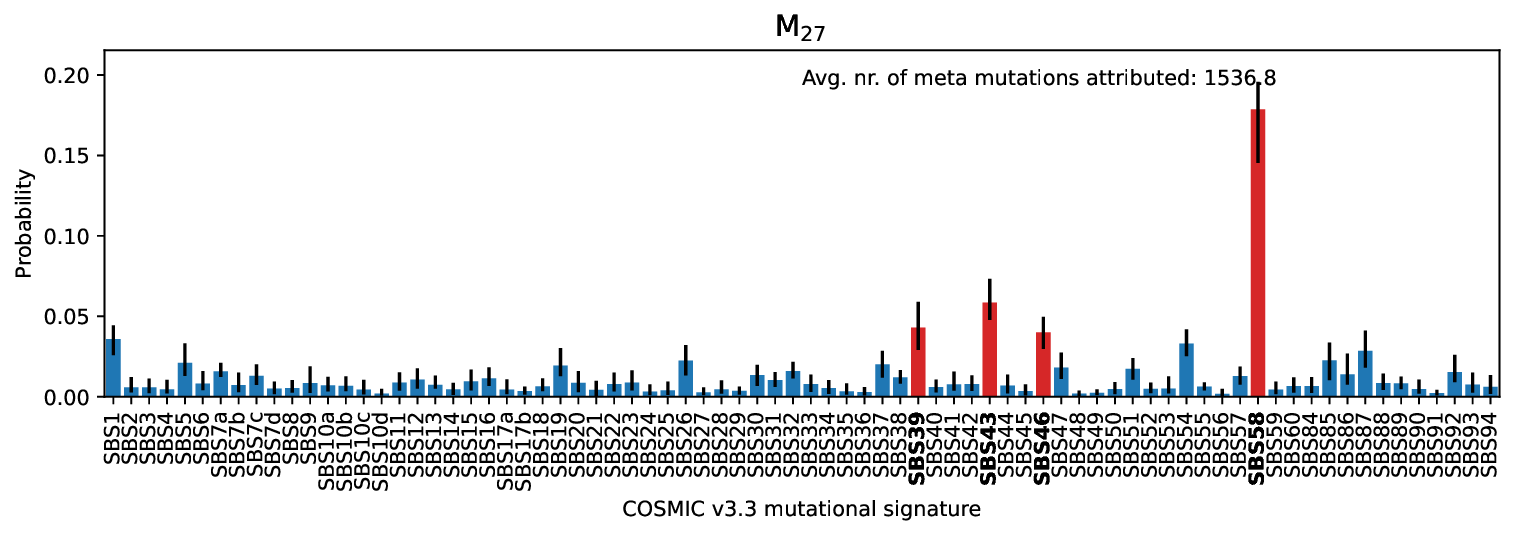}
        &
        \includegraphics[width=0.5\textwidth]{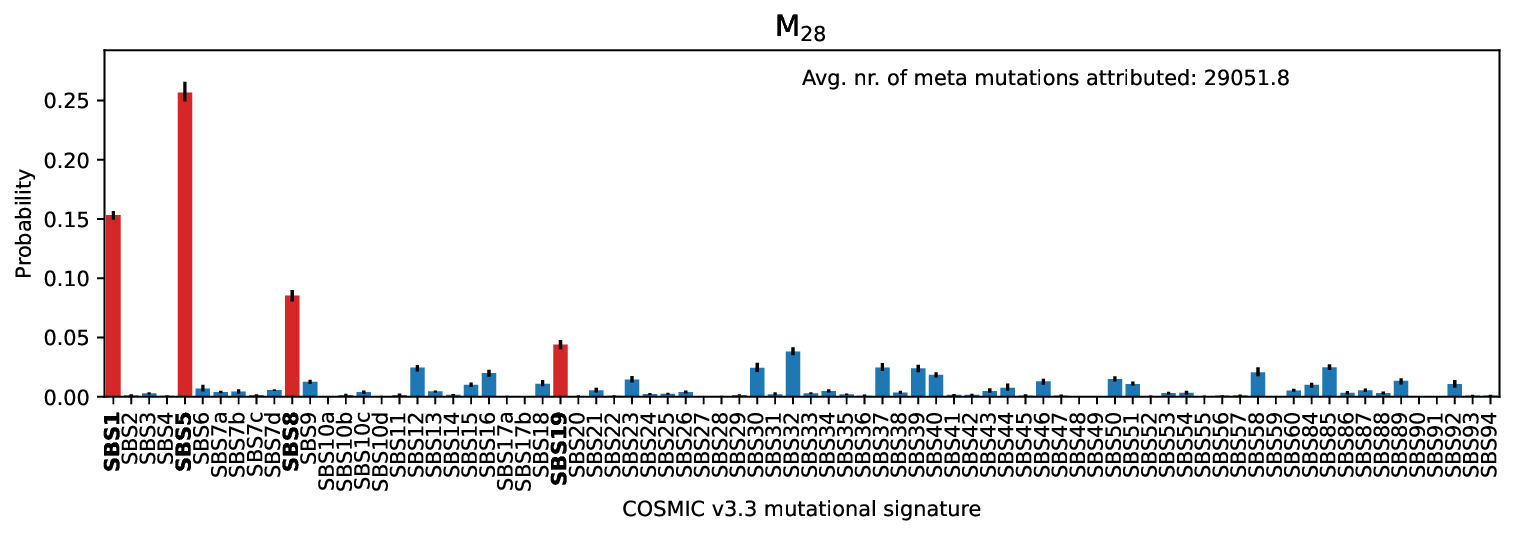}
        \\
        \includegraphics[width=0.5\textwidth]{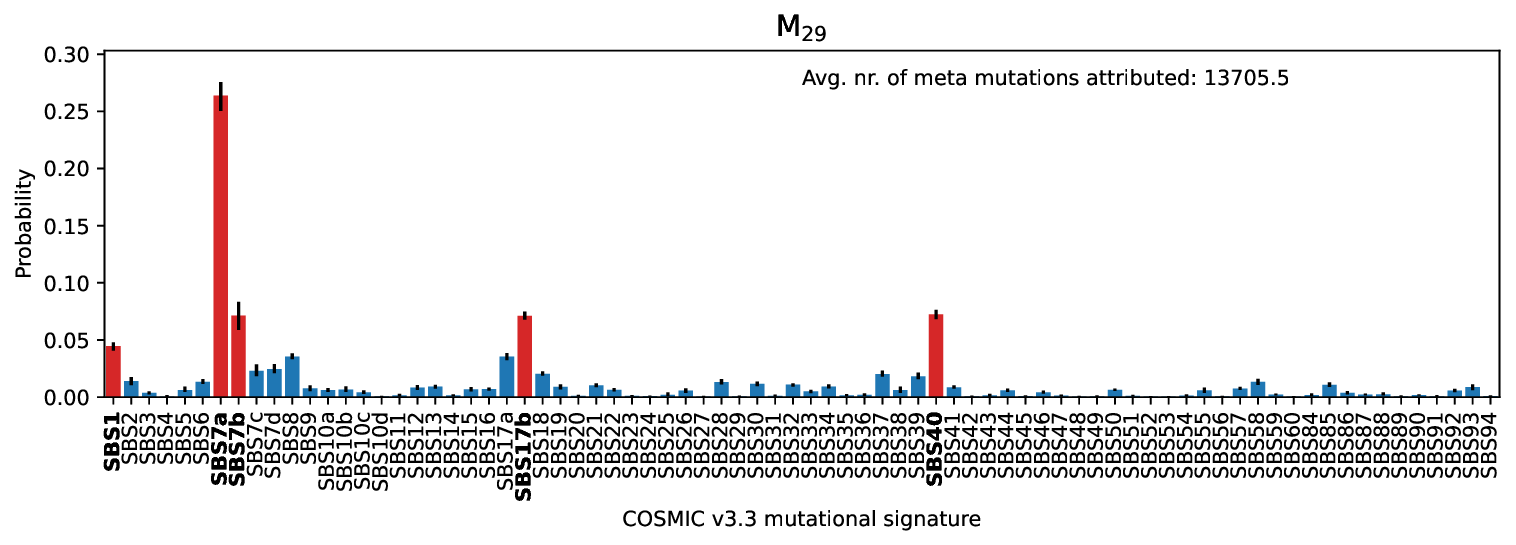}
        &
        \includegraphics[width=0.5\textwidth]{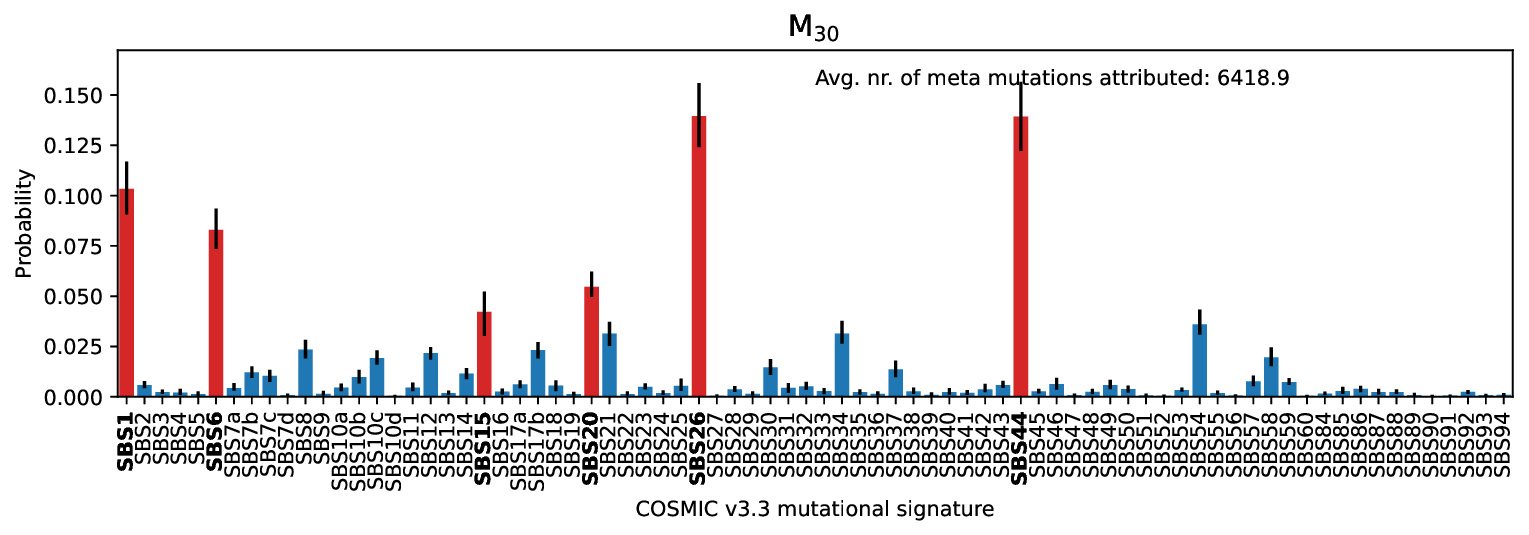}
        \\
        \includegraphics[width=0.5\textwidth]{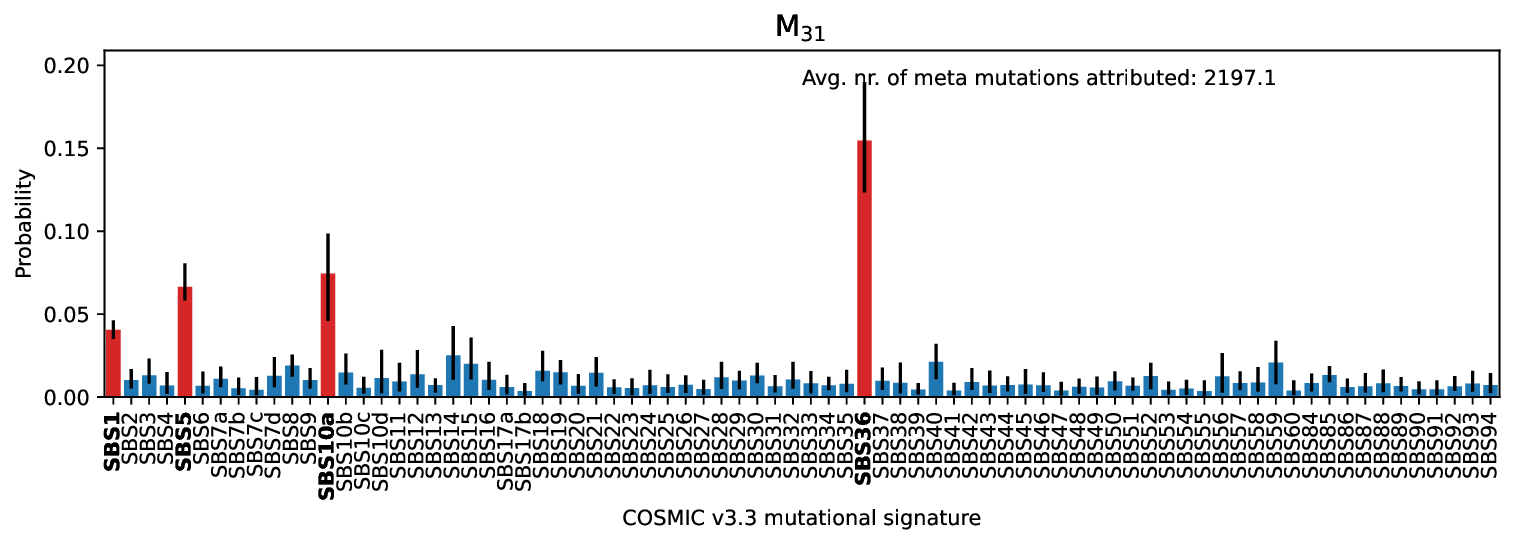}
        &
        \includegraphics[width=0.5\textwidth]{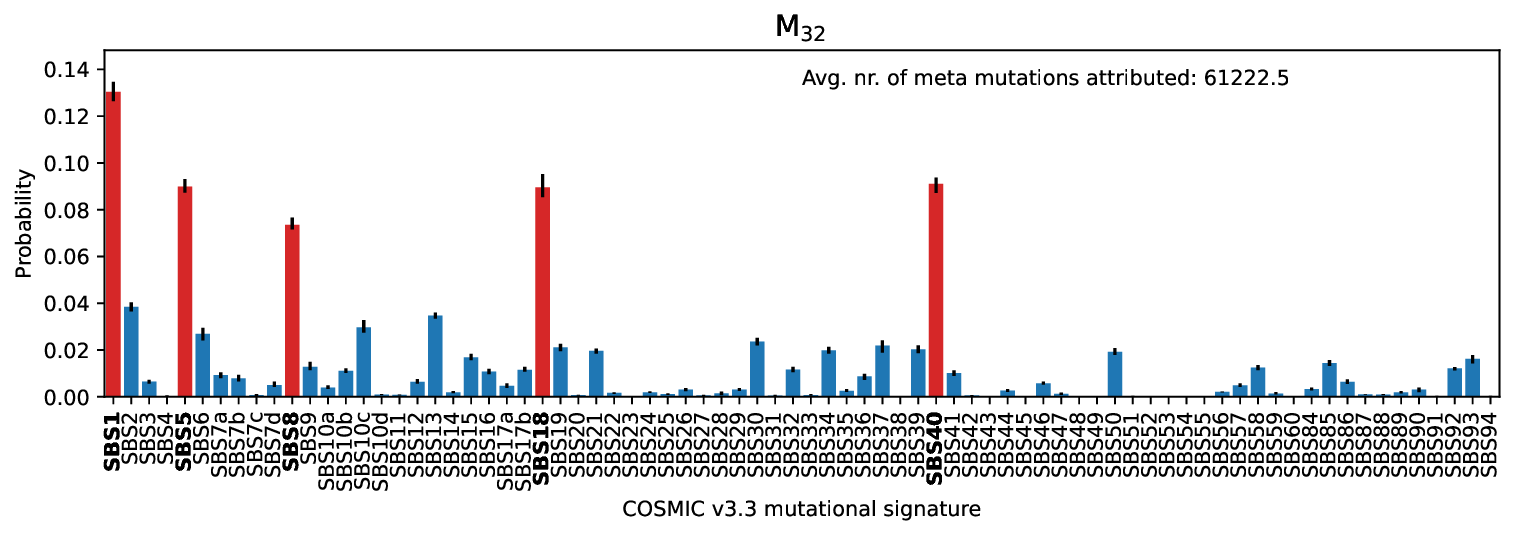}
        \\
    \end{tabular}
    \caption{Continuation of Fig.~\ref{fig:meta_signatures_rest_a} listing $k = \text{M}_{19},\dots, \text{M}_{32}$ (meta signatures $k = \text{M}_{33},\dots, \text{M}_{41}$ are listed in the subsequent figure).}
    \label{fig:meta_signatures_rest_b}
\end{figure}

\begin{figure}
    \centering
    \begin{tabular}{cc}
        \includegraphics[width=0.5\textwidth]{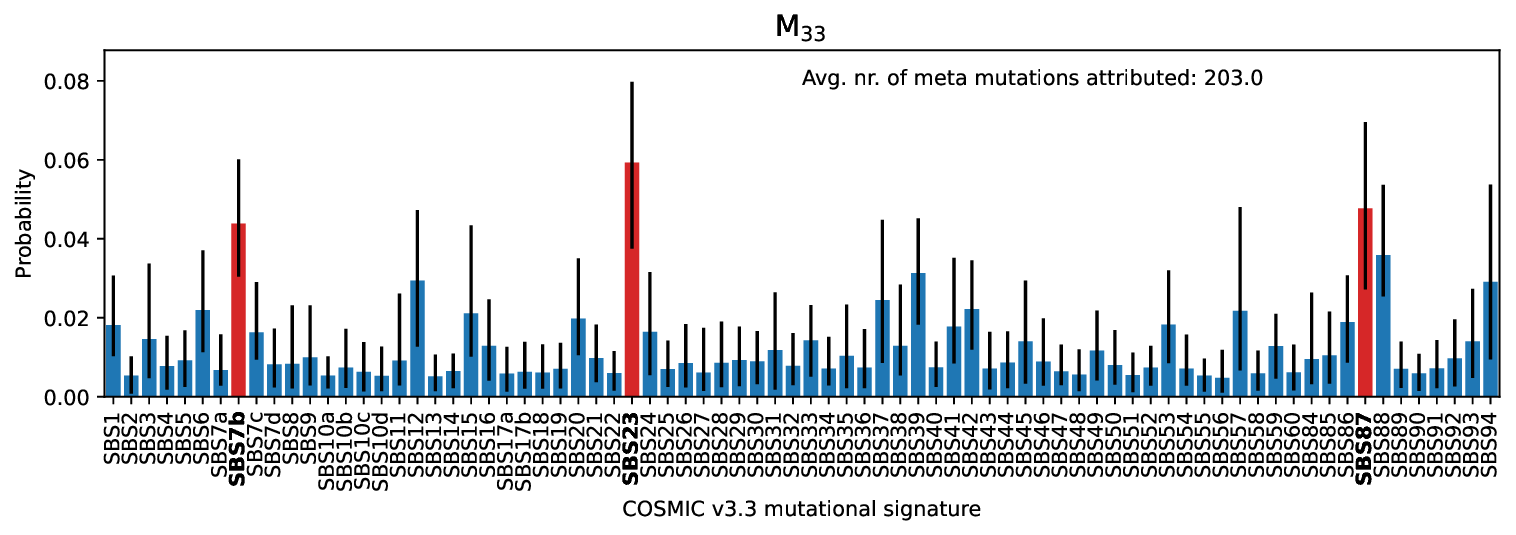}
        &
        \includegraphics[width=0.5\textwidth]{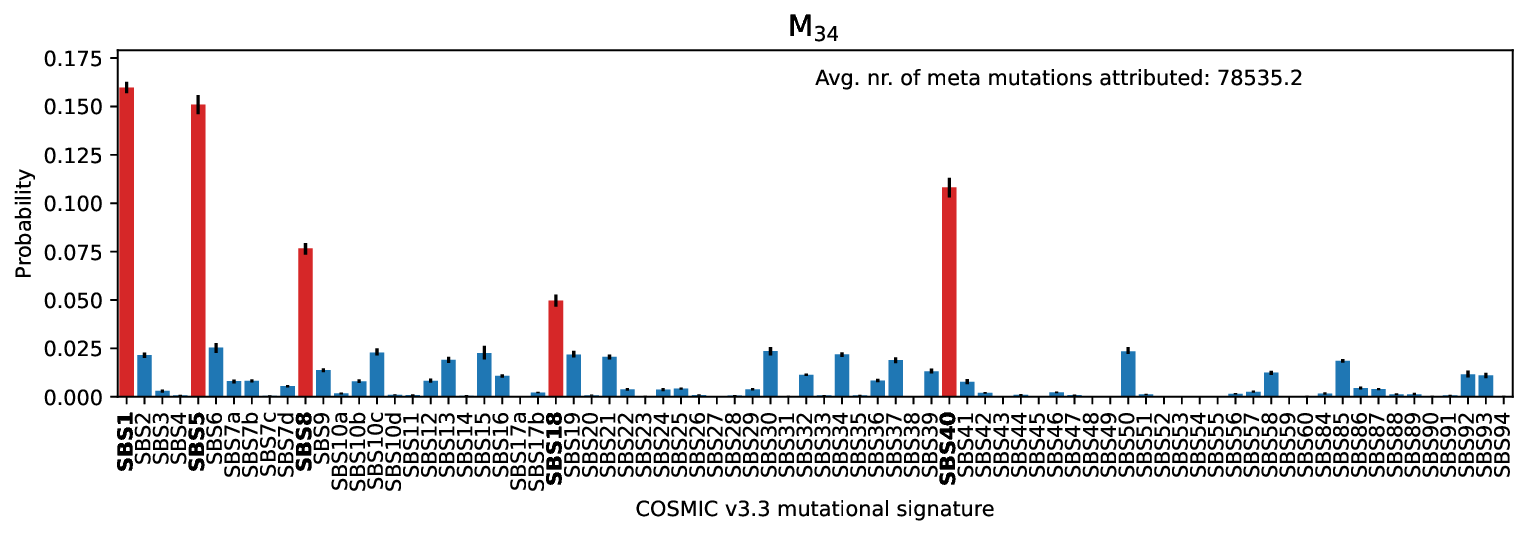}
        \\
        \includegraphics[width=0.5\textwidth]{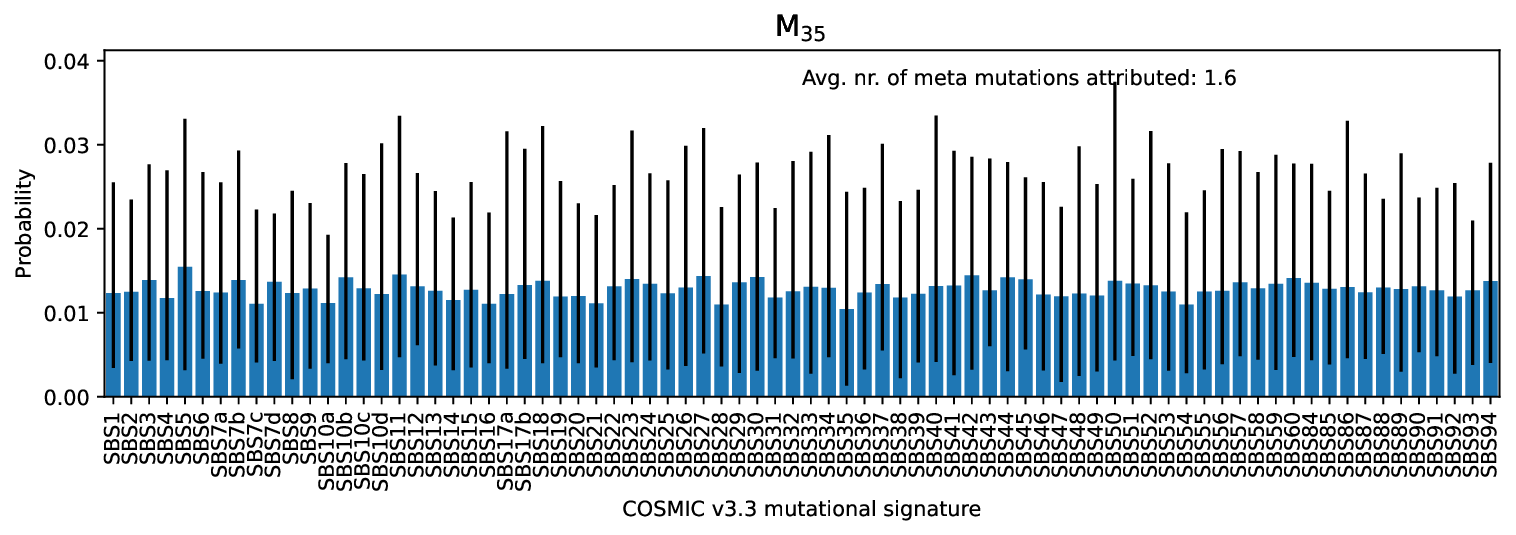}
        &
        \includegraphics[width=0.5\textwidth]{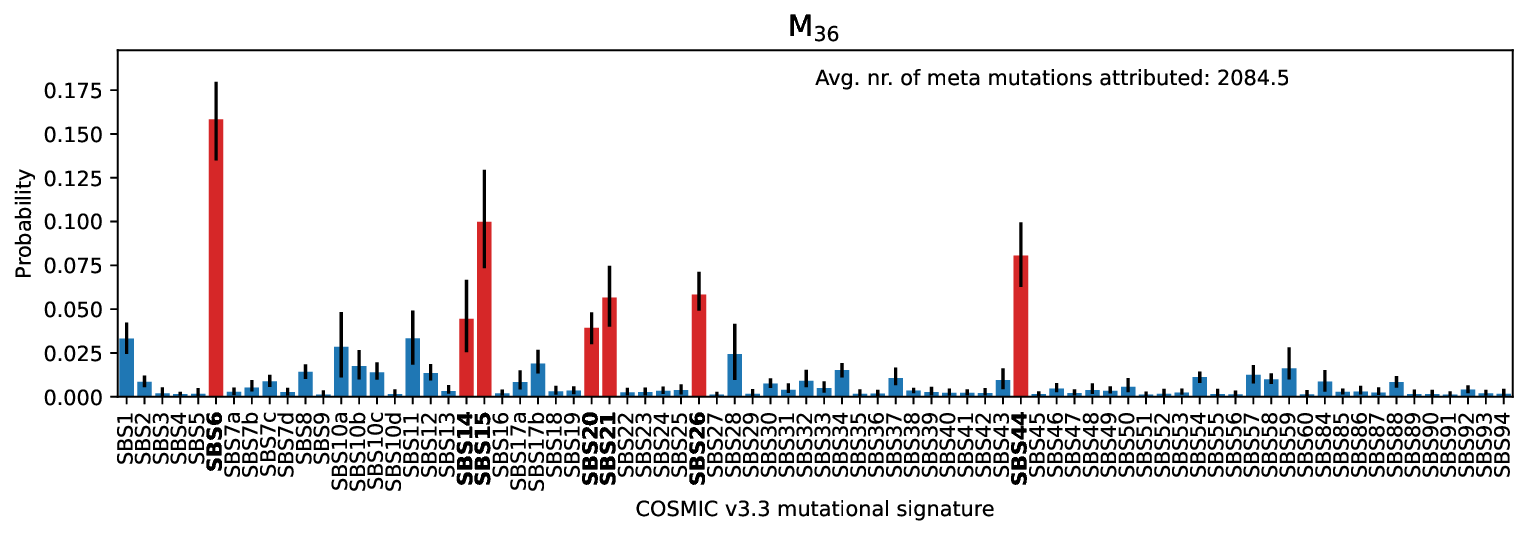}
        \\
        \includegraphics[width=0.5\textwidth]{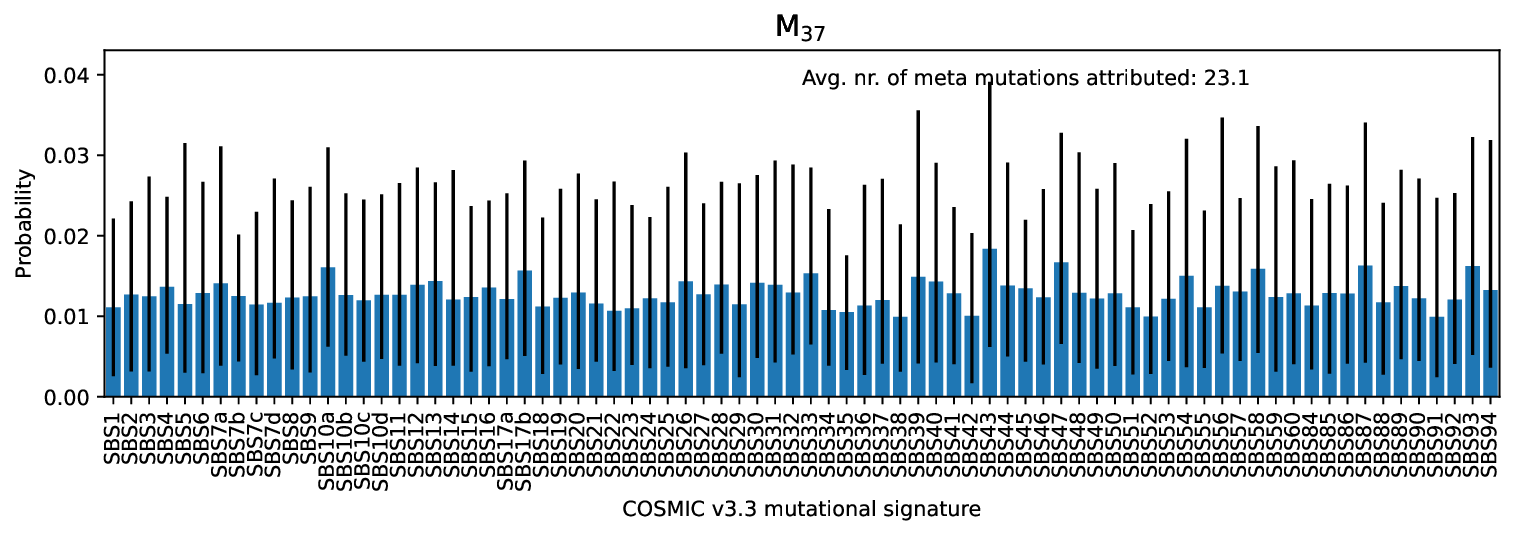}
        &
        \includegraphics[width=0.5\textwidth]{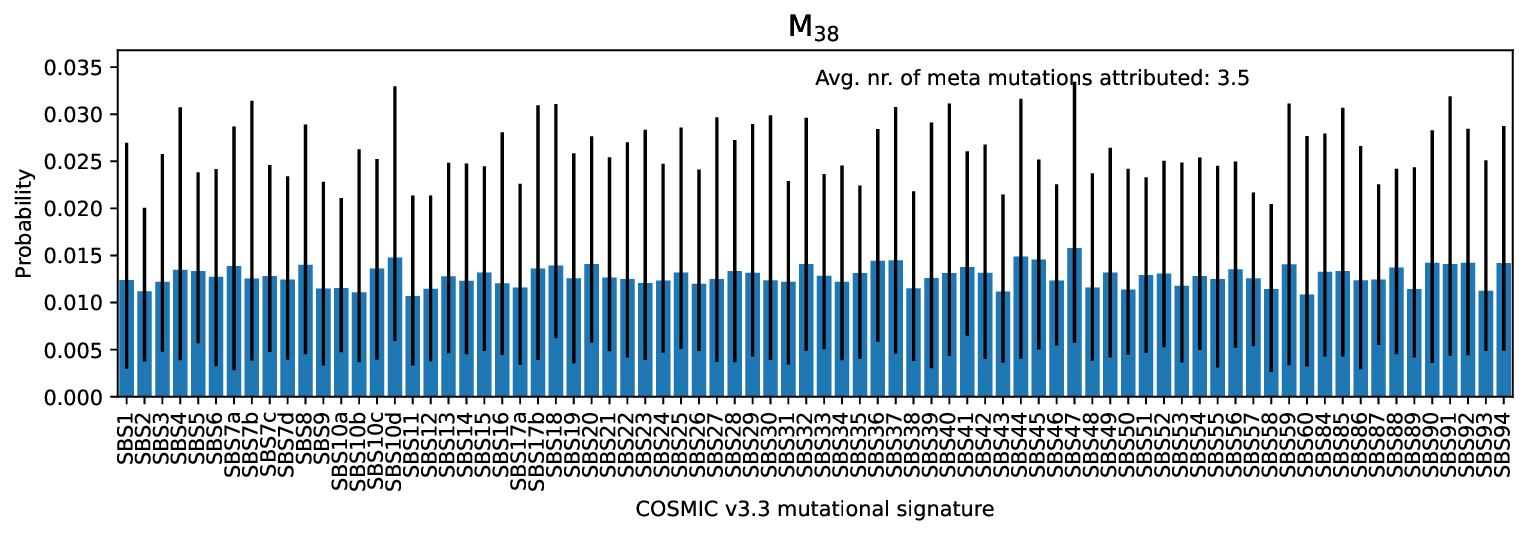}
        \\
        \includegraphics[width=0.5\textwidth]{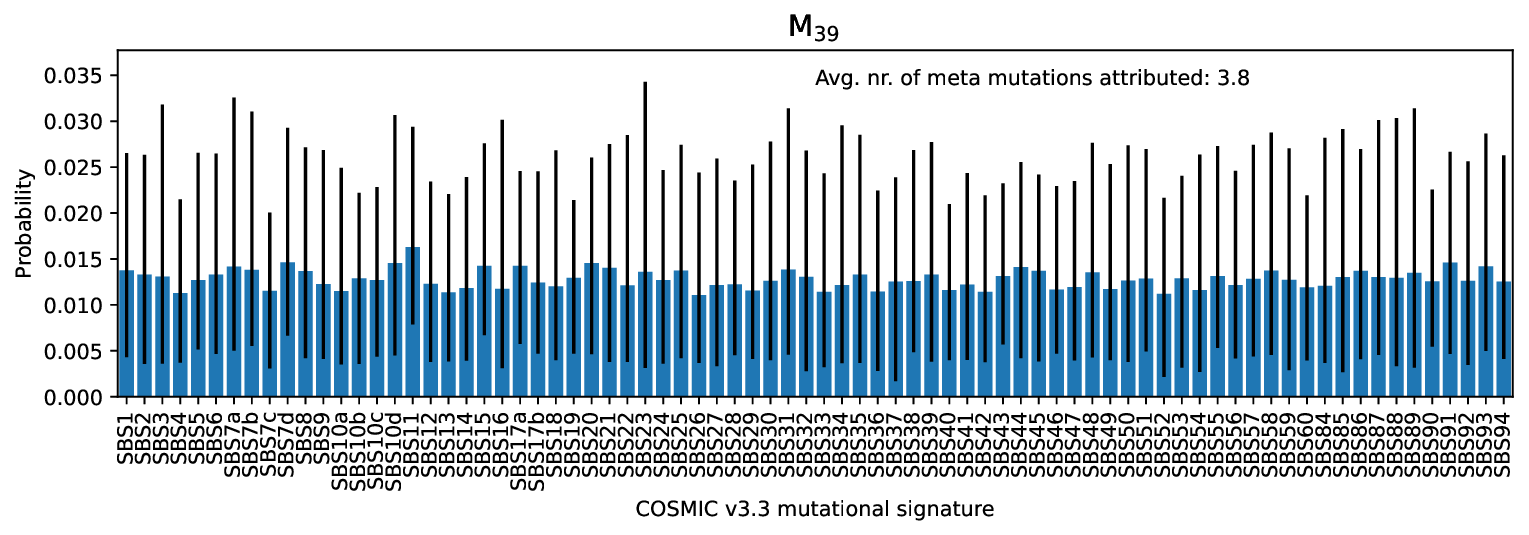}
        &
        \includegraphics[width=0.5\textwidth]{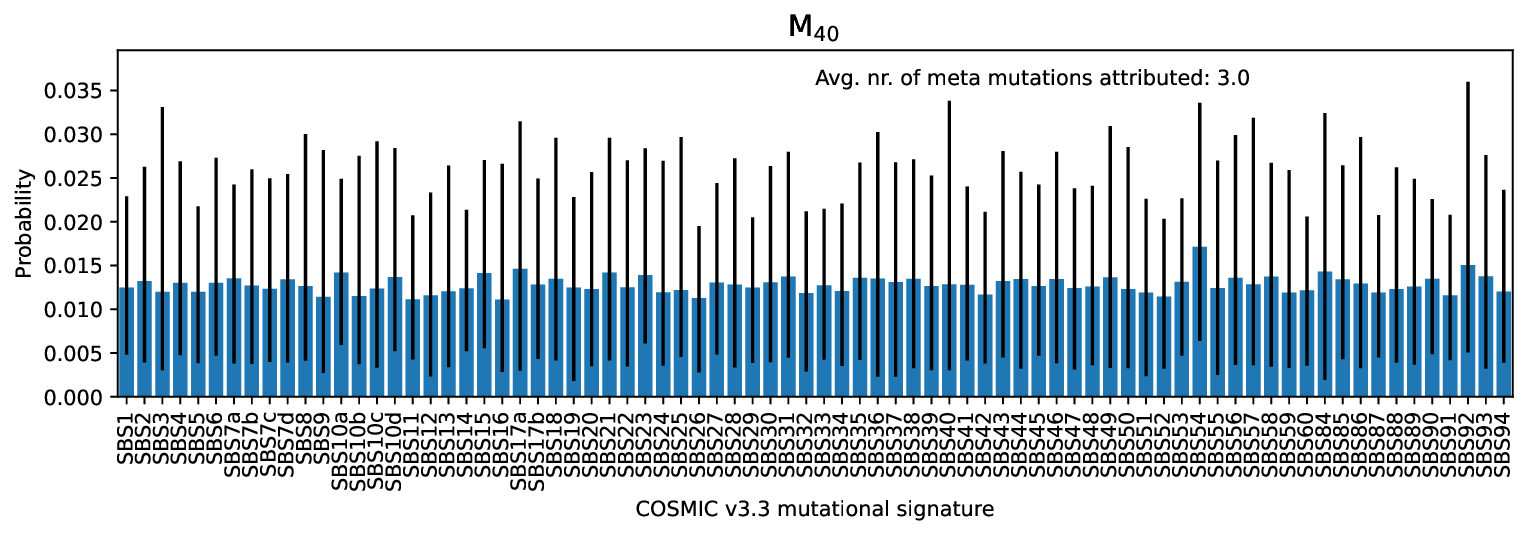}
        \\
        \includegraphics[width=0.5\textwidth]{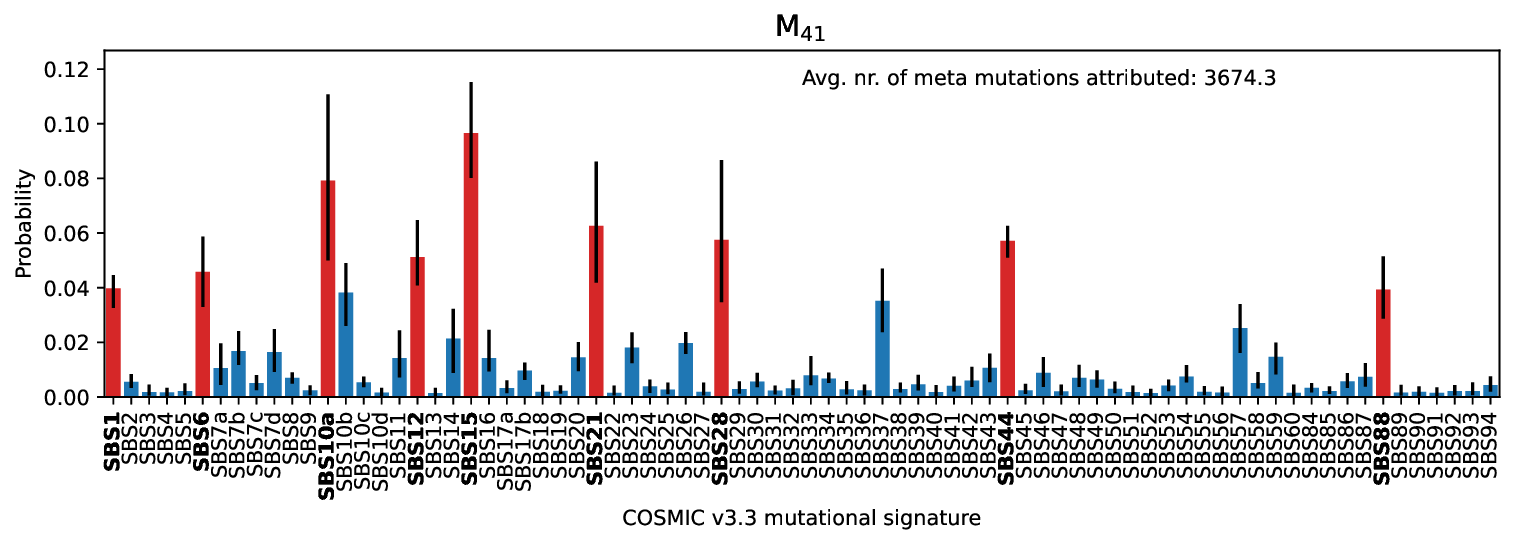}
    \end{tabular}
    \caption{Continuation of Figs.~\ref{fig:meta_signatures_rest_a} and~\ref{fig:meta_signatures_rest_b} listing $k = \text{M}_{33},\dots, \text{M}_{41}$.}
    \label{fig:meta_signatures_rest_c}
\end{figure}

\begin{table}[]
    \caption{Summary statistics of the hyperparameter $c^{(t)}$.}
    \centering
    \begin{tabular}{lrrrrrrrrr}
        \toprule
        {} &    mean &     sd &  hdi\_3\% &  hdi\_97\% &  mcse\_mean &  mcse\_sd &  ess\_bulk &  ess\_tail &  r\_hat \\
        \midrule
        $c^{(3)}$ &   0.298 &  0.036 &   0.230 &    0.351 &      0.016 &    0.012 &       5.0 &      12.0 &   3.10 \\
        $c^{(2)}$ &  33.978 &  4.545 &  26.305 &   39.319 &      2.125 &    1.611 &       5.0 &      13.0 &   3.73 \\
        \bottomrule
    \end{tabular}
\end{table}

\begin{table}[]
    \centering
    \caption{Summary statistics of the hyperparameter $\{r_k\}$ (top-level activation), per meta signature $k$.}
    \resizebox{\columnwidth}{!}{
    \begin{tabular}{lrrrrrrrrr}

        \toprule
        {Meta-signature} &    mean &      sd &  hdi\_3\% &  hdi\_97\% &  mcse\_mean &  mcse\_sd &  ess\_bulk &  ess\_tail &   r\_hat \\
        \midrule
        M$_{1}$  &  0.0005 &  0.0003 &  0.0000 &   0.0011 &     0.0001 &   0.0001 &    7.8437 &   45.1667 &  1.5635 \\
        M$_{2}$  &  0.0274 &  0.0171 &  0.0051 &   0.0489 &     0.0080 &   0.0060 &    5.2076 &   25.3881 &  2.9207 \\
        M$_{3}$  &  0.0462 &  0.0275 &  0.0180 &   0.0934 &     0.0129 &   0.0098 &    5.2138 &   35.0239 &  2.8226 \\
        M$_{4}$  &  0.0023 &  0.0009 &  0.0008 &   0.0037 &     0.0003 &   0.0002 &    7.2637 &   57.5069 &  1.6368 \\
        M$_{5}$  &  0.0090 &  0.0091 &  0.0014 &   0.0255 &     0.0043 &   0.0032 &    5.6255 &   42.9691 &  2.3436 \\
        M$_{6}$  &  0.0174 &  0.0141 &  0.0025 &   0.0425 &     0.0066 &   0.0050 &    5.1255 &   18.2017 &  3.0118 \\
        M$_{7}$  &  0.0086 &  0.0077 &  0.0009 &   0.0208 &     0.0036 &   0.0027 &    5.9037 &   37.2047 &  2.1290 \\
        M$_{8}$  &  0.0047 &  0.0056 &  0.0006 &   0.0154 &     0.0026 &   0.0020 &    7.2067 &   56.4231 &  1.6564 \\
        M$_{9}$  &  0.0807 &  0.0395 &  0.0297 &   0.1444 &     0.0185 &   0.0140 &    5.1425 &   33.8033 &  2.9455 \\
        M$_{10}$ &  0.0173 &  0.0102 &  0.0033 &   0.0298 &     0.0047 &   0.0036 &    5.6516 &   25.2606 &  2.3022 \\
        M$_{11}$ &  0.0424 &  0.0294 &  0.0070 &   0.0889 &     0.0137 &   0.0104 &    5.2041 &   39.3170 &  2.8706 \\
        M$_{12}$ &  0.0125 &  0.0045 &  0.0049 &   0.0189 &     0.0020 &   0.0015 &    5.4742 &   18.1715 &  2.4575 \\
        M$_{13}$ &  0.0119 &  0.0190 &  0.0000 &   0.0463 &     0.0089 &   0.0067 &    5.3522 &   42.5339 &  2.6005 \\
        M$_{14}$ &  0.0121 &  0.0108 &  0.0000 &   0.0307 &     0.0051 &   0.0038 &    5.1627 &   23.8320 &  2.9830 \\
        M$_{15}$ &  0.0139 &  0.0141 &  0.0000 &   0.0335 &     0.0066 &   0.0050 &    5.3330 &   24.4206 &  2.6387 \\
        M$_{16}$ &  0.0094 &  0.0015 &  0.0066 &   0.0121 &     0.0005 &   0.0004 &    8.5465 &   70.0544 &  1.4770 \\
        M$_{17}$ &  0.0007 &  0.0005 &  0.0000 &   0.0017 &     0.0002 &   0.0002 &    6.3390 &   41.6786 &  1.8971 \\
        M$_{18}$ &  0.0015 &  0.0009 &  0.0001 &   0.0029 &     0.0004 &   0.0003 &    6.1458 &   22.9947 &  1.9816 \\
        M$_{19}$ &  0.0069 &  0.0036 &  0.0015 &   0.0117 &     0.0016 &   0.0012 &    5.9197 &   36.6124 &  2.0946 \\
        M$_{20}$ &  0.0441 &  0.0335 &  0.0124 &   0.1036 &     0.0157 &   0.0119 &    5.0797 &   15.1183 &  3.1108 \\
        M$_{21}$ &  0.2091 &  0.0914 &  0.1374 &   0.3724 &     0.0428 &   0.0324 &    5.2223 &   40.6697 &  2.7940 \\
        M$_{22}$ &  0.0003 &  0.0003 &  0.0000 &   0.0009 &     0.0001 &   0.0001 &    8.9105 &   41.9669 &  1.4437 \\
        M$_{23}$ &  0.0001 &  0.0001 &  0.0000 &   0.0003 &     0.0000 &   0.0000 &   17.0880 &   69.2559 &  1.1846 \\
        M$_{24}$ &  0.0590 &  0.0462 &  0.0209 &   0.1413 &     0.0216 &   0.0164 &    5.4445 &   25.3216 &  2.4660 \\
        M$_{25}$ &  0.0124 &  0.0071 &  0.0048 &   0.0249 &     0.0033 &   0.0025 &    5.8507 &   32.4049 &  2.1428 \\
        M$_{26}$ &  0.0003 &  0.0003 &  0.0000 &   0.0010 &     0.0001 &   0.0001 &    8.8749 &   41.5220 &  1.4720 \\
        M$_{27}$ &  0.0030 &  0.0019 &  0.0001 &   0.0056 &     0.0008 &   0.0006 &    5.8643 &   17.1875 &  2.1408 \\
        M$_{28}$ &  0.0626 &  0.0459 &  0.0172 &   0.1442 &     0.0215 &   0.0163 &    5.1546 &   24.8252 &  2.9731 \\
        M$_{29}$ &  0.0221 &  0.0235 &  0.0007 &   0.0601 &     0.0110 &   0.0083 &    5.5399 &   45.9551 &  2.4353 \\
        M$_{30}$ &  0.0056 &  0.0035 &  0.0004 &   0.0101 &     0.0016 &   0.0012 &    5.3785 &   26.2501 &  2.5512 \\
        M$_{31}$ &  0.0044 &  0.0070 &  0.0000 &   0.0175 &     0.0033 &   0.0025 &    6.0873 &   22.1597 &  2.0126 \\
        M$_{32}$ &  0.0991 &  0.0535 &  0.0183 &   0.1604 &     0.0251 &   0.0190 &    5.2073 &   48.6473 &  2.8941 \\
        M$_{33}$ &  0.0004 &  0.0003 &  0.0000 &   0.0010 &     0.0001 &   0.0001 &    8.3580 &   80.6760 &  1.4931 \\
        M$_{34}$ &  0.1459 &  0.0950 &  0.0333 &   0.2602 &     0.0445 &   0.0338 &    5.1767 &   33.8603 &  2.9398 \\
        M$_{35}$ &  0.0000 &  0.0001 &  0.0000 &   0.0002 &     0.0000 &   0.0000 &  115.4890 &   95.5287 &  1.0455 \\
        M$_{36}$ &  0.0020 &  0.0017 &  0.0000 &   0.0050 &     0.0008 &   0.0006 &    5.4310 &   59.6550 &  2.5273 \\
        M$_{37}$ &  0.0001 &  0.0001 &  0.0000 &   0.0003 &     0.0000 &   0.0000 &   13.4681 &   93.1817 &  1.2433 \\
        M$_{38}$ &  0.0001 &  0.0001 &  0.0000 &   0.0002 &     0.0000 &   0.0000 &   78.6768 &  160.1208 &  1.0418 \\
        M$_{39}$ &  0.0001 &  0.0001 &  0.0000 &   0.0003 &     0.0000 &   0.0000 &   20.6372 &  145.3196 &  1.1420 \\
        M$_{40}$ &  0.0001 &  0.0001 &  0.0000 &   0.0002 &     0.0000 &   0.0000 &   14.2321 &   22.7601 &  1.2294 \\
        M$_{41}$ &  0.0042 &  0.0057 &  0.0000 &   0.0150 &     0.0027 &   0.0020 &    5.4181 &   21.5255 &  2.5456 \\
        \bottomrule
    \end{tabular}
    }
\end{table}

\end{document}